\documentclass[]{youtu}

\usepackage{mathpazo}

\usepackage{graphicx}
\usepackage{tikz}
\usepackage{colortbl}
\usepackage{arydshln}
\usepackage{subcaption}
\usepackage{natbib}
\usepackage{amsmath}
\setcitestyle{numbers,square,comma}
\hypersetup{
  pdftitle={STBridge: Shared-Target Alignment for Bridging Understanding and Generation in UMMs},
  pdfauthor={Ye Wang Hongjun Wang Hao Fang Tongyuan Bai Zuwei Long Peixian Chen Wei Liu Weibo Gu Xing Sun Rui Ma},
  pdfsubject={Unified multimodal models},
  pdfkeywords={STBridge, shared-target alignment, unified multimodal models, image editing, post-training}
}
\newcommand{\corrauthicon}{%
  \tikz[baseline=-0.32ex, x=0.72em, y=0.72em, line width=0.045em]{%
    \draw (0,0) rectangle (1.20,0.72);
    \draw (0,0.72) -- (0.60,0.32) -- (1.20,0.72);
    \draw (0,0) -- (0.48,0.38);
    \draw (1.20,0) -- (0.72,0.38);
  }%
}

\title{STBridge: Shared-Target Alignment for Bridging Understanding and Generation in UMMs}
\renewcommand\authorlist{%
\begin{center}
{\normalsize\rmfamily
Ye Wang$^{1,2 \, *}$ \qquad Hongjun Wang$^{2}$ \qquad Hao Fang$^{2}$ \qquad Tongyuan Bai$^{1}$ \qquad Zuwei Long$^{2}$\\[0.25em]
Peixian Chen$^{2}$ \qquad Wei Liu$^{2}$\,\textsuperscript{\corrauthicon} \qquad Weibo Gu$^{2 \, \dagger}$ \qquad Xing Sun$^{2}$ \qquad Rui Ma$^{1}$\,\textsuperscript{\corrauthicon}%
}
\end{center}}
\renewcommand\affiliationlist{%
\begin{center}
{\small\rmfamily
$^{1}$Jilin University \qquad $^{2}$Tencent YouTu Lab
}
\end{center}}
\leftheaderlogo{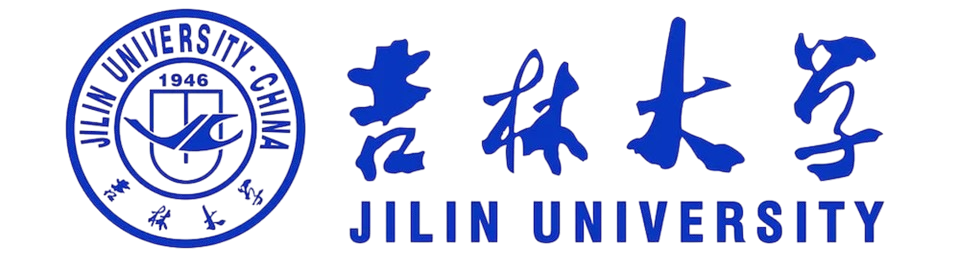}
\rightheaderlogo{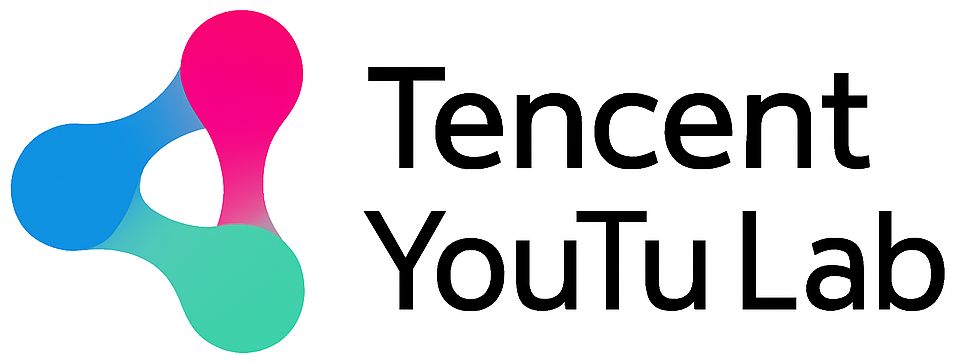}
\makeatletter
\fancypagestyle{titlepagefooter}{%
  \fancyhf{}
  \fancyhead[L]{\@youtuleftheader}
  \fancyhead[R]{\@youturightheader}
  \fancyfoot[C]{\thepage}

}
\makeatother

\date{\today}
\correspondence{frankweiliu@tencent.com}

\begin{document}

\abstract{Unified multimodal models (UMMs) aim to integrate visual understanding and visual generation within a single architecture, but architectural unification alone does not ensure semantic consistency. A model may describe the intended target state correctly while generating an inconsistent edit, or produce an image that contradicts its own interpretation of the instruction. This exposes an understanding-generation alignment gap: linguistic and visual outputs live in different spaces, yet should be governed by the same target semantics.
We study this gap in image editing, where an instruction defines a target state that can be both described and visually realized. Given a source image and an edit instruction, we compare a UMM’s target caption with its edited image to test whether the two outputs converge on the same visual result. Our analysis shows that existing UMMs remain weakly aligned, especially for fine-grained entities, attributes, spatial relations, and local details, indicating that semantic unification may not be achieved by architecture alone.
To bridge this gap, we propose \emph{STBridge}, a shared-target alignment framework that connects understanding and generation through a common target state. In this formulation, the target caption expresses the desired visual result, while the edited image realizes it visually. This replaces separate task-specific optimization paths with a shared information flow from target expression to target realization. STBridge follows an align-then-optimize strategy: supervised fine-tuning first establishes the shared-target channel, and sequential reinforcement learning further refines target-centered coordination.
Across visual understanding, image generation, and image editing benchmarks, STBridge consistently improves over the initialization model, with gains on BLINK (+5.15), CVBench-2D/3D (+1.19/+1.92), OddGridBench (+5.14), MMVP (+2.70), WISE (+1.20), ImgEdit (+0.46), and RISE (+21.00). Alignment analysis further confirms that STBridge narrows the gap between what the model describes and what it generates. These results demonstrate shared-target alignment as an effective post-training strategy for bridging understanding and generation in UMMs.
}
\maketitle
\thispagestyle{titlepagefooter}
\vspace{0.04in}
\begin{center}
\makebox[\textwidth][c]{\includegraphics[width=0.97\textwidth]{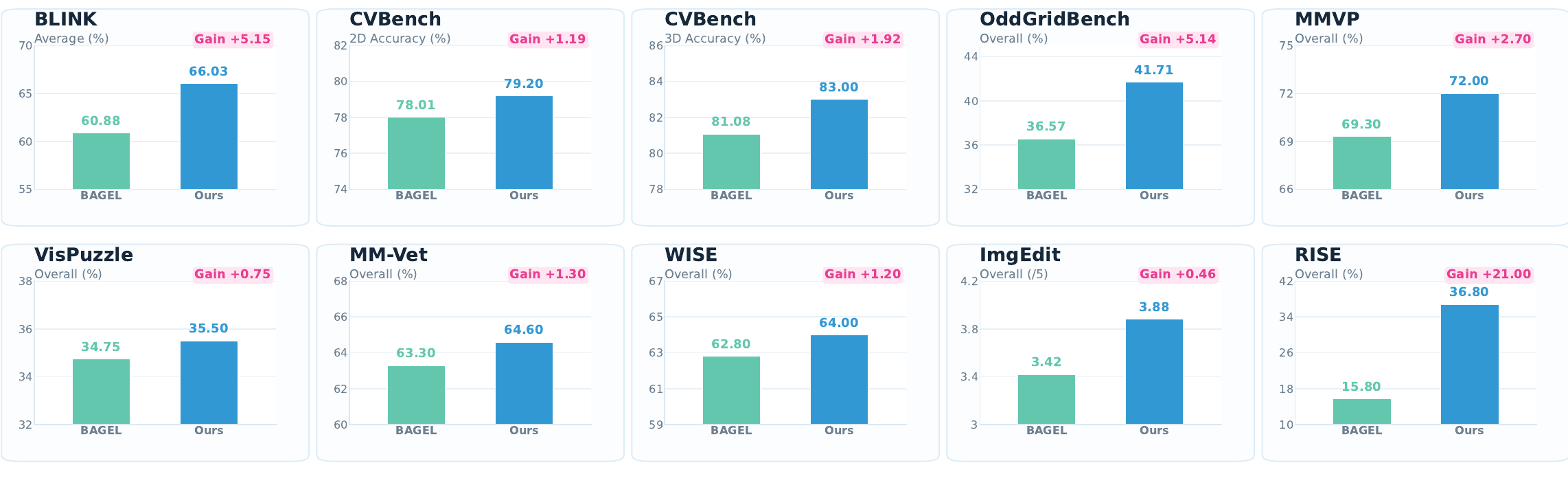}}\\[3pt]
\captionsetup{type=figure,hypcap=false}
\caption{\textbf{Aggregate results.} We compare BAGEL and Ours across visual understanding, image generation, and image editing benchmarks, reporting benchmark-specific metrics and absolute gains.}
\label{fig:aggregate_summary}
\par\vspace{2pt}
\makebox[\textwidth][l]{\raisebox{0pt}[0pt][0pt]{\begin{minipage}[t]{0.62\textwidth}\footnotesize\setbox0=\hbox{$^{*}$Work was done when Ye Wang interned at Tencent YouTu Lab.}\rule{\wd0}{0.4pt}\par\vspace{1.5pt}\begin{tabular}[t]{@{}l@{}}\textsuperscript{\corrauthicon}Co-corresponding authors \qquad $^{\dagger}$Project leader\\$^{*}$Work was done when Ye Wang interned at Tencent YouTu Lab.\end{tabular}\end{minipage}}}
\end{center}
\clearpage
\begin{figure}[p]
\centering
\vspace*{-0.25in}
\makebox[\textwidth][c]{\includegraphics[width=\textwidth]{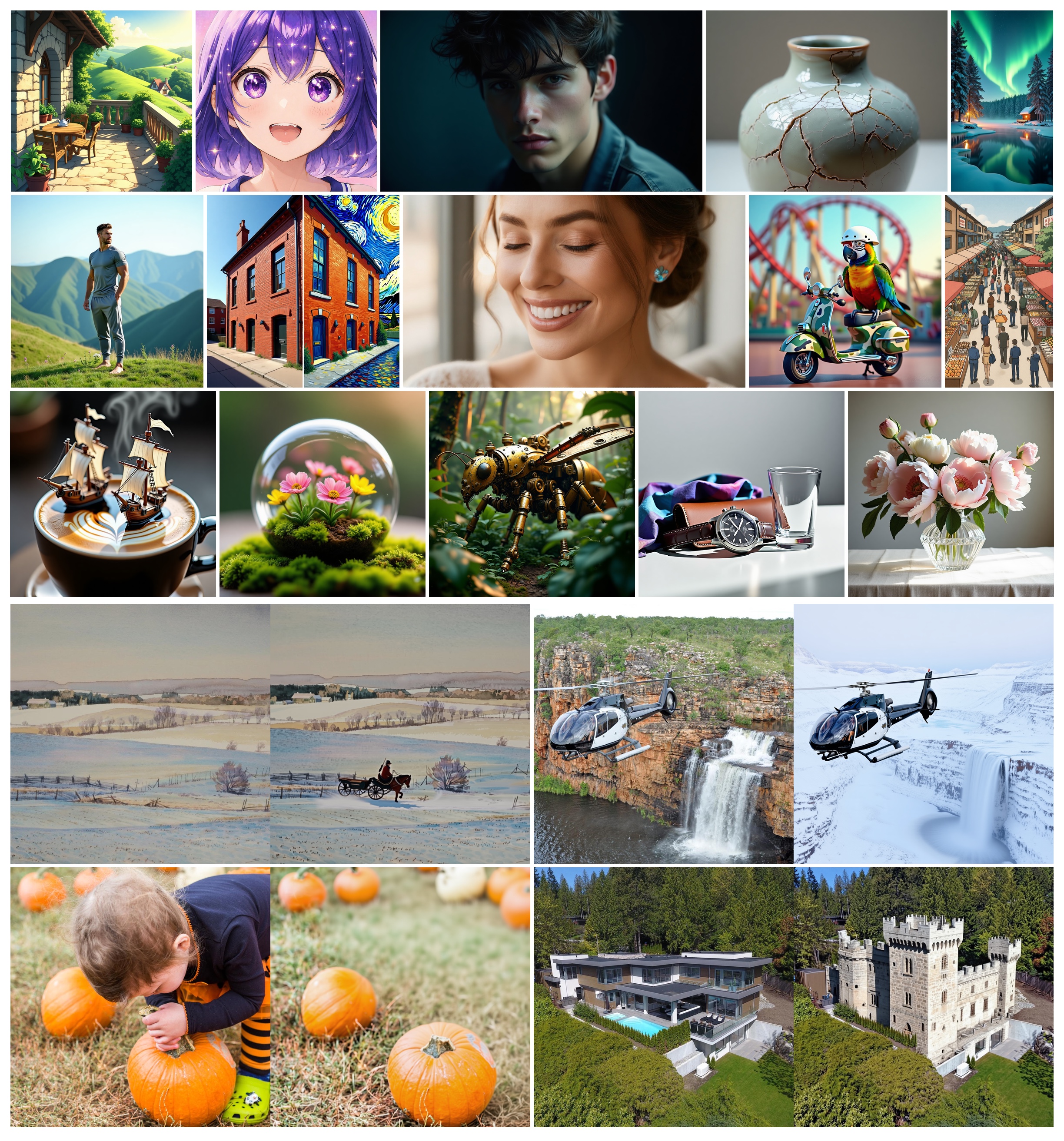}}
\caption{\textbf{Generation and editing examples.} Representative qualitative examples from STBridge are arranged as a staggered mosaic. \textbf{Above:} image-generation examples with diverse resolutions and aspect ratios. \textbf{Below:} instruction-guided image-editing examples.}
\label{fig:layout_mosaic}
\end{figure}
\clearpage

\begin{figure}[p]
\centering
\begingroup
\vspace*{-0.50in}
\setlength{\abovecaptionskip}{2pt}
\setlength{\belowcaptionskip}{0pt}
\newcommand{\wisepanelheight}{0.140\textheight}
\newcommand{\wisepromptheight}{0.023\textheight}
\newcommand{\wiserowheight}{\dimexpr\wisepromptheight+\wisepanelheight+0.8pt\relax}
\newcommand{\wisepromptfont}{\fontsize{7pt}{6.2pt}\selectfont}
\definecolor{wiseLogoMagenta}{RGB}{223,66,146}
\newcommand{\wisekey}[1]{\textcolor{wiseLogoMagenta}{\textbf{#1}}}
\newcommand{\wisepanel}[4]{%
\begin{minipage}[t]{0.498\linewidth}
\begin{minipage}[t][\wisepromptheight][c]{\linewidth}
\centering\wisepromptfont\textbf{Prompt:} \emph{#1}
\end{minipage}\par\vspace{0.8pt}
\begin{minipage}[t]{0.49\linewidth}
\vspace{0pt}
\begin{tcolorbox}[colback=youtuLightBlue,colframe=youtuBlue!25,boxrule=0.25pt,arc=1.2pt,left=1.4pt,right=1.4pt,top=1.4pt,bottom=1.4pt,width=\linewidth,fit to height=\wisepanelheight,fit basedim=6.4pt,fit skip=1.05,fit algorithm=fontsize]
{\raggedright\setlength{\parindent}{0pt}\setlength{\parskip}{0pt}\textbf{#2.} #3\par}
\end{tcolorbox}
\end{minipage}\hfill
\begin{minipage}[t]{0.505\linewidth}
\vspace{0pt}
\centering
\includegraphics[width=\linewidth,height=\wisepanelheight,keepaspectratio]{#4}
\end{minipage}
\end{minipage}}
\newcommand{\wisefolderlabel}[1]{%
\begin{minipage}[t][\wiserowheight][c]{0.030\linewidth}
\centering\rotatebox{90}{\textbf{\footnotesize #1}}
\end{minipage}}
\newcommand{\wiserow}[3]{%
\noindent\wisefolderlabel{#1}\hspace{0.006\linewidth}%
\begin{minipage}[t]{0.964\linewidth}
#2\hfill
#3
\end{minipage}\par\vspace{1.0pt}}
\newcommand{\wiseculturalone}{The model should generate an image of a \wisekey{hummingbird}, highlighting its \wisekey{small size} and \wisekey{iridescent feathers}, perched on a \wisekey{flower or branch}. The complete, polished prompt is: A tiny hummingbird perched delicately on a vibrant flower or thin branch, showcasing its iridescent feathers that shimmer with hues of green, blue, and purple under soft, natural lighting. The scene is captured in an ultra-realistic style with a shallow depth of field, creating a blurred background of warm bokeh tones that emphasize the bird's intricate details and small stature. The composition is intimate and focused, highlighting the bird's agility and elegance in its natural habitat, with a serene and tranquil atmosphere.}
\newcommand{\wiseculturaltwo}{The model should generate an image of the \wisekey{Wright brothers' first successful powered aircraft}, the \wisekey{Wright Flyer}, in its early 20th-century design, featuring \wisekey{wooden wings}, a \wisekey{fabric-covered fuselage}, and a \wisekey{propeller-driven engine}. The complete, polished prompt is: A detailed depiction of the Wright Flyer, the first successful powered aircraft, captured in its early 20th-century design, featuring wooden wings with a smooth, polished finish and a fabric-covered fuselage that appears lightweight and aerodynamic. The aircraft is shown in mid-flight, with its propeller-driven engine clearly visible at the front, surrounded by a clear sky with soft, natural lighting that highlights the texture of the wood and fabric. The scene is rendered in a realistic style with sharp details, emphasizing the historical significance and engineering ingenuity of this pioneering invention, set against a backdrop of open air with faint clouds, evoking a sense of early aviation history.}
\newcommand{\wisephysicsone}{The model should generate an image of a \wisekey{sandy beach} with \wisekey{visible footprints} scattered across the surface, showing the texture of the sand being \wisekey{compressed and displaced} by the weight of the feet, while the overall scene remains \wisekey{bright and natural}. The complete, polished prompt is: A bright and natural beach scene featuring a sandy shoreline under clear daylight, with scattered footprints clearly visible across the surface of the sand, showing the texture and compression of the grains displaced by the weight of the feet. The lighting is soft and even, highlighting the golden tones of the sand and the fine details of the footprints, while the overall atmosphere feels serene and untouched, with no other objects or people present in the frame.}
\newcommand{\wisephysicstwo}{The model should generate an image of a \wisekey{soda can} placed outdoors on a \wisekey{humid day}, showing \wisekey{condensation} forming on the surface, with \wisekey{droplets visible on the metal} and a \wisekey{slightly foggy atmosphere} to indicate moisture in the air. Here's the full, detailed prompt: A close-up, ultra-realistic image of a single soda can resting outdoors on a humid day, its metallic surface glistening with visible droplets of condensation forming along the curves and edges, while a faint, misty atmosphere surrounds the can to emphasize the moisture in the air. The lighting is soft and diffused, suggesting overcast or cloudy weather conditions, and the background is slightly blurred to focus attention on the can, with natural tones and realistic textures that highlight the interaction between the liquid and the metal surface.}
\newcommand{\wisespaceone}{The model should generate an image of a \wisekey{glass of red wine}, highlighting its \wisekey{rich color} and \wisekey{elegant presentation}, representing a common drink associated with the region. The comprehensive prompt is: A crystal-clear glass of red wine, filled to the brim, sits elegantly on a polished wooden table in soft, warm lighting that enhances the rich, deep crimson hue of the liquid. The wine displays subtle reflections and a velvety texture, with a slight sheen on the surface that catches the ambient light. The background is softly blurred with a bokeh effect, featuring hints of a rustic wine cellar or vineyard atmosphere, creating a refined and inviting mood. The composition is focused on the wine glass, with a realistic and detailed rendering of the glassware and liquid, emphasizing its elegance and sophistication, captured in a high-resolution, photorealistic style.}
\newcommand{\wisespacetwo}{The model should generate an image of the \wisekey{Eiffel Tower}, a famous \wisekey{iron lattice tower} in \wisekey{Paris, France}. The complete, polished prompt is: A majestic view of the Eiffel Tower, a towering iron lattice structure in Paris, France, captured in a clear and bright daylight setting with soft, natural lighting that highlights the intricate geometric patterns of its steel framework. The perspective is slightly low-angle to emphasize the tower's grandeur, set against a vibrant blue sky with faint wisps of clouds, and framed by lush green trees in the foreground for a balanced composition. The image should feel realistic and visually striking, with fine details visible in the metalwork and a crisp, high-resolution finish that conveys the iconic status of this landmark.}
\newcommand{\wisetimetableone}{The image should depict a \wisekey{quiet urban street in the evening}, with \wisekey{streetlights glowing} and \wisekey{shop shutters pulled down}, indicating the \wisekey{transition from day to night}. Complete prompt generated: A quiet urban street in the evening, captured in a realistic and cinematic style with soft ambient lighting and a warm, golden glow from glowing streetlights. The scene features closed shop shutters lining the sidewalks, and the street is empty, conveying a peaceful and still atmosphere. The perspective is slightly elevated, looking down the length of the road, with subtle shadows and reflections on the pavement to emphasize the transition from day to night. The color palette is muted and earthy, dominated by deep blues and warm yellows, with a slightly overcast sky in the background to suggest the fading light of dusk. The overall mood is calm and serene, with fine architectural details on the buildings visible in the background and foreground.}
\newcommand{\wisetimetabletwo}{The prompt requires \wisekey{time-zone reasoning}: when it is \wisekey{1 AM in Melbourne}, \wisekey{London is on the previous afternoon or early evening} depending on daylight saving time, so the campus should appear in \wisekey{daylight rather than at night}. Complete prompt generated: A vibrant university campus in London during the previous afternoon or early evening, featuring students walking in groups between historic stone buildings and modern architecture under natural daylight with warm, low-angle sunlight. The atmosphere is lively yet calm, with soft shadows across the grassy lawns and paved walkways, making it clear that the scene is not midnight in London but the corresponding local daytime period. The scene includes details like students carrying backpacks, reading books, or chatting, surrounded by mature trees and manicured greenery, with a mix of classical and contemporary architectural styles visible in the background. The perspective is slightly elevated, capturing the scale of the campus and the time-zone-informed lighting, rendered in ultra-realistic detail with cinematic lighting and natural color tones.}
\newcommand{\wisebiologyone}{The image should depict a \wisekey{maple tree in autumn}, with its leaves transitioning from green to \wisekey{orange, yellow, and red}, indicating the \wisekey{breakdown of chlorophyll} and the exposure of other pigments. The complete, polished prompt is: A majestic maple tree in the autumn season, its branches adorned with leaves in vibrant shades of orange, yellow, and red, showcasing the natural process of chlorophyll breakdown and the emergence of other pigments. The scene is set in soft, warm lighting that enhances the richness of the colors and creates a serene, atmospheric mood, with a shallow depth of field and gentle bokeh in the background to emphasize the tree's intricate details and the interplay of light and shadow on the foliage. Ultra-realistic rendering with cinematic clarity and natural textures, capturing the beauty of seasonal change in a tranquil outdoor environment.}
\newcommand{\wisebiologytwo}{The model should generate an image of a \wisekey{loaf of bread} with visible patches of \wisekey{fuzzy, greenish or blackish mold} growing on its surface, indicating the presence of \wisekey{fungi and decay}. The complete, polished prompt is: A close-up, ultra-realistic image of a loaf of bread with a textured crust, showing distinct, fuzzy patches of greenish and blackish mold growing unevenly across its surface, highlighting the decay and fungal presence. The lighting is soft and natural, emphasizing the organic details and texture of the bread and mold, with a slightly grainy, macro-style aesthetic to enhance realism and focus on the subject matter.}
\makebox[\textwidth][c]{%
\begin{minipage}[t]{1.08\textwidth}
\wiserow{Cultural}{\wisepanel{the smallest bird in the world}{Cultural-1}{\wiseculturalone}{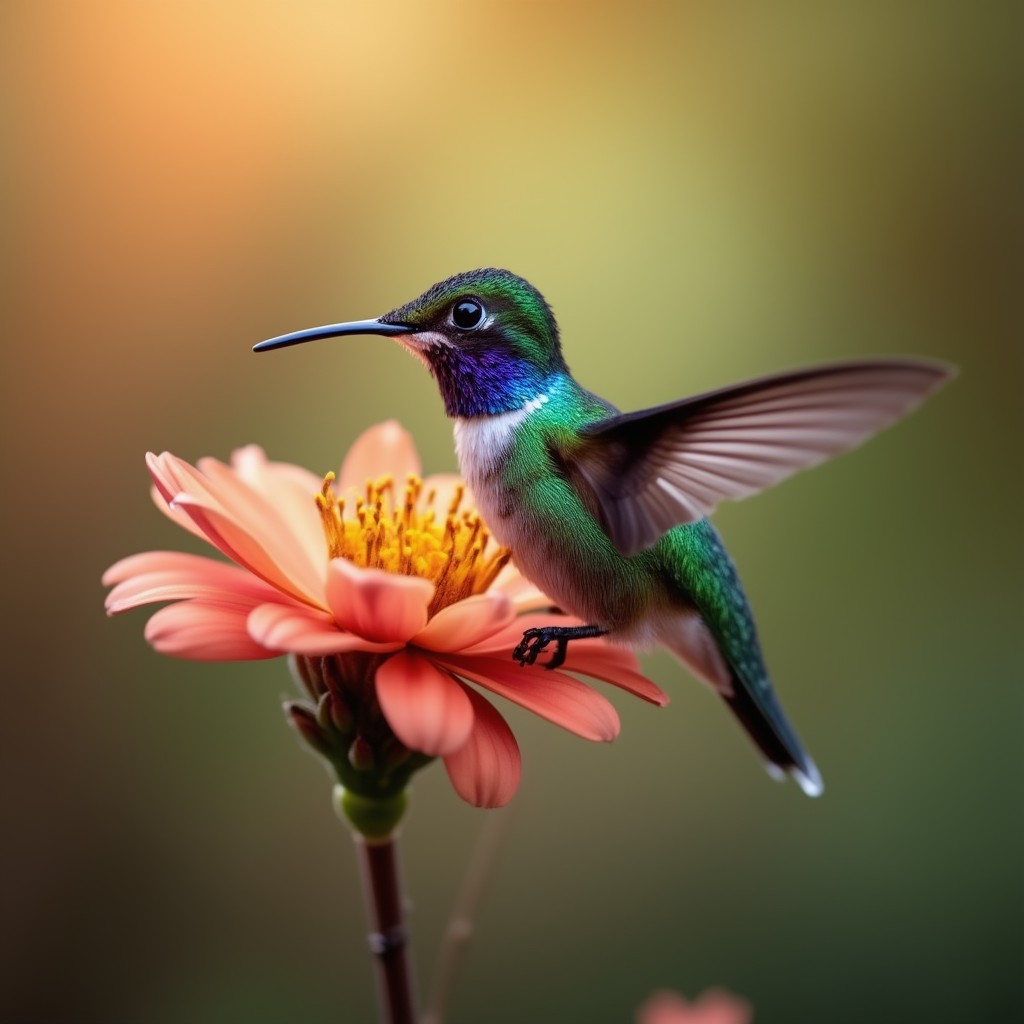}}{\wisepanel{wright brothers greatest invention}{Cultural-2}{\wiseculturaltwo}{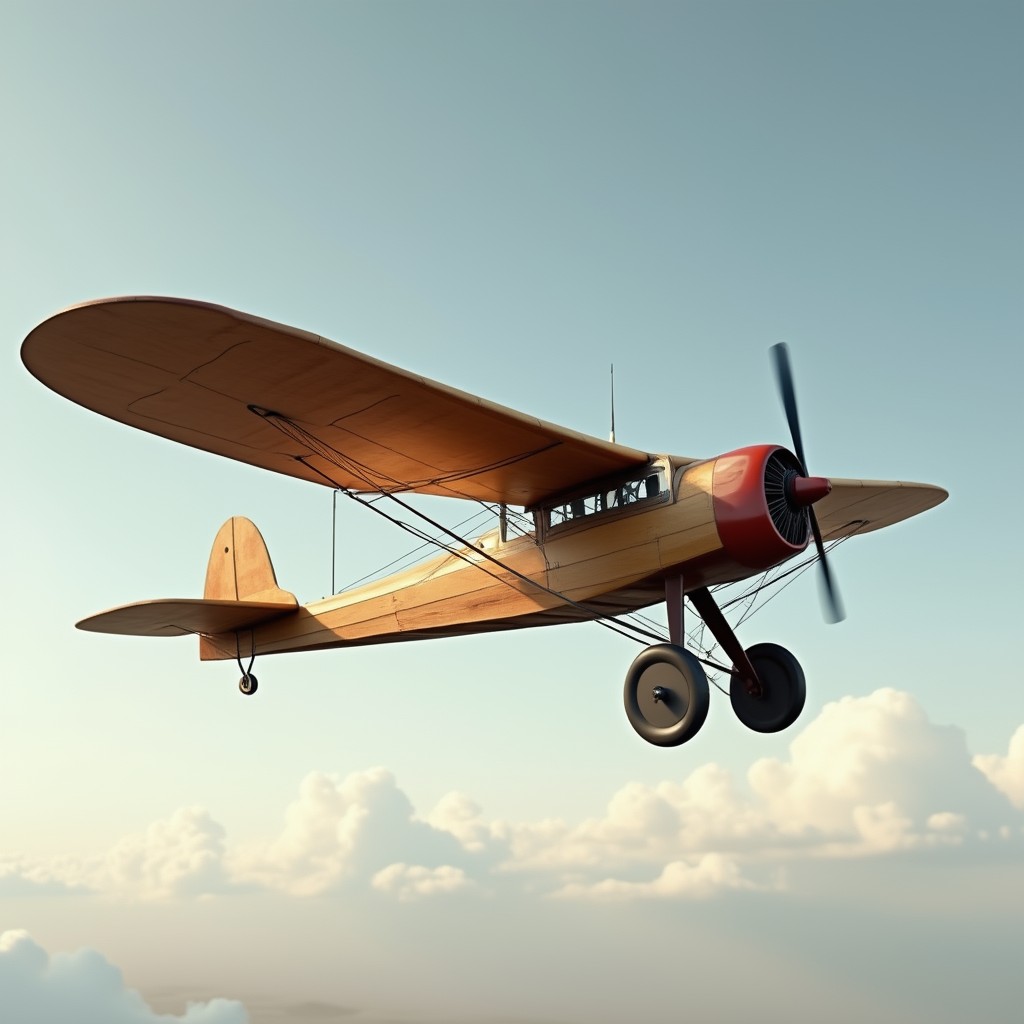}}
\wiserow{Physics}{\wisepanel{a beach after many footsteps}{Physics-1}{\wisephysicsone}{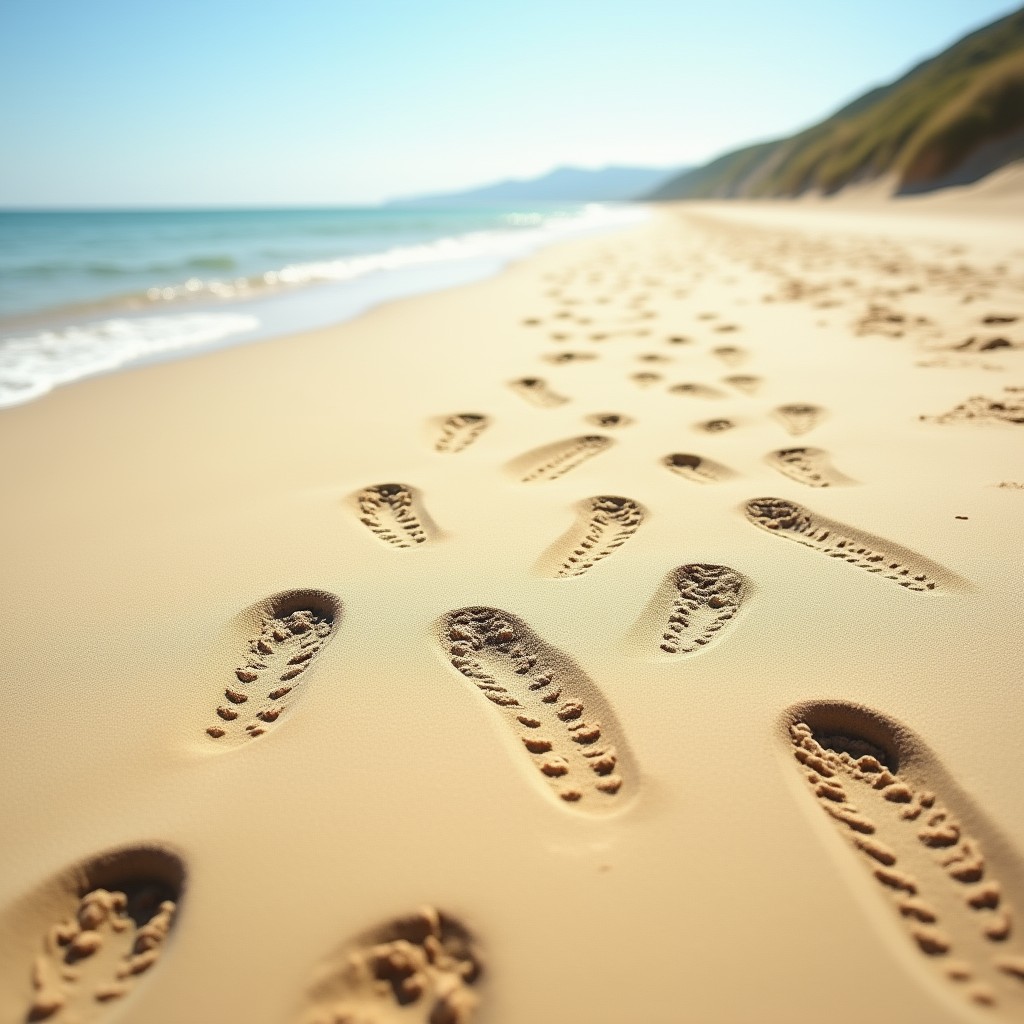}}{\wisepanel{a cold soda can left outside on a humid day, highlighting the state of the can}{Physics-2}{\wisephysicstwo}{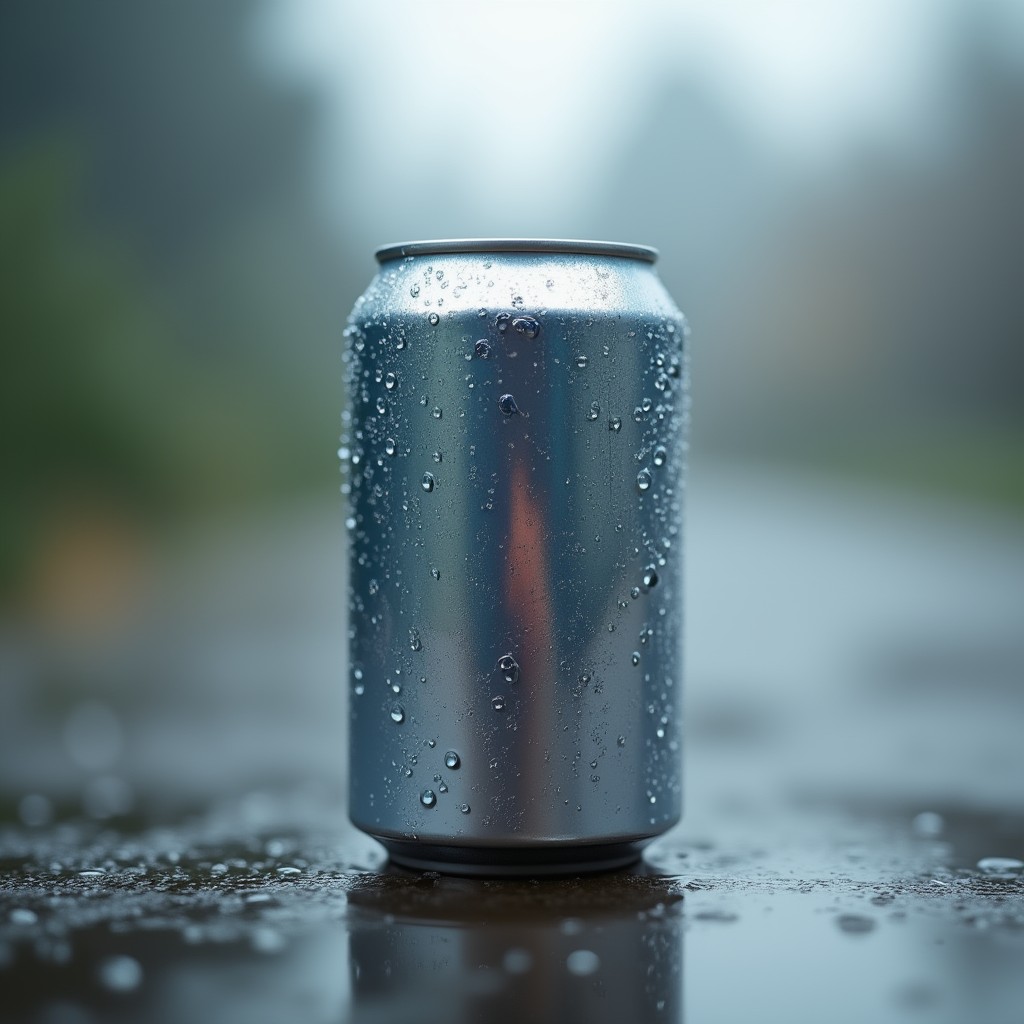}}
\wiserow{Space}{\wisepanel{a typical beverage produced in the country where Bordeaux is located}{Space-1}{\wisespaceone}{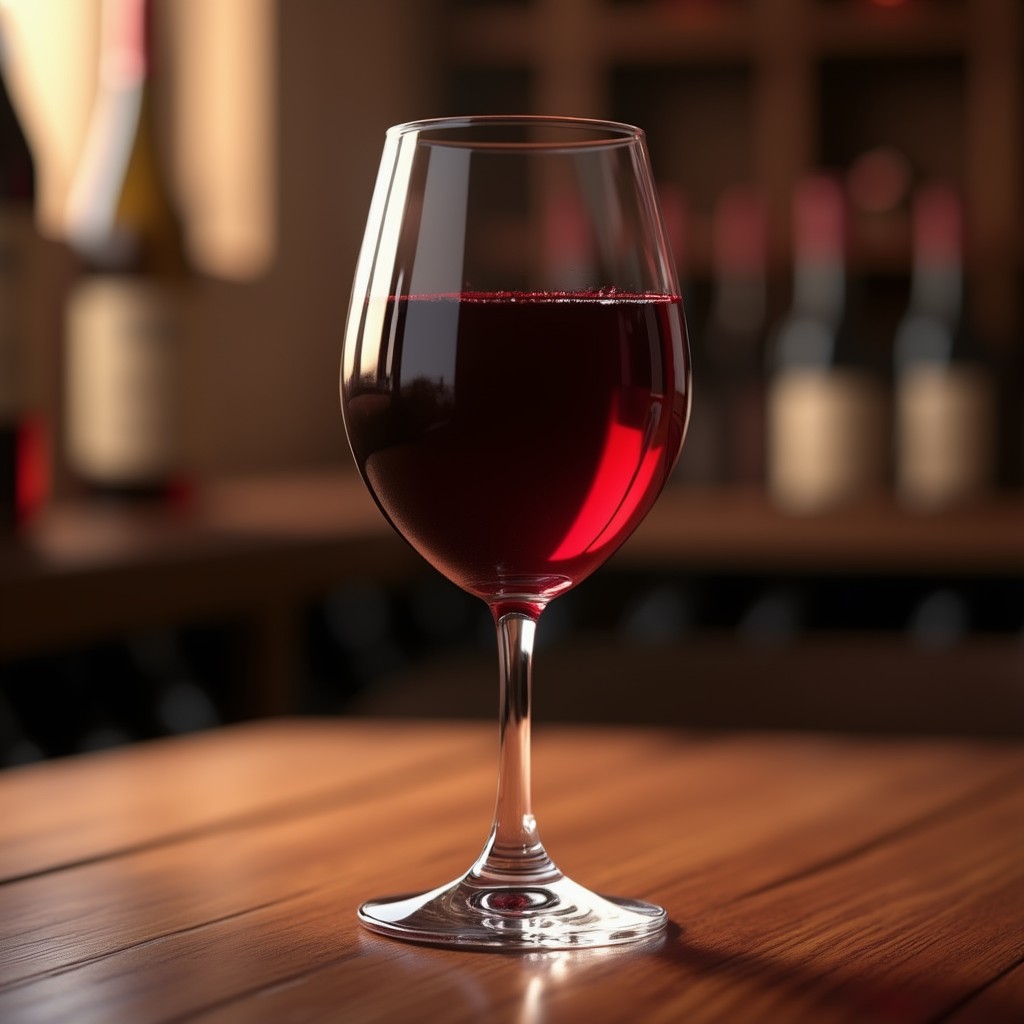}}{\wisepanel{the landmark of the country where Paris is located}{Space-2}{\wisespacetwo}{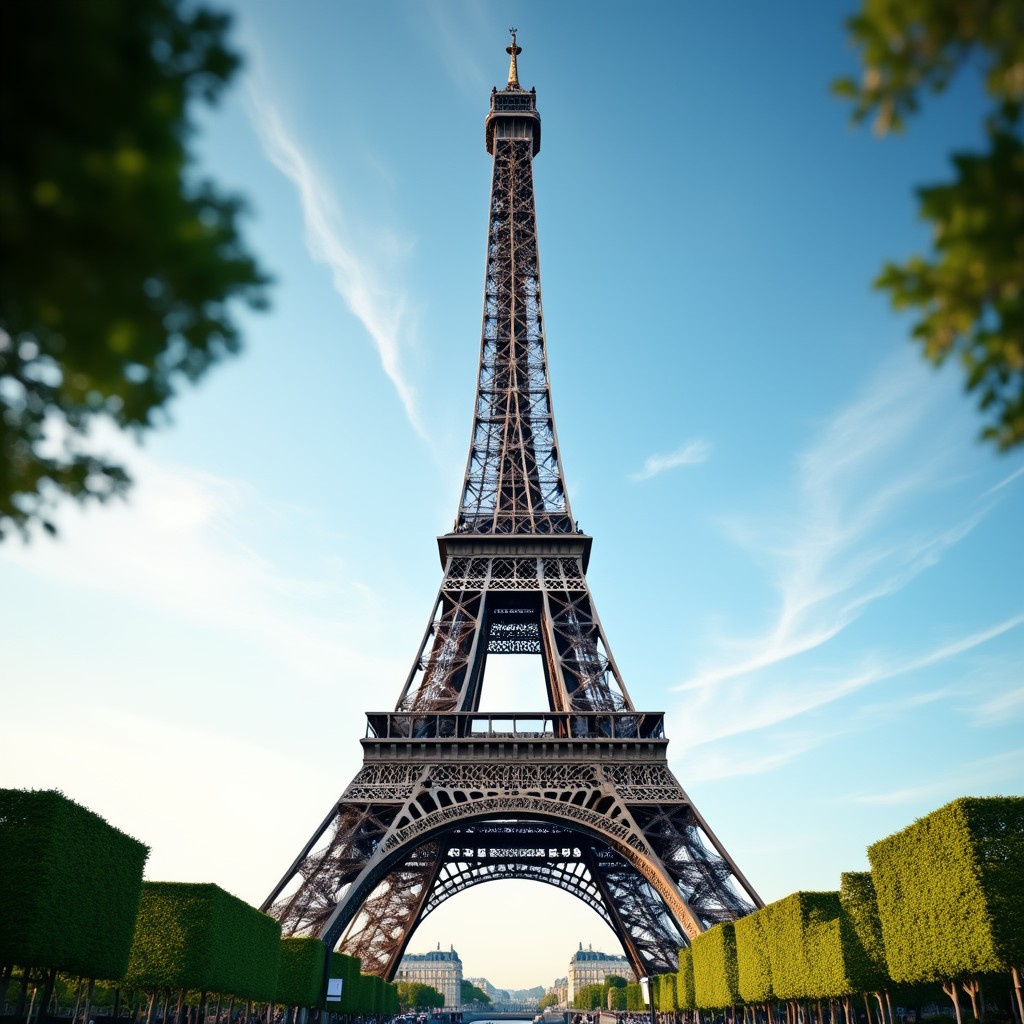}}
\wiserow{Time}{\wisepanel{the street when the streetlights are turned on and the shops are closing}{Time-1}{\wisetimetableone}{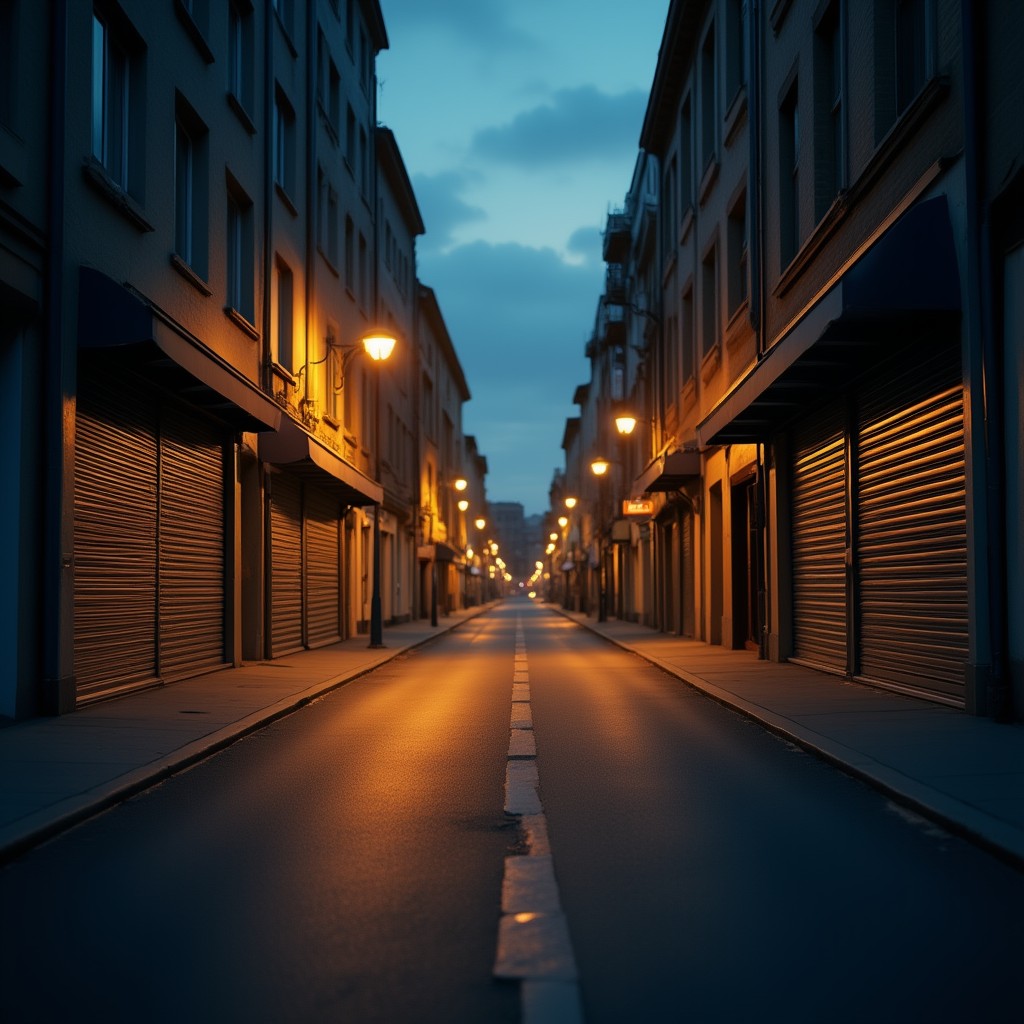}}{\wisepanel{a university campus in London at 1 am Melbourne time}{Time-2}{\wisetimetabletwo}{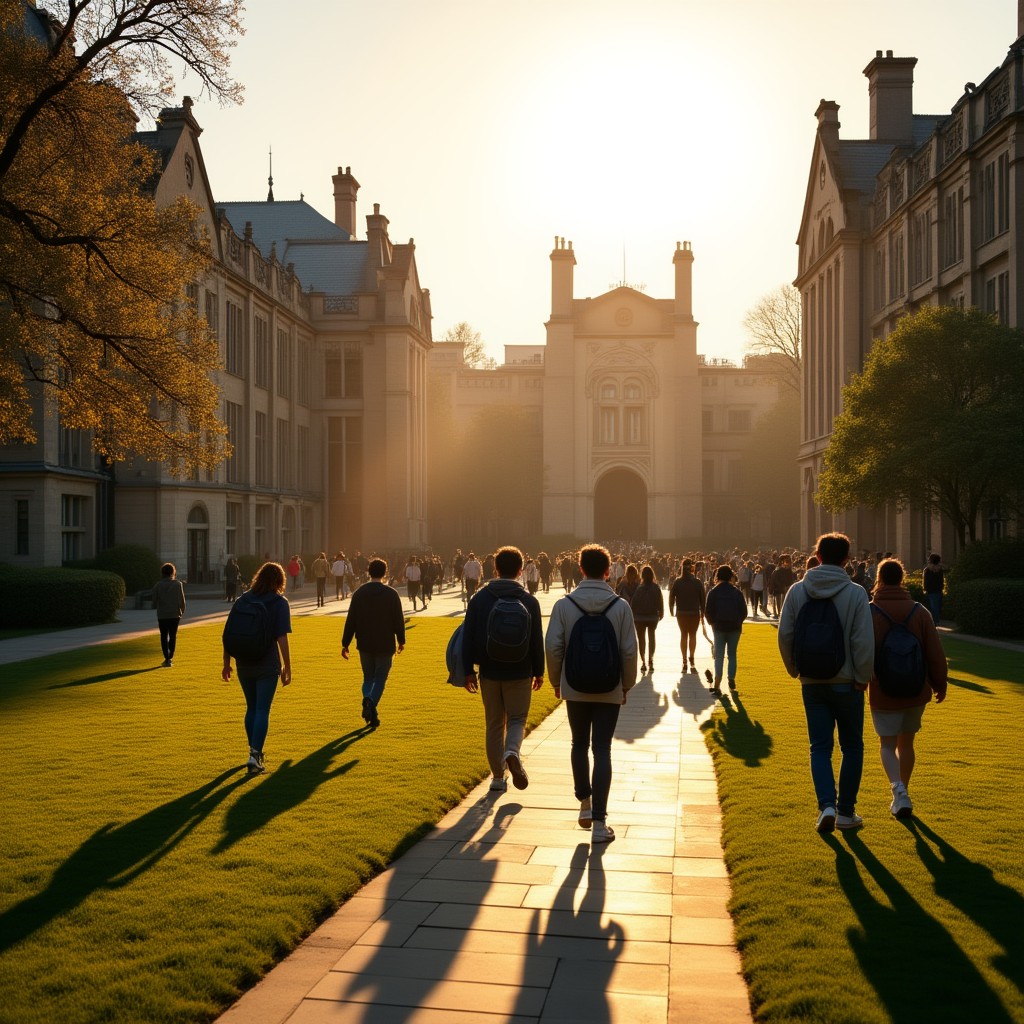}}
\wiserow{Biology}{\wisepanel{a maple tree with leaves exhibiting chlorophyll breakdown}{Biology-1}{\wisebiologyone}{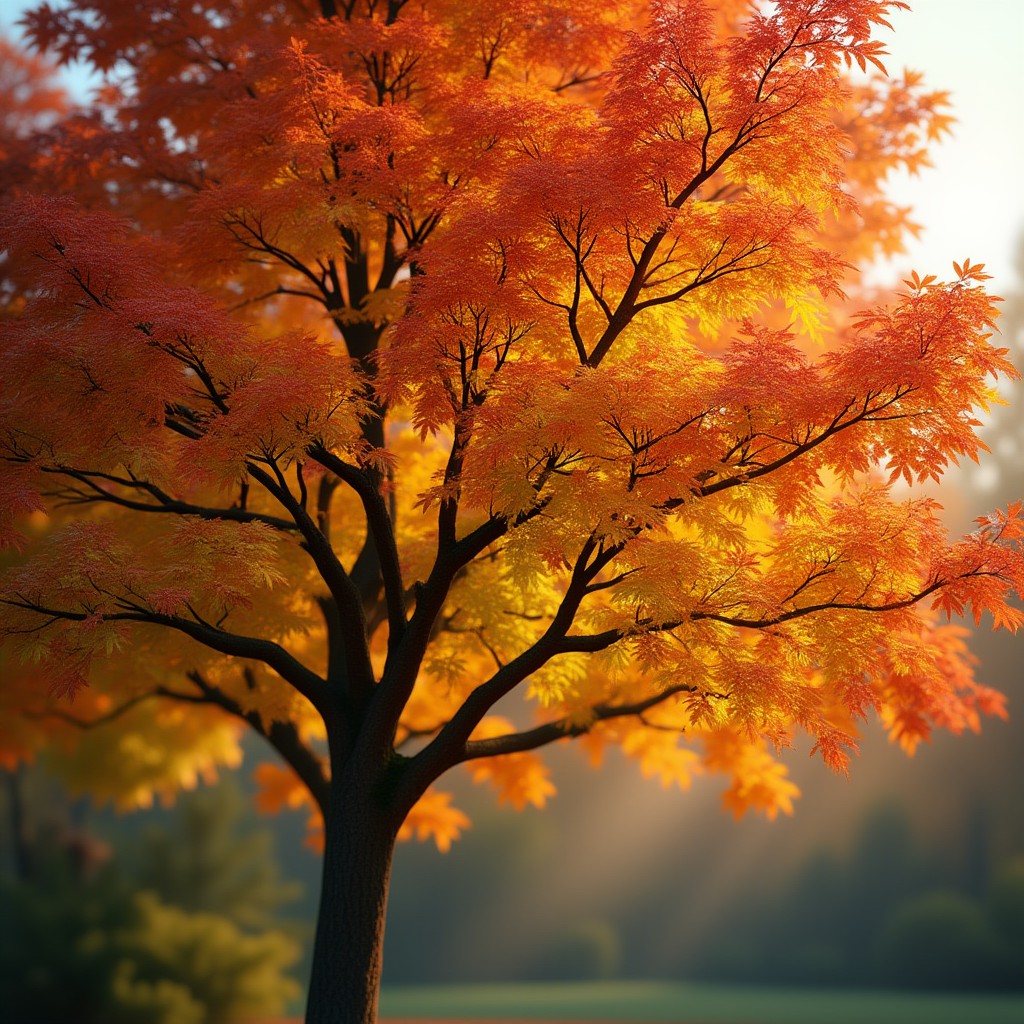}}{\wisepanel{moldy bread}{Biology-2}{\wisebiologytwo}{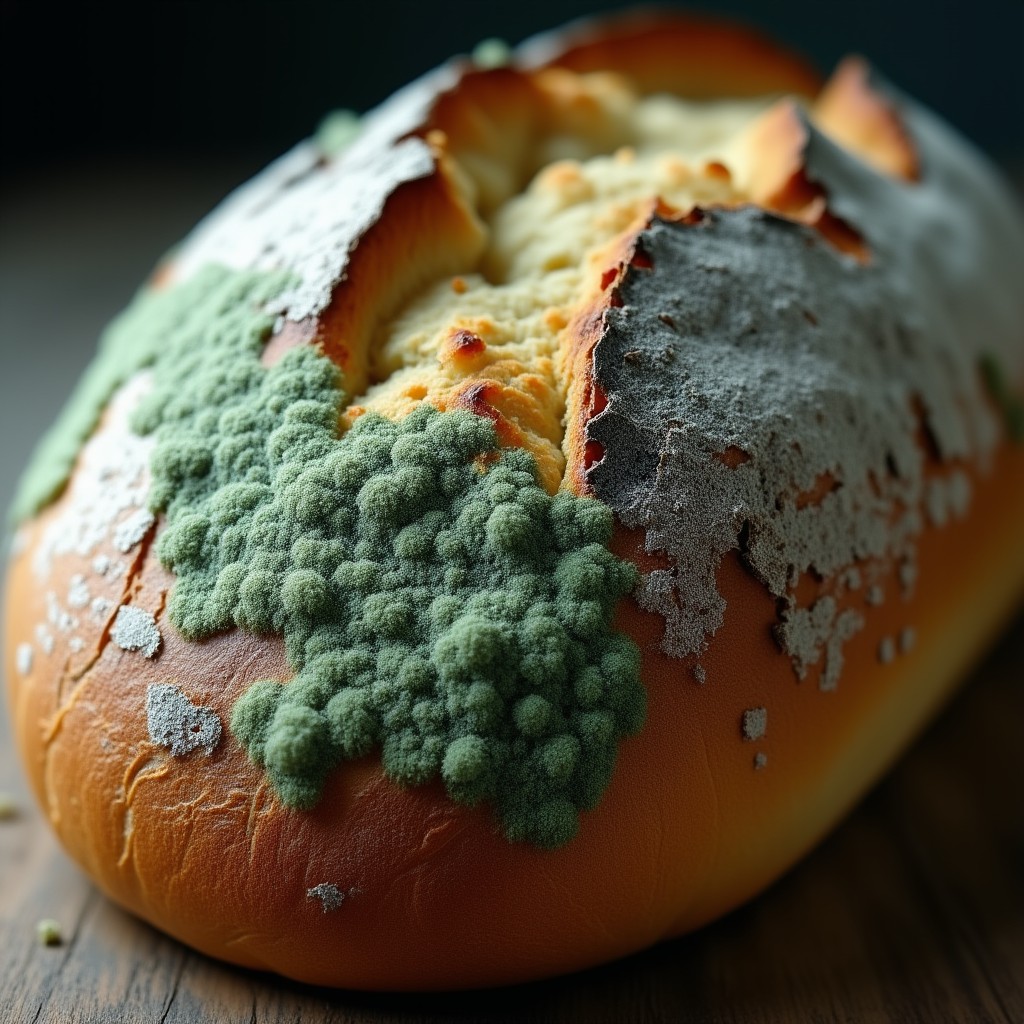}}
\end{minipage}}
\caption{\textbf{Knowledge-rich generation on WISE.} We show image generation examples from STBridge on WISE. Each row corresponds to a knowledge dimension, with two sets, each containing a prompt, a thinking trace, and an image. Blue boxes contain the model's thinking for translating the prompt into a target expression; magenta text highlights key knowledge constraints and visual target cues.}
\label{fig:wise_qualitative_layout}
\endgroup
\end{figure}
\clearpage

\begin{figure}[p]
\centering
\begingroup
\vspace*{-1.16in}
\setlength{\abovecaptionskip}{2pt}
\setlength{\belowcaptionskip}{0pt}
\newcommand{\understandpanelheight}{0.142\textheight}
\newcommand{\understandpromptheight}{0.023\textheight}
\newcommand{\understandpromptfont}{\fontsize{6.8pt}{6.5pt}\selectfont}
\newcommand{\understandtextfont}{\fontsize{5.7pt}{6.05pt}\selectfont}
\newcommand{\understandlabelfont}{\fontsize{5.45pt}{5.65pt}\selectfont}
\definecolor{understandLogoMagenta}{RGB}{223,66,146}
\newcommand{\understandkey}[1]{\textcolor{understandLogoMagenta}{\textbf{#1}}}
\newcommand{\understandentry}[7]{%
\begin{minipage}[t]{0.497\linewidth}
\begin{minipage}[t][\understandpromptheight][c]{\linewidth}
\centering\understandpromptfont\textbf{Question:} \emph{#1}
\end{minipage}\par\vspace{0.8pt}
\begin{minipage}[t]{0.485\linewidth}
\vspace{0pt}
\centering
\includegraphics[width=\linewidth,height=\understandpanelheight,keepaspectratio]{#2}
\end{minipage}\hfill
\begin{minipage}[t]{0.505\linewidth}
\vspace{0pt}
\begin{tcolorbox}[colback=youtuLightBlue,colframe=youtuBlue!25,boxrule=0.25pt,arc=1.2pt,left=1.4pt,right=1.4pt,top=1.1pt,bottom=1.1pt,width=\linewidth,fit to height=\understandpanelheight,fit basedim=5.6pt,fit skip=0.85,fit algorithm=fontsize]
{\raggedright\understandtextfont\setlength{\parindent}{0pt}\setlength{\parskip}{0pt}%
\textbf{Task.} #3\par\vspace{0.8pt}%
\textbf{Options.} #4\par\vspace{0.8pt}%
\textbf{Answer:} \textcolor{youtuBlue}{\textbf{#5}}\quad\textbf{GT:} \understandkey{#6}\par}
\end{tcolorbox}
\end{minipage}
\end{minipage}}
\newcommand{\understandgridentry}[6]{%
\begin{minipage}[t]{0.497\linewidth}
\begin{minipage}[t][\understandpromptheight][c]{\linewidth}
\centering\understandpromptfont\textbf{Task:} \emph{#1}
\end{minipage}\par\vspace{0.8pt}
\begin{minipage}[t]{0.485\linewidth}
\vspace{0pt}
\centering
\includegraphics[width=\linewidth,height=\understandpanelheight,keepaspectratio]{#2}
\end{minipage}\hfill
\begin{minipage}[t]{0.505\linewidth}
\vspace{0pt}
\begin{tcolorbox}[colback=youtuLightBlue,colframe=youtuBlue!25,boxrule=0.25pt,arc=1.2pt,left=1.4pt,right=1.4pt,top=1.1pt,bottom=1.1pt,width=\linewidth,fit to height=\understandpanelheight,fit basedim=5.6pt,fit skip=0.85,fit algorithm=fontsize]
{\raggedright\understandtextfont\setlength{\parindent}{0pt}\setlength{\parskip}{0pt}%
\textbf{Grid.} #3\par\vspace{0.8pt}%
\textbf{Oddity.} #4\par\vspace{0.8pt}%
\textbf{Answer:} \textcolor{youtuBlue}{\textbf{#5}}\par\textbf{GT:} \understandkey{#6}\par}
\end{tcolorbox}
\end{minipage}
\end{minipage}}
\makebox[\textwidth][c]{%
\begin{minipage}[t]{\textwidth}
\noindent
\understandentry
{Which object is closer to the camera?}
{}
{Compare the table highlighted by a red box with the bookcase highlighted by a blue box.}
{A. table; B. bookcase}
{A}
{A}
{}
\hfill
\understandentry
{How many pendant lamps are in the image?}
{}
{Count the pendant lamps visible in the scene.}
{A. 3; B. 5; C. 4.0; D. 0.0; E. 2.0; F. 6.0}
{B}
{B}
{}
\par\vspace{7.0pt}
\noindent
\understandentry
{Which object is closer to the table?}
{}
{Estimate distances from the red-box table to the blue-box lamp and green-box television.}
{A. lamp; B. television}
{A}
{A}
{}
\hfill
\understandentry
{Which point corresponds to the reference point?}
{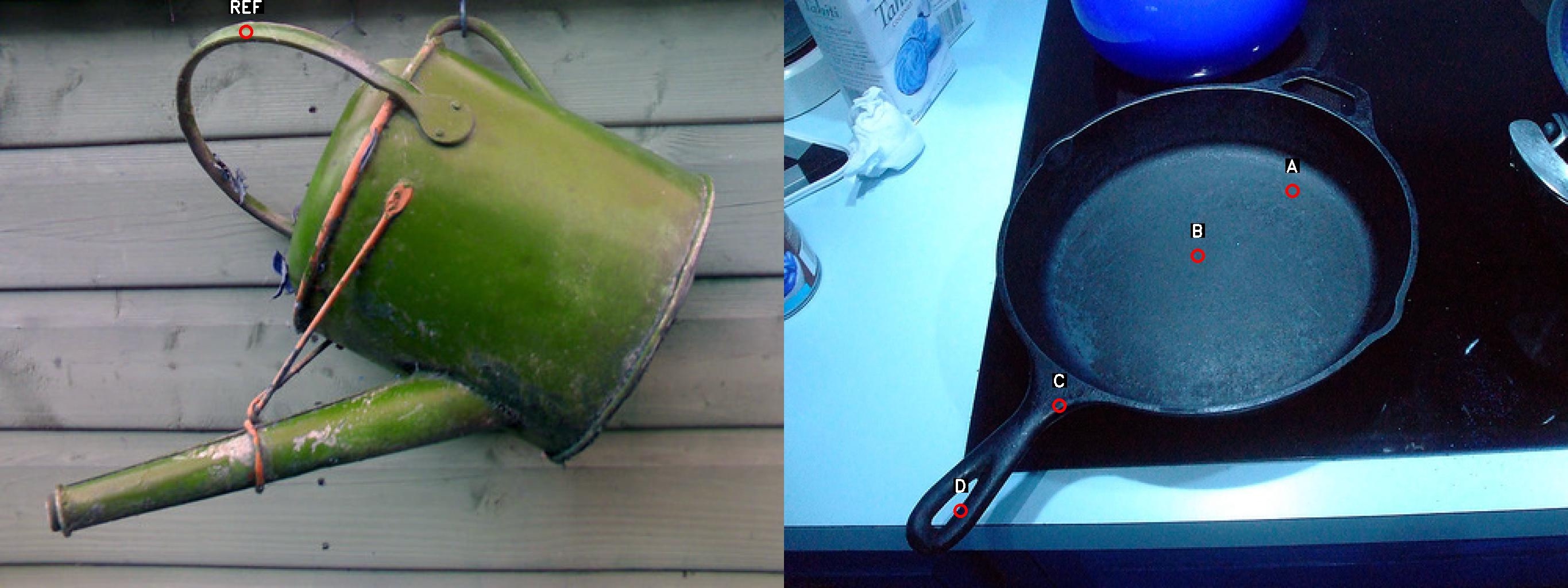}
{Identify the point matching the reference location.}
{A. Point A; B. Point B; C. Point C; D. Point D}
{(D)}
{(D)}
{}
\par\vspace{7.0pt}
\noindent
\understandentry
{Which image is most similar to the reference image?}
{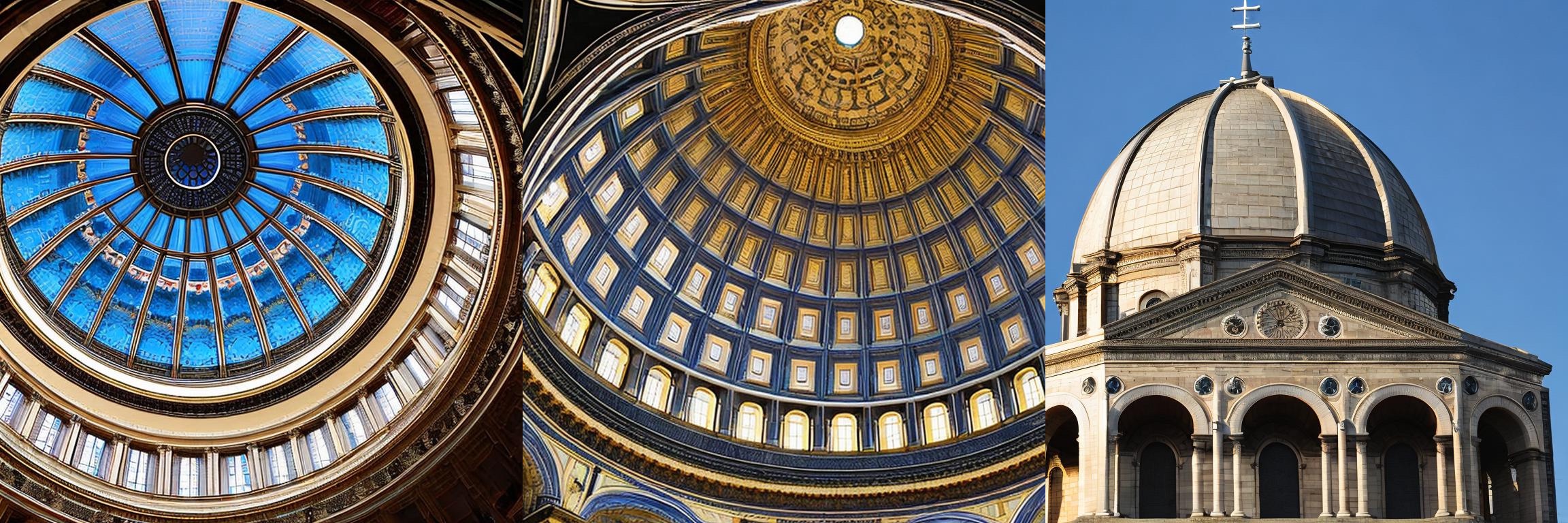}
{Select the candidate image most similar to the reference.}
{A. the second image; B. the third image}
{(A)}
{(A)}
{}
\hfill
\understandgridentry
{Find the odd gamepad in the grid.}
{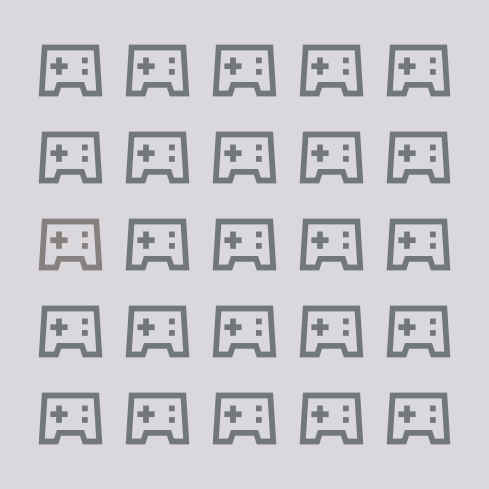}
{5$\times$5 gamepad grid}
{color}
{Row 3, Column 1}
{Row 3, Column 1}
\par\vspace{7.0pt}
\noindent
\understandentry
{Which box more accurately encloses the cake?}
{}
{Compare two candidate boxes for localizing the cake.}
{A. Box A; B. Box B}
{(A)}
{(A)}
{}
\hfill
\understandgridentry
{Find the odd word-shape item in the grid.}
{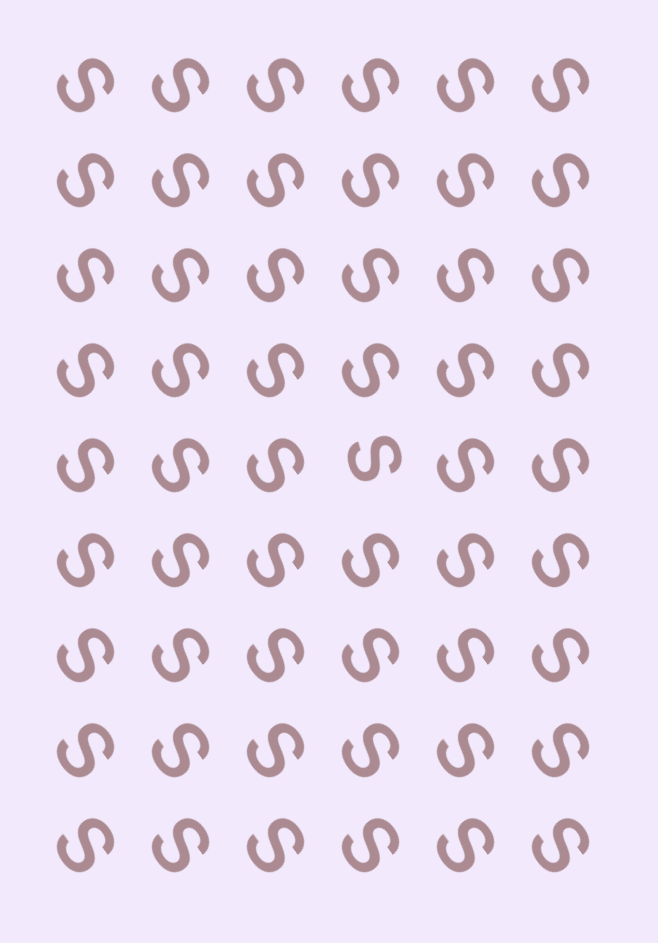}
{9$\times$6 word-s grid}
{size, position, rotation}
{Row 5, Column 4}
{Row 5, Column 4}
\end{minipage}}
\caption{\textbf{Image-understanding examples.} Representative image-understanding examples from STBridge are arranged in a compact two-column qualitative grid. Each panel pairs the visual input on the left with a compact blue question-answer block on the right. Blue text marks the model answer, and magenta text marks the ground-truth answer.}
\label{fig:understand_qualitative_layout}
\endgroup
\end{figure}
\clearpage

\begin{figure}[p]
\centering
\begingroup
\vspace*{-0.73in}
\setlength{\abovecaptionskip}{2pt}
\setlength{\belowcaptionskip}{0pt}
\newcommand{\risepromptheight}{0.020\textheight}
\newcommand{\risethinkheight}{0.041\textheight}
\newcommand{\riseimageheight}{0.122\textheight}
\newcommand{\risepromptfont}{\fontsize{7pt}{6.2pt}\selectfont}
\newcommand{\risethinkfont}{\fontsize{6.4pt}{6.7pt}\selectfont}
\newcommand{\riselabelfont}{\fontsize{5.8pt}{5.8pt}\selectfont}
\definecolor{riseLogoMagenta}{RGB}{223,66,146}
\newcommand{\risekey}[1]{\textcolor{riseLogoMagenta}{\textbf{#1}}}
\newcommand{\riseentry}[4]{%
\begin{minipage}[t]{0.497\linewidth}
\begin{minipage}[t]{\linewidth}
\centering\risepromptfont\emph{#1}
\end{minipage}\par\vspace{-1.2pt}
\begin{tcolorbox}[colback=youtuLightBlue,colframe=youtuBlue!25,boxrule=0.25pt,arc=1.2pt,left=1.4pt,right=1.4pt,top=1.0pt,bottom=1.0pt,width=\linewidth,height=\risethinkheight,valign=top,before skip=0pt,after skip=0pt]
{\risethinkfont\raggedright\setlength{\parindent}{0pt}\setlength{\parskip}{0pt}\textbf{Thinking.} #2\par}
\end{tcolorbox}\vspace{1.5pt}
\begin{minipage}[t]{0.493\linewidth}
\centering{\riselabelfont\textbf{Source}}\par\vspace{-0.5pt}
\includegraphics[width=\linewidth,height=\riseimageheight,keepaspectratio]{#3}
\end{minipage}\hfill
\begin{minipage}[t]{0.493\linewidth}
\centering{\riselabelfont\textbf{\textcolor{youtuBlue}{Result}}}\par\vspace{-0.5pt}
\includegraphics[width=\linewidth,height=\riseimageheight,keepaspectratio]{#4}
\end{minipage}
\end{minipage}}
\makebox[\textwidth][c]{%
\begin{minipage}[t]{1.14\textwidth}
\noindent
\riseentry
{Draw what it will look like ten years later.}
{Transform the young sapling into a \risekey{mature tree} with a \risekey{thick trunk} and \risekey{lush canopy}, while preserving the \risekey{natural outdoor lighting} and growth setting.}
{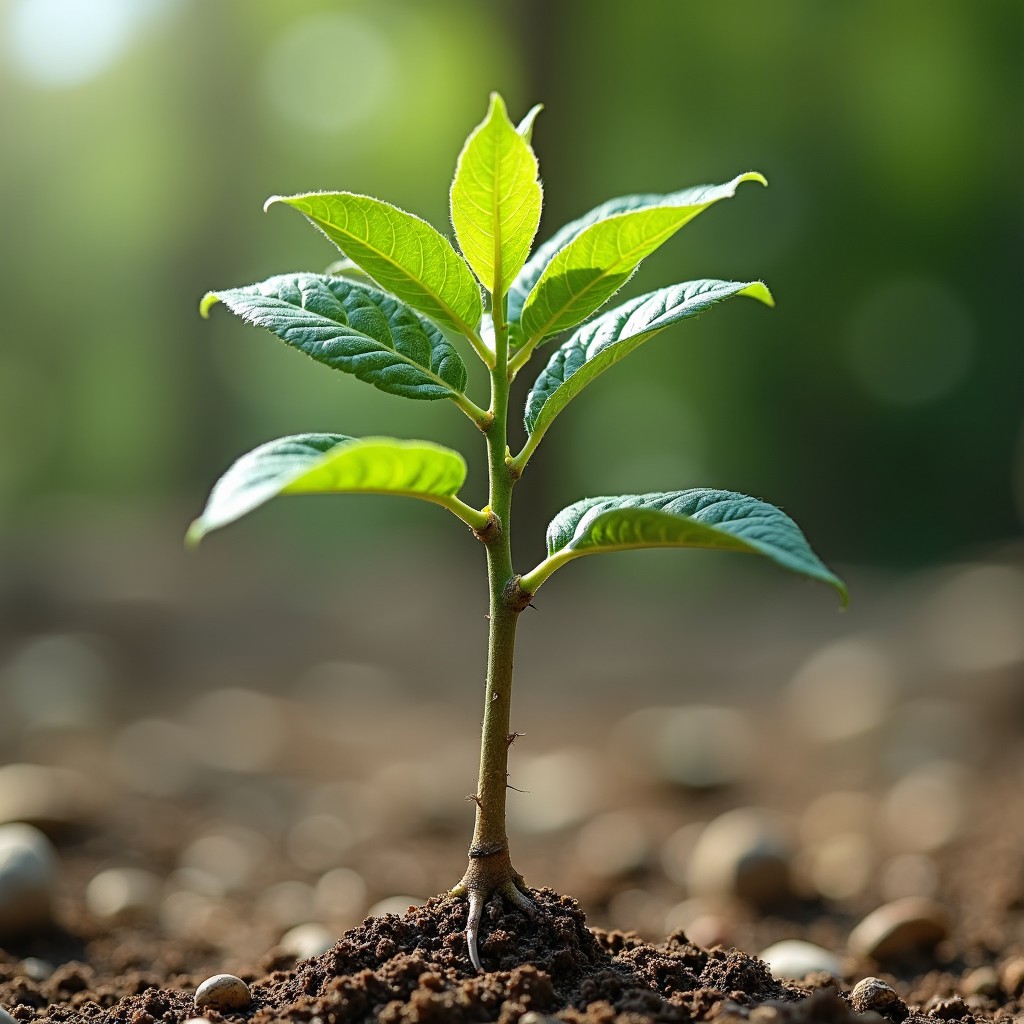}
{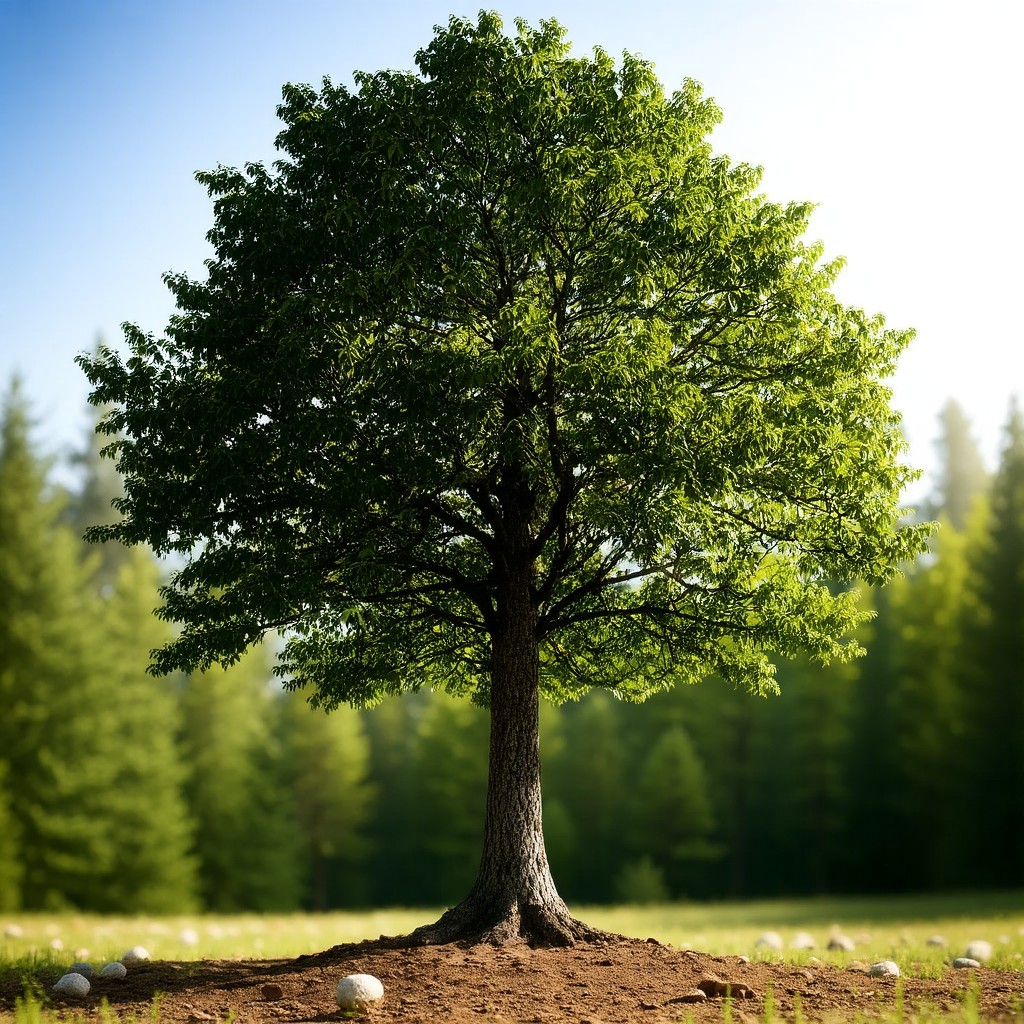}
\hfill
\riseentry
{Draw what it will look like after six hours outdoors on a snowy day.}
{Cover the bench, grass, and branches with a \risekey{thick layer of snow}; keep the \risekey{bare trees}, \risekey{diffuse winter lighting}, and quiet park composition.}
{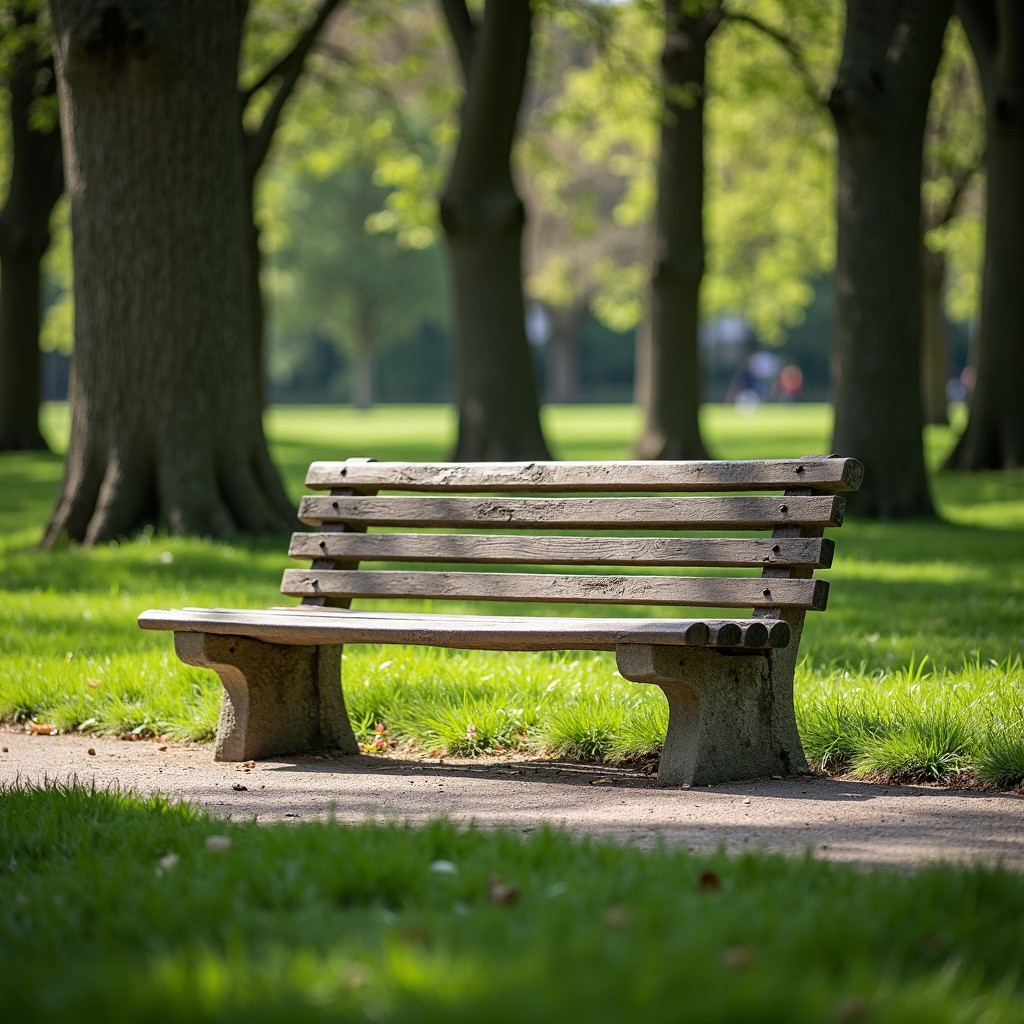}
{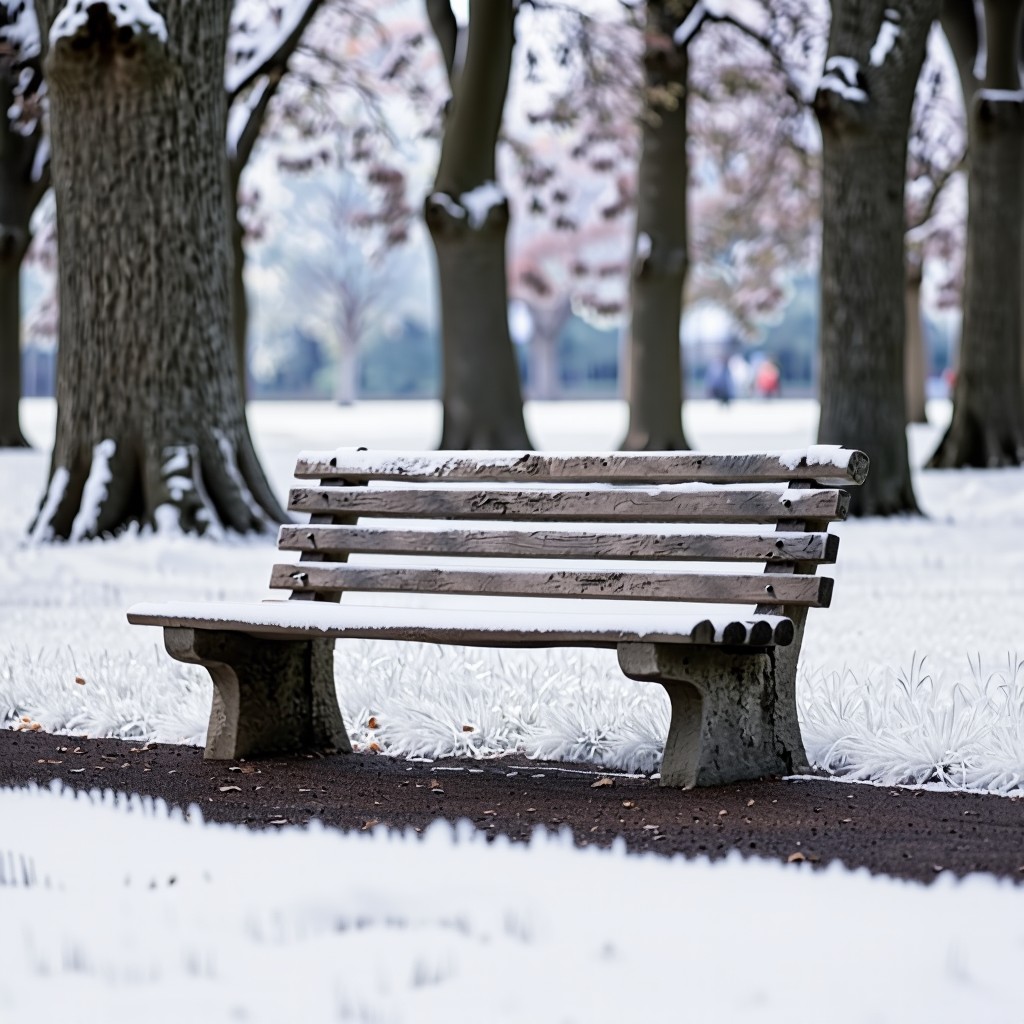}
\par\vspace{8.0pt}
\noindent
\riseentry
{Draw what it will look like after it has been painted red.}
{Keep the bicycle \risekey{geometry} and studio composition unchanged, but repaint the frame a \risekey{uniform glossy red} while preserving \risekey{black components}.}
{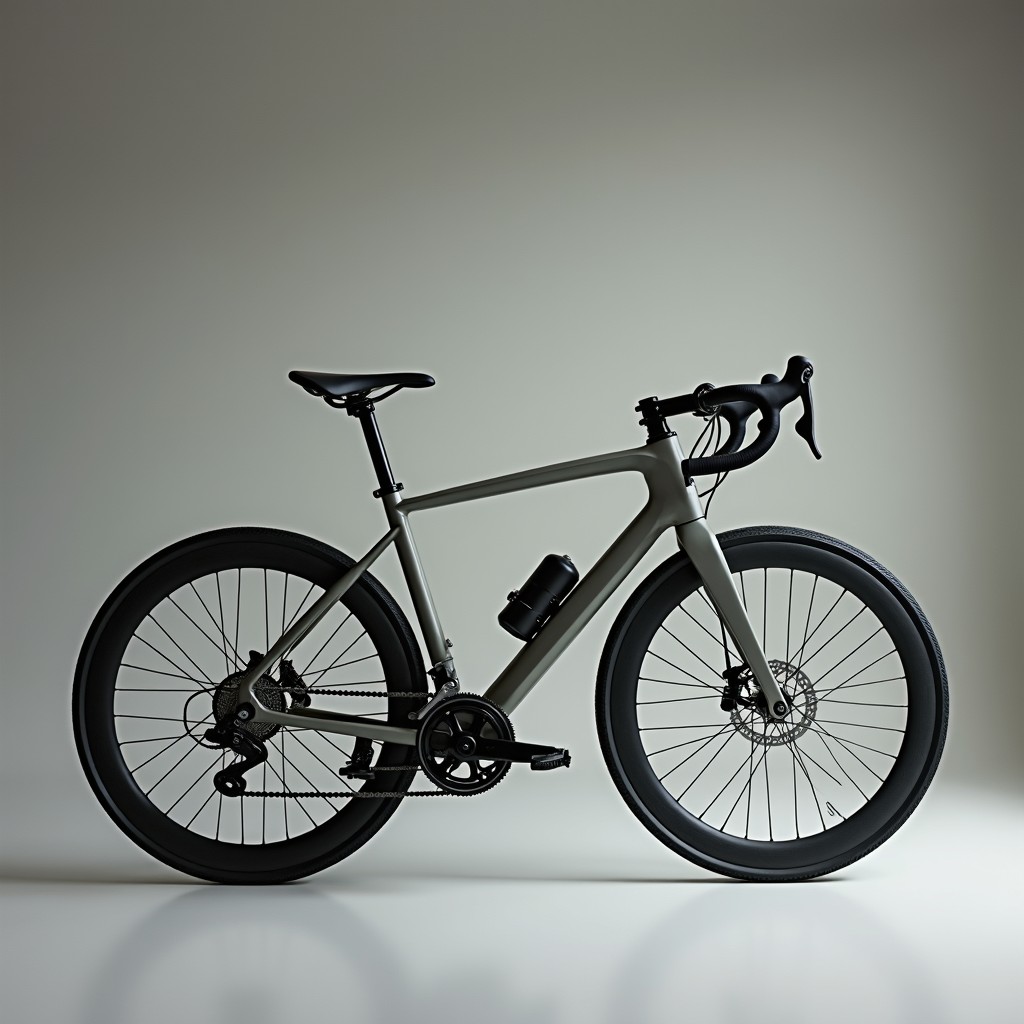}
{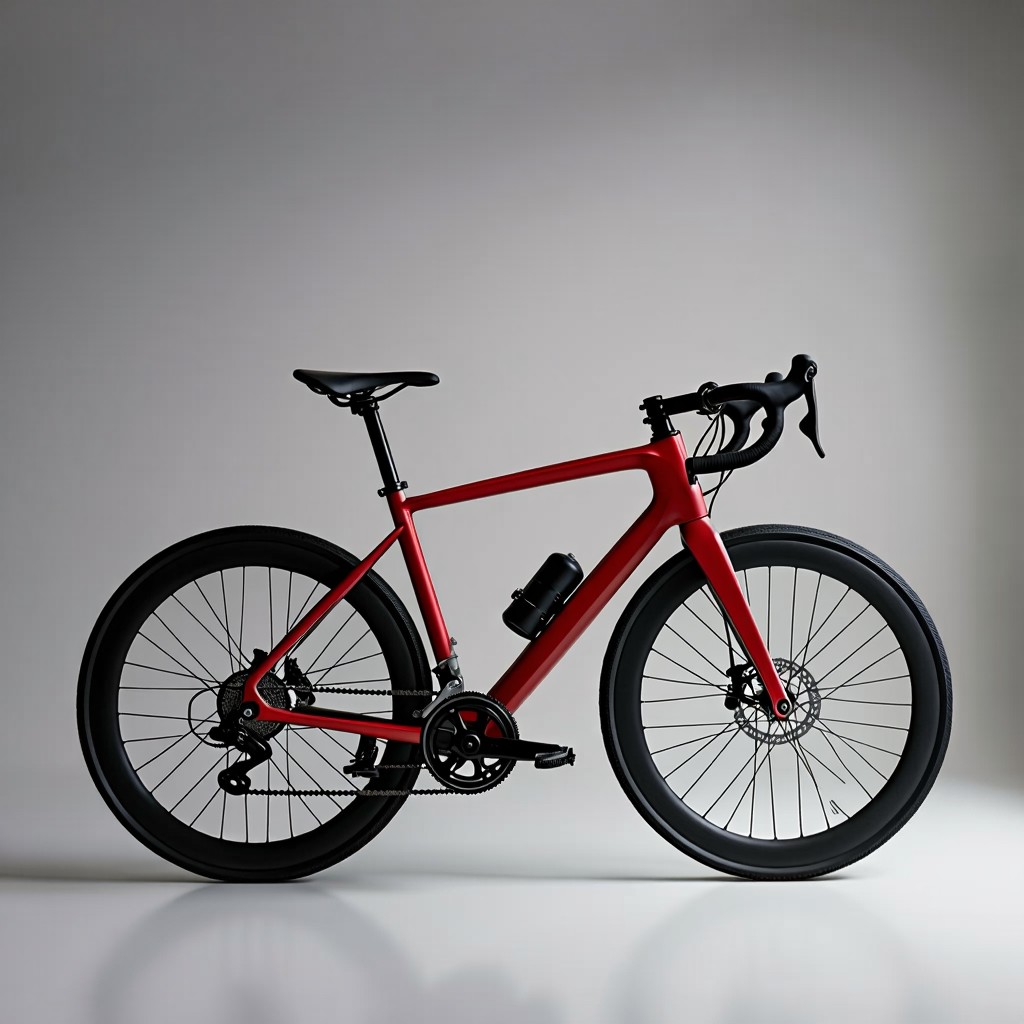}
\hfill
\riseentry
{Draw what it will look like after it gets stuck in mud.}
{Preserve the \risekey{sneaker silhouette} and studio viewpoint, then add \risekey{realistic brown mud} across the \risekey{sole}, \risekey{lateral panels}, and \risekey{laces}.}
{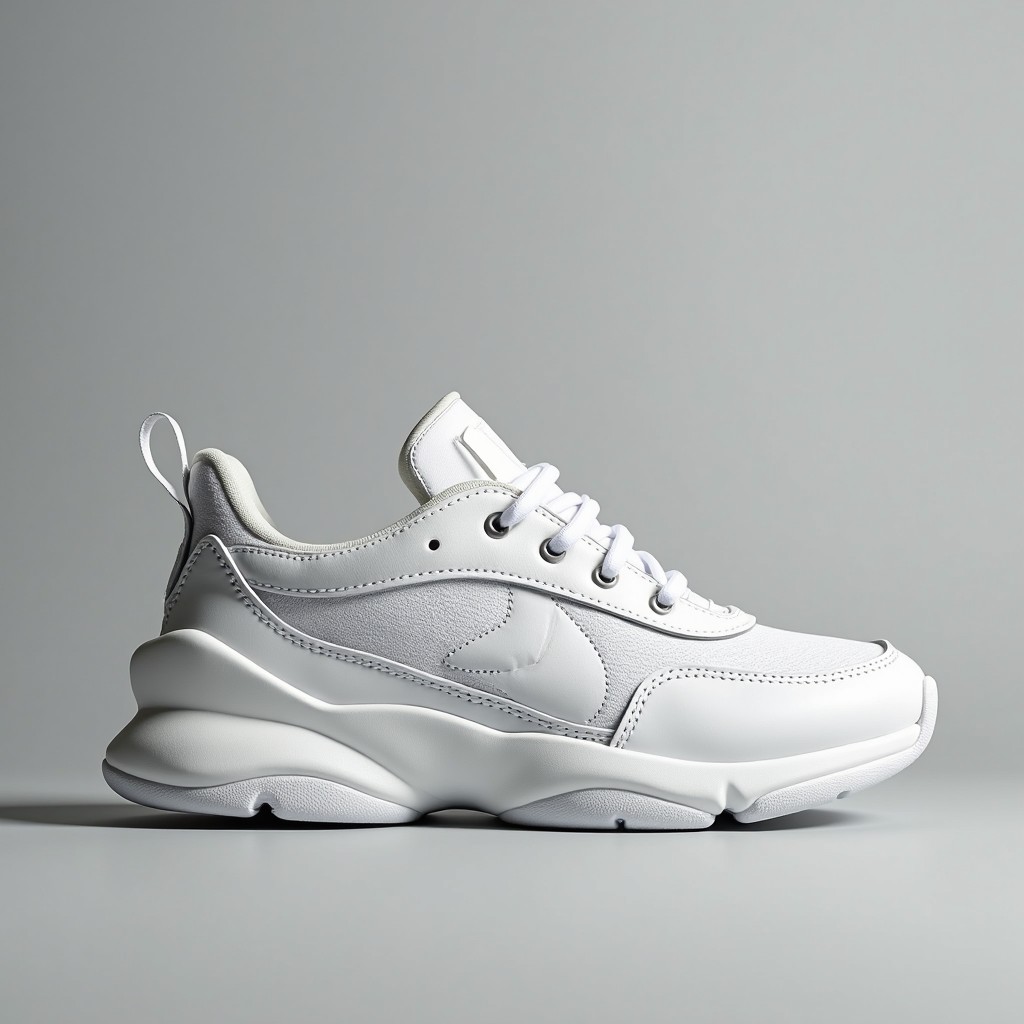}
{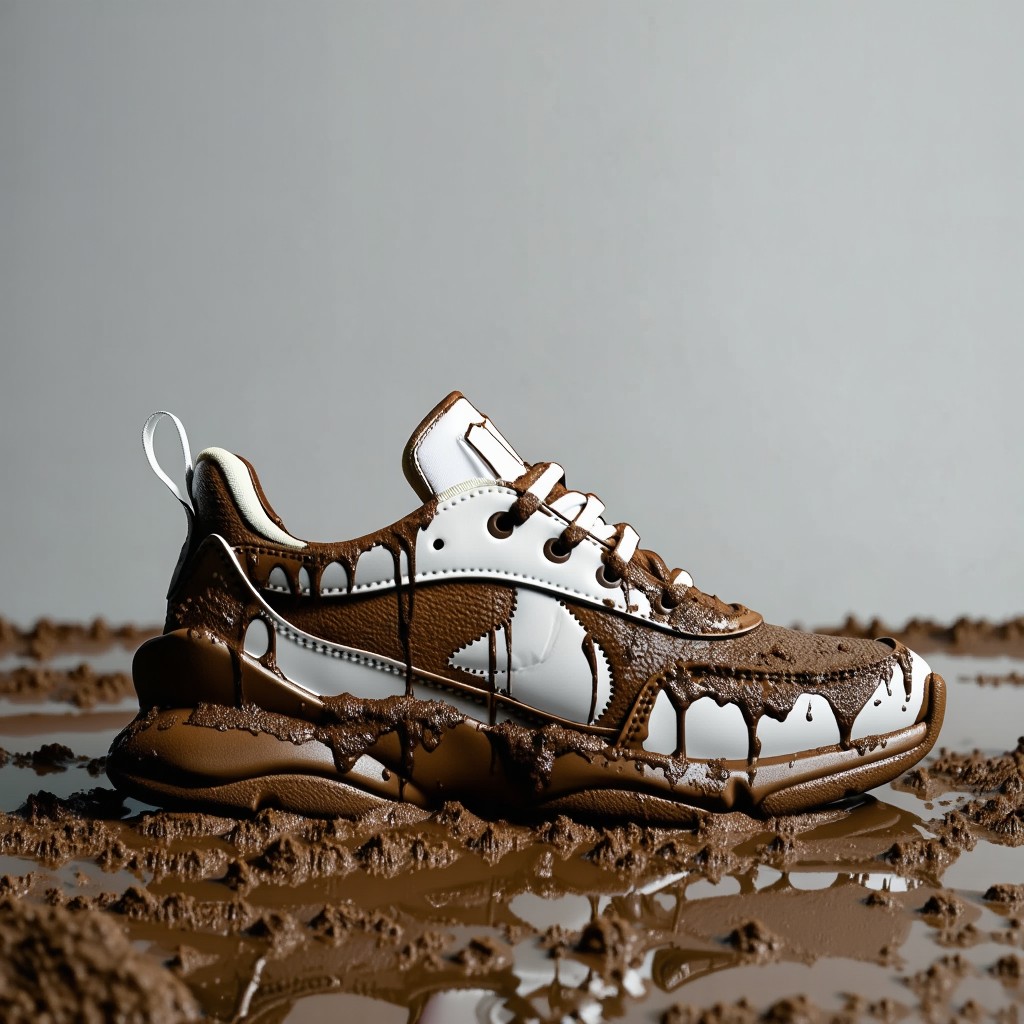}
\par\vspace{8.0pt}
\noindent
\riseentry
{Draw what it will look like after being scribbled on by a child.}
{Add \risekey{colorful child-like scribbles} across the wall while keeping the \risekey{plain room}, \risekey{clean floor}, and centered viewpoint unchanged.}
{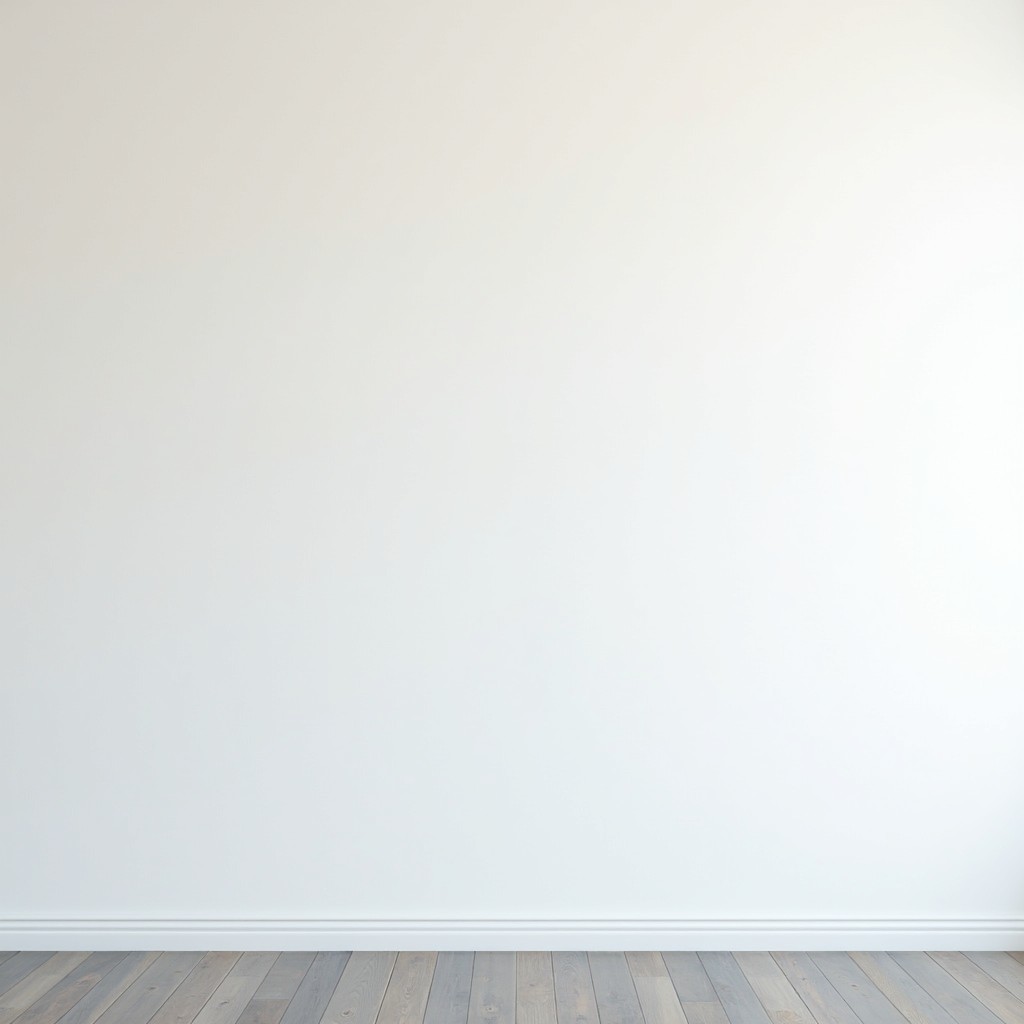}
{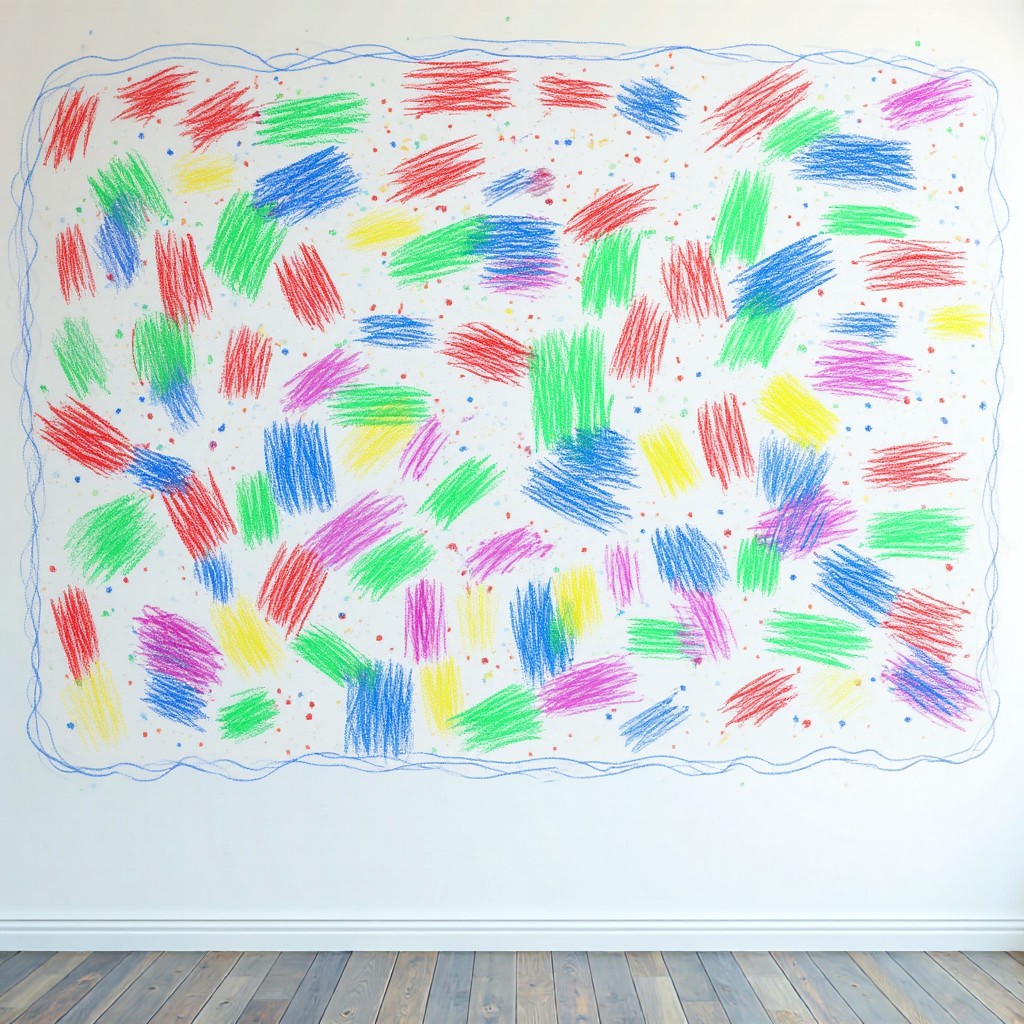}
\hfill
\riseentry
{Draw what it will look like after being turned on.}
{Turn on the lamp by adding a \risekey{warm visible glow} from the shade, with \risekey{nearby illumination} while preserving the interior layout.}
{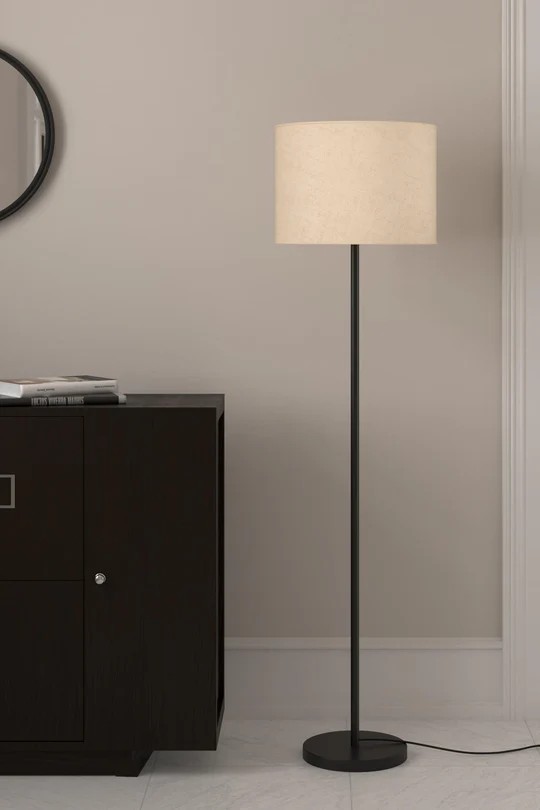}
{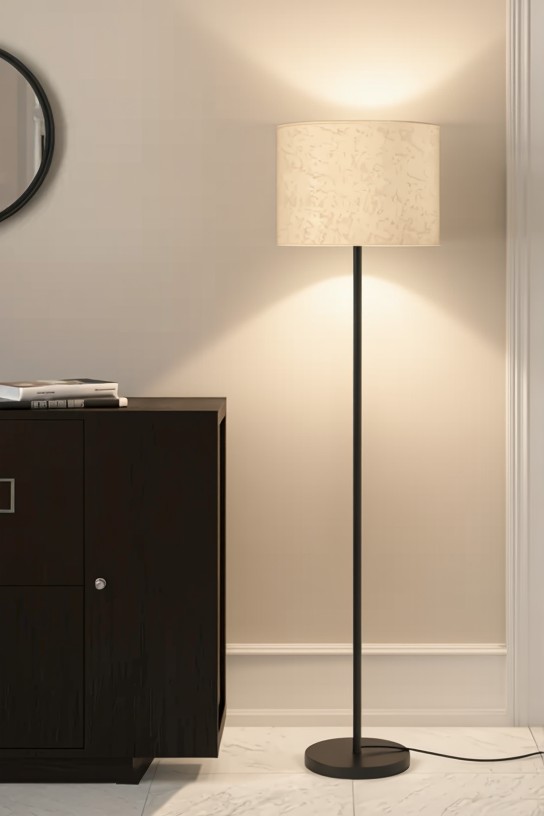}
\par\vspace{8.0pt}
\noindent
\riseentry
{Draw an image showing the front view of the ladder.}
{Infer the \risekey{front-facing view}: center the \risekey{two vertical rails}, make the \risekey{horizontal rungs evenly spaced}, and keep a clean background.}
{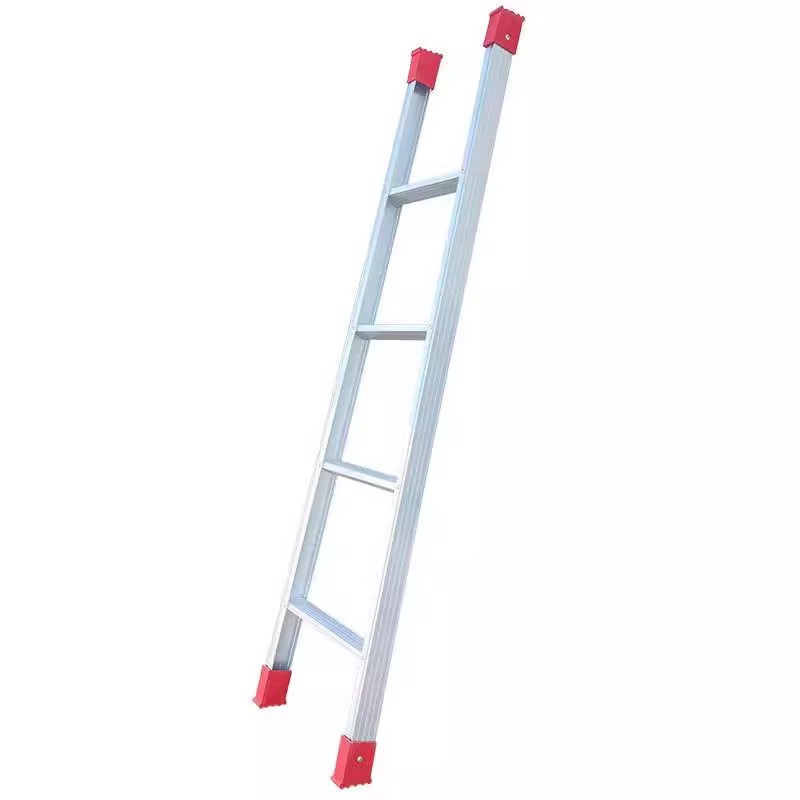}
{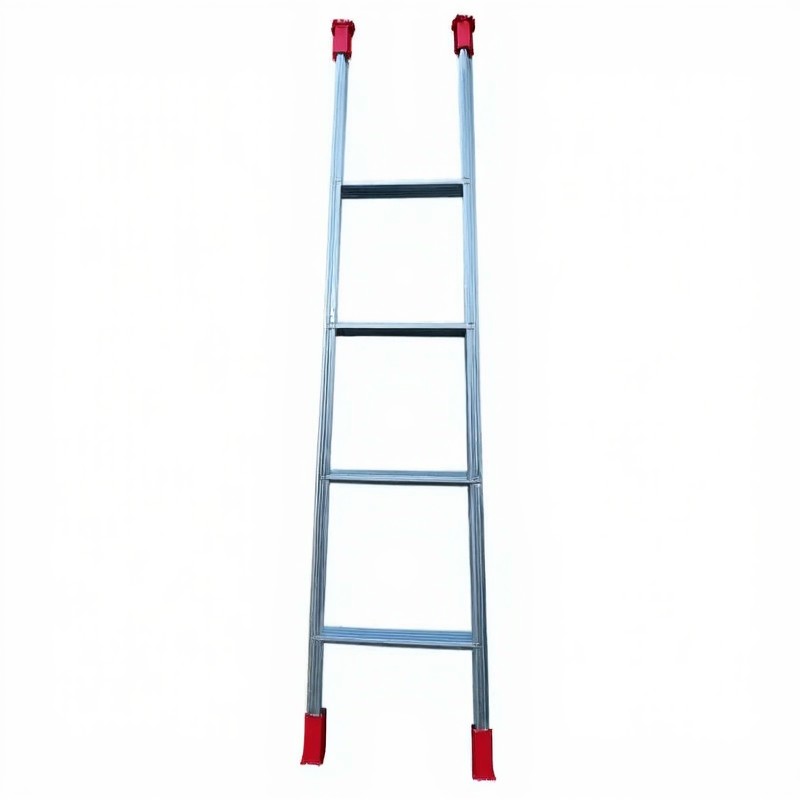}
\hfill
\riseentry
{Draw an image showing the provided incomplete shape completed into a whole circle.}
{Complete the missing sector into a \risekey{whole circle} with a \risekey{smooth continuous edge}, preserving the \risekey{light blue fill} and clean background.}
{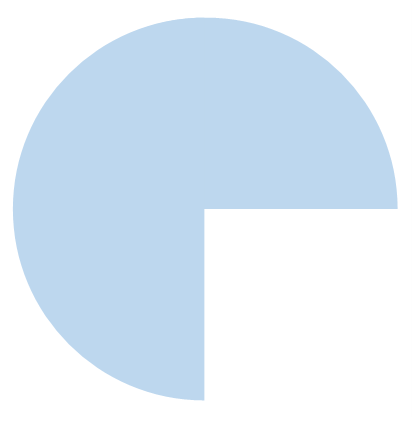}
{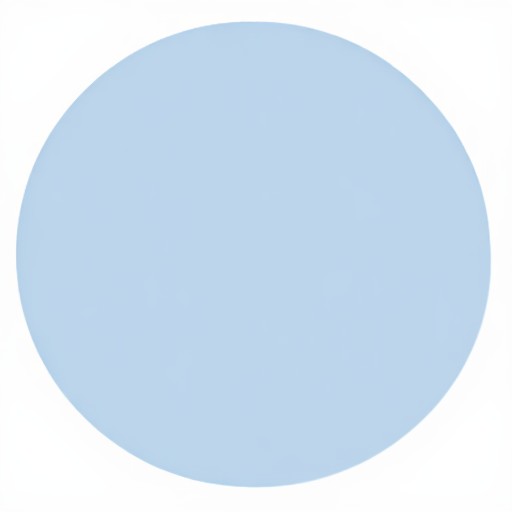}
\end{minipage}}
\caption{\textbf{RISE editing examples.} Representative STBridge RISE image-editing examples pair the source image with the edited result and show an abbreviated thinking summary with key visual changes or preservation cues highlighted in magenta.}
\label{fig:rise_qualitative_layout}
\endgroup
\end{figure}

\clearpage

\section{Introduction}

Unified multimodal models (UMMs)~\cite{Wang_2022_OFA,Lu_2023_UnifiedIO2,Dong_2023_DreamLLM,Chameleon_2024,Xie_2024_Showo,Wang_2024_Emu3,Wu_2024_Janus,deng2025emerging} aim to integrate visual understanding and visual generation within a single model. They support a broad range of tasks, including image question answering, visual reasoning, object grounding, image generation, and image editing~\cite{Ge_2024_SEEDX,Wu_2024_VILAU,Chen_2025_BLIP3o,Tian_2026_InternVLU,Liu_2025_TUNA,Liu_2026_Tuna2,Diao_2026_SenseNovaU1}. Recent UMMs have achieved strong performance across both understanding and generation benchmarks, suggesting that perception and synthesis can share a unified architecture and representation space. This progress makes UMMs a promising direction toward general-purpose visual intelligence.

However, architectural unification does not necessarily lead to semantic unification. A truly unified multimodal model should not only solve understanding and generation tasks separately, but also coordinate the two capabilities around the same user intent. In particular, the textual output produced by the understanding pathway should describe the visual result that the generation pathway is expected to realize, while the generated image should faithfully reflect the semantics expressed by understanding. When these two outputs diverge, the model may appear unified at the architectural level, yet remain misaligned at the level of target semantics.

\begin{figure}[!t]
\centering
\includegraphics[width=0.92\textwidth]{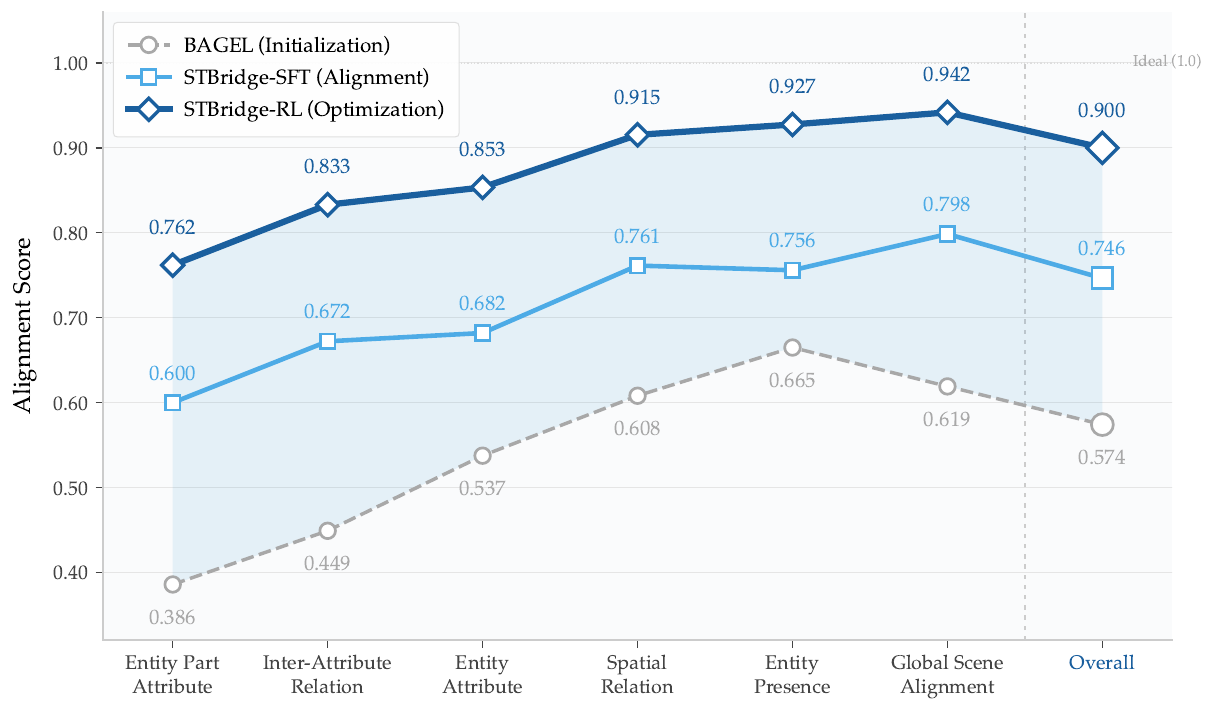}
\caption{\textbf{Understanding-generation alignment analysis.}
We evaluate the consistency between the understanding target description and the generation edited image produced by BAGEL across seven alignment metrics. BAGEL scores remain low across all fine-grained dimensions, revealing a substantial alignment gap between what the model describes and what it generates.}
\label{fig:teaser_alignment}
\end{figure}

We examine this problem through image editing, where a user instruction defines a target visual state that can be both described in language and realized as an image. Given the same source image and edit instruction, we ask the understanding pathway to predict a textual description of the edited target and the generation pathway to produce the corresponding edited image. We then design a VLM-based evaluation protocol to assess whether the two outputs converge to the same target result. The protocol measures consistency at both the global scene level and several fine-grained dimensions, including entities, attributes, spatial relations, entity presence, and local details, as shown in Fig.~\ref{fig:teaser_alignment}. Applying this protocol to 400 diverse editing cases, we find that BAGEL exhibits limited alignment across all fine-grained dimensions. These results indicate that current UMMs can preserve coarse semantic intent while still failing to coordinate detailed information between what they describe and what they generate.

\begin{figure}[!t]
\centering
\begingroup
\setlength{\abovecaptionskip}{3pt}
\setlength{\belowcaptionskip}{0pt}
\newcommand{\alignheadfont}{\fontsize{7.2pt}{7.4pt}\selectfont}
\newcommand{\aligncasefont}{\fontsize{7.1pt}{7.4pt}\selectfont}
\newcommand{\aligninstfont}{\fontsize{6.5pt}{6.8pt}\selectfont}
\newcommand{\aligntextfont}{\fontsize{7.0pt}{7.25pt}\selectfont}
\newcommand{\alignimageheight}{0.143\textheight}
\newcommand{\alignrowheight}{0.149\textheight}
\definecolor{alignGood}{RGB}{38,132,82}
\definecolor{alignBad}{RGB}{190,72,72}
\definecolor{alignGap}{RGB}{172,117,24}
\newcommand{\alignsmile}{%
\tikz[baseline=-0.46ex,x=1.30em,y=1.30em,line width=0.060em]{%
\draw[alignGood] (0,0) circle (0.46);
\fill[alignGood] (-0.15,0.12) circle (0.045);
\fill[alignGood] (0.15,0.12) circle (0.045);
\draw[alignGood] (-0.20,-0.08) .. controls (0,-0.25) .. (0.20,-0.08);
}}
\newcommand{\alignsad}{%
\tikz[baseline=-0.46ex,x=1.30em,y=1.30em,line width=0.060em]{%
\draw[alignBad] (0,0) circle (0.46);
\fill[alignBad] (-0.15,0.12) circle (0.045);
\fill[alignBad] (0.15,0.12) circle (0.045);
\draw[alignBad] (-0.20,-0.21) .. controls (0,-0.04) .. (0.20,-0.21);
}}
\newcommand{\aligngapface}{%
\tikz[baseline=-0.46ex,x=1.30em,y=1.30em,line width=0.060em]{%
\draw[alignGap] (0,0) circle (0.46);
\fill[alignGap] (-0.15,0.12) circle (0.045);
\fill[alignGap] (0.15,0.12) circle (0.045);
\draw[alignGap] (-0.22,-0.14) -- (-0.06,-0.14);
\draw[alignGap] (0.07,-0.14) -- (0.22,-0.14);
}}
\newcommand{\alignheader}[1]{\centering{\alignheadfont\textbf{\textcolor{youtuBlue}{#1}}}}
\newcommand{\aligninstruction}[1]{%
{\aligninstfont\noindent\textbf{\textcolor{youtuBlue}{Edit:}} #1\par}
\vspace{-2.8pt}}
\newcommand{\aligncase}[3]{%
\begin{minipage}[c][\alignrowheight][c]{\linewidth}
\centering
{\aligncasefont\textbf{\textcolor{black!78}{#1}}\par\vspace{2.2pt}
\textcolor{black!66}{Gen. #2}\par\vspace{1.6pt}
\textcolor{black!66}{Und. #3}\par}
\end{minipage}}
\newcommand{\alignimagebox}[1]{%
\begin{tcolorbox}[colback=white,colframe=youtuBlue!18,boxrule=0.25pt,arc=1.0pt,left=0.8pt,right=0.8pt,top=0.8pt,bottom=0.8pt,width=\linewidth,height=\alignrowheight,valign=center,before skip=0pt,after skip=0pt]
\centering\includegraphics[width=\linewidth,height=\alignimageheight,keepaspectratio]{#1}
\end{tcolorbox}}
\newcommand{\aligntextbox}[1]{%
\noindent\hspace*{0.10\linewidth}%
\begin{tcolorbox}[colback=youtuLightBlue!30,colframe=youtuBlue!16,boxrule=0.25pt,arc=1.0pt,left=2.0pt,right=2.0pt,top=1.7pt,bottom=1.7pt,width=0.80\linewidth,height=\alignrowheight,valign=top,before skip=0pt,after skip=0pt]
{\raggedright\aligntextfont\setlength{\parindent}{0pt}\setlength{\parskip}{0pt}#1\par}
\end{tcolorbox}}
\newcommand{\alignrow}[6]{%
\begin{minipage}[t]{0.255\linewidth}
\vspace{0pt}
\alignimagebox{#1}
\end{minipage}\hfill
\begin{minipage}[t]{0.255\linewidth}
\vspace{0pt}
\alignimagebox{#2}
\end{minipage}\hfill
\begin{minipage}[t]{0.300\linewidth}
\vspace{0pt}
\aligntextbox{#3}
\end{minipage}\hfill
\begin{minipage}[t]{0.145\linewidth}
\vspace{0pt}
\aligncase{#4}{#5}{#6}
\end{minipage}\par}
\newcommand{\alignentry}[7]{%
\aligninstruction{#1}
\alignrow{#2}{#3}{#4}{#5}{#6}{#7}}
\newcommand{\alignentrysep}[7]{%
\par\vspace{6.2pt}
\alignentry{#1}{#2}{#3}{#4}{#5}{#6}{#7}}
\makebox[\textwidth][c]{%
\begin{minipage}{\textwidth}
\noindent
\begin{minipage}[c]{0.255\linewidth}\alignheader{Source Image}\end{minipage}\hfill
\begin{minipage}[c]{0.255\linewidth}\alignheader{Generation Output}\end{minipage}\hfill
\begin{minipage}[c]{0.300\linewidth}\alignheader{Understanding Output}\end{minipage}\hfill
\begin{minipage}[c]{0.145\linewidth}\alignheader{Diagnosis}\end{minipage}
\par\vspace{1.1pt}
\alignentry
{Add a small vintage bicycle leaning against the wooden panel wall near the lamp.}
{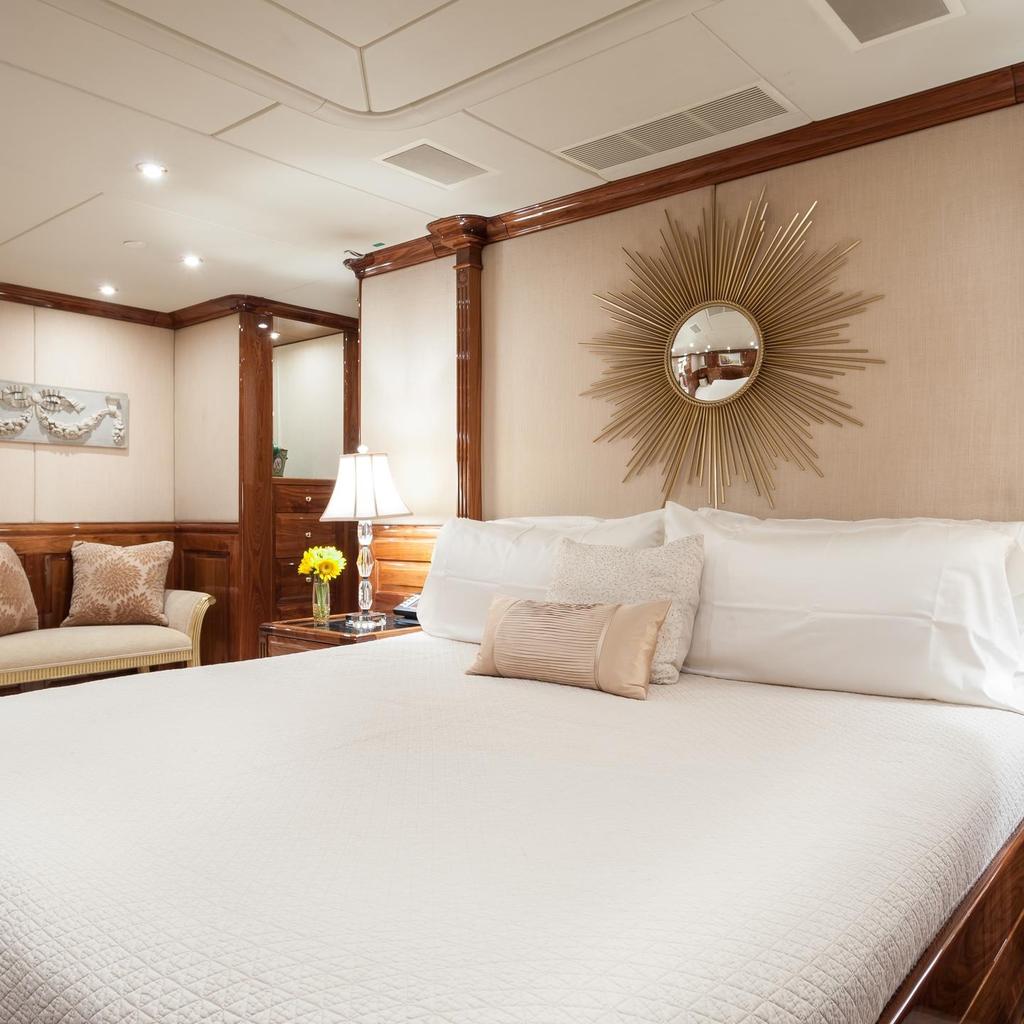}
{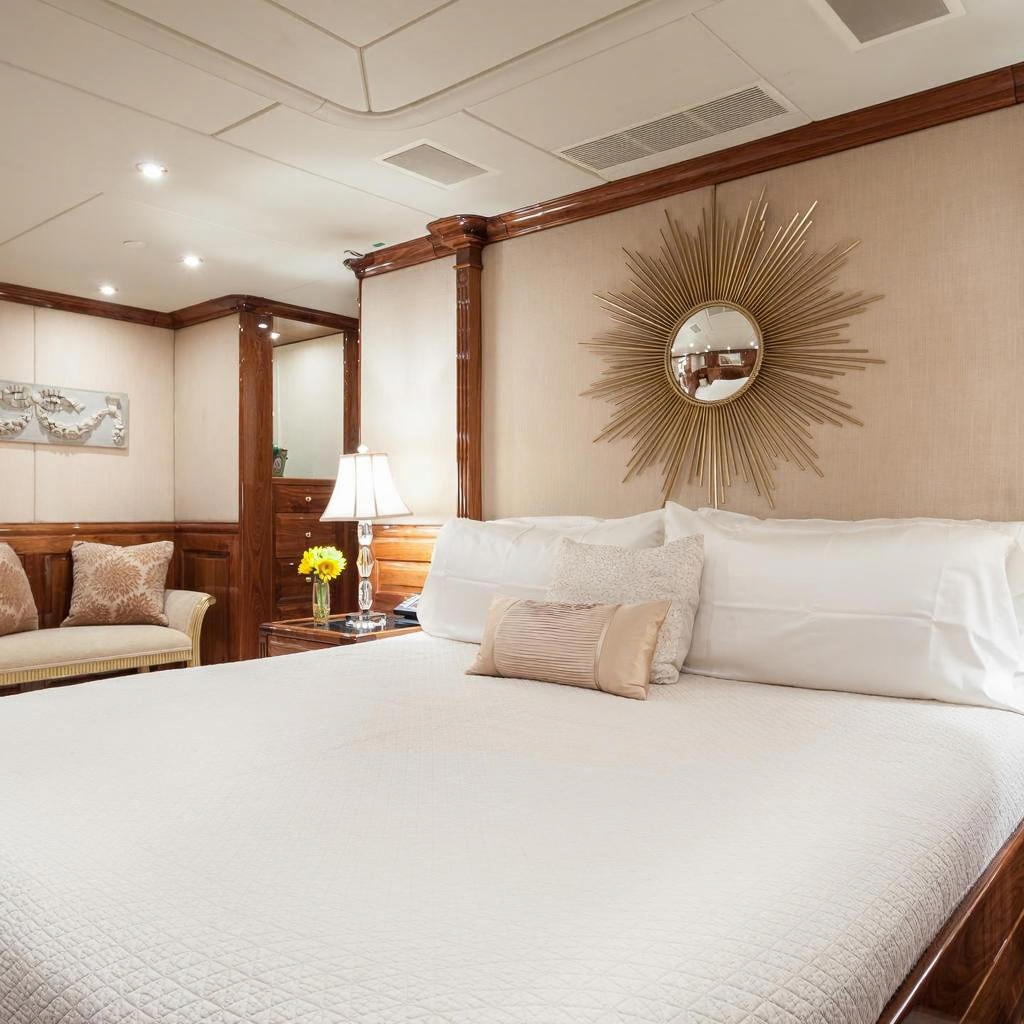}
{A large bed with white sheets and pillows is in a room with a wooden panel wall. There is a lamp on a table next to the bed. There is a small vintage bicycle leaning against the wooden panel wall near the lamp.}
{(a) Generation miss}
{\alignsad}
{\alignsmile}
\alignentrysep
{Transfer the image into a loose, flowing watercolor-wash style.}
{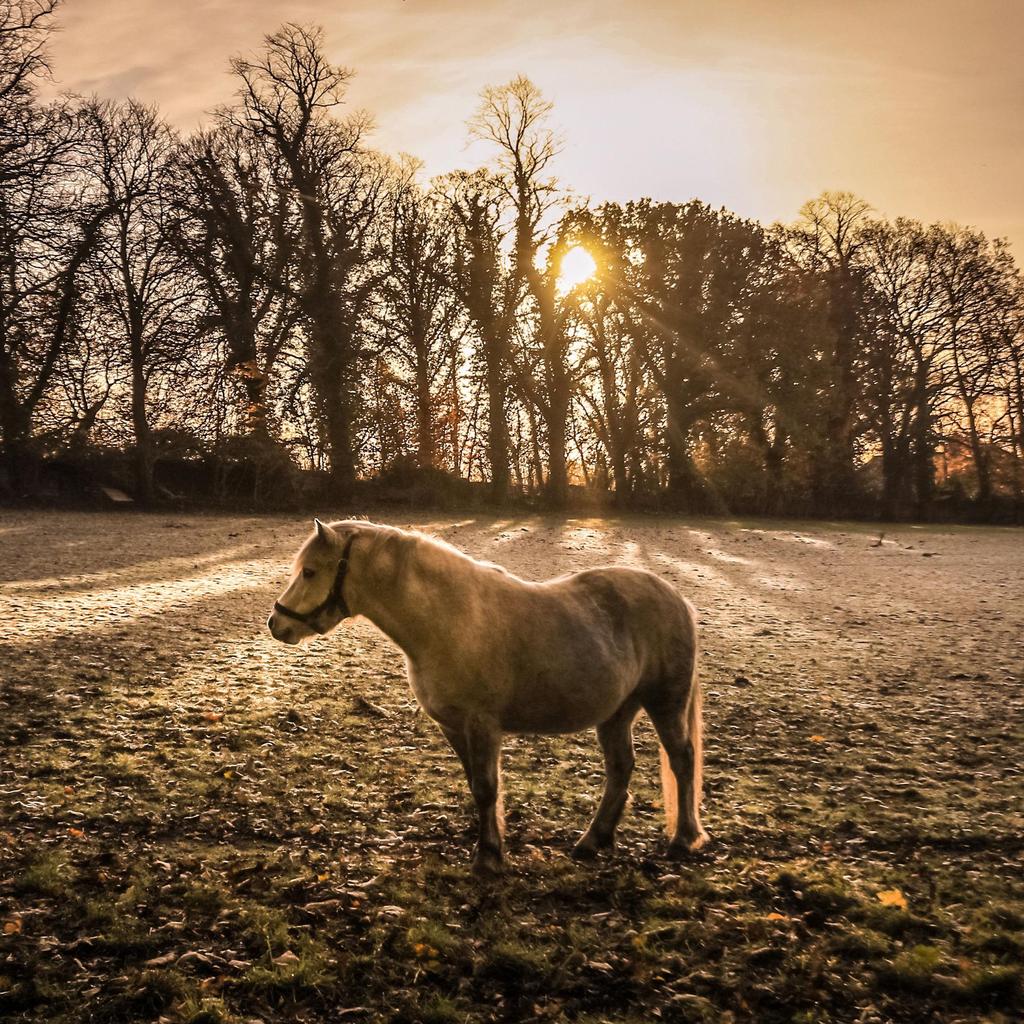}
{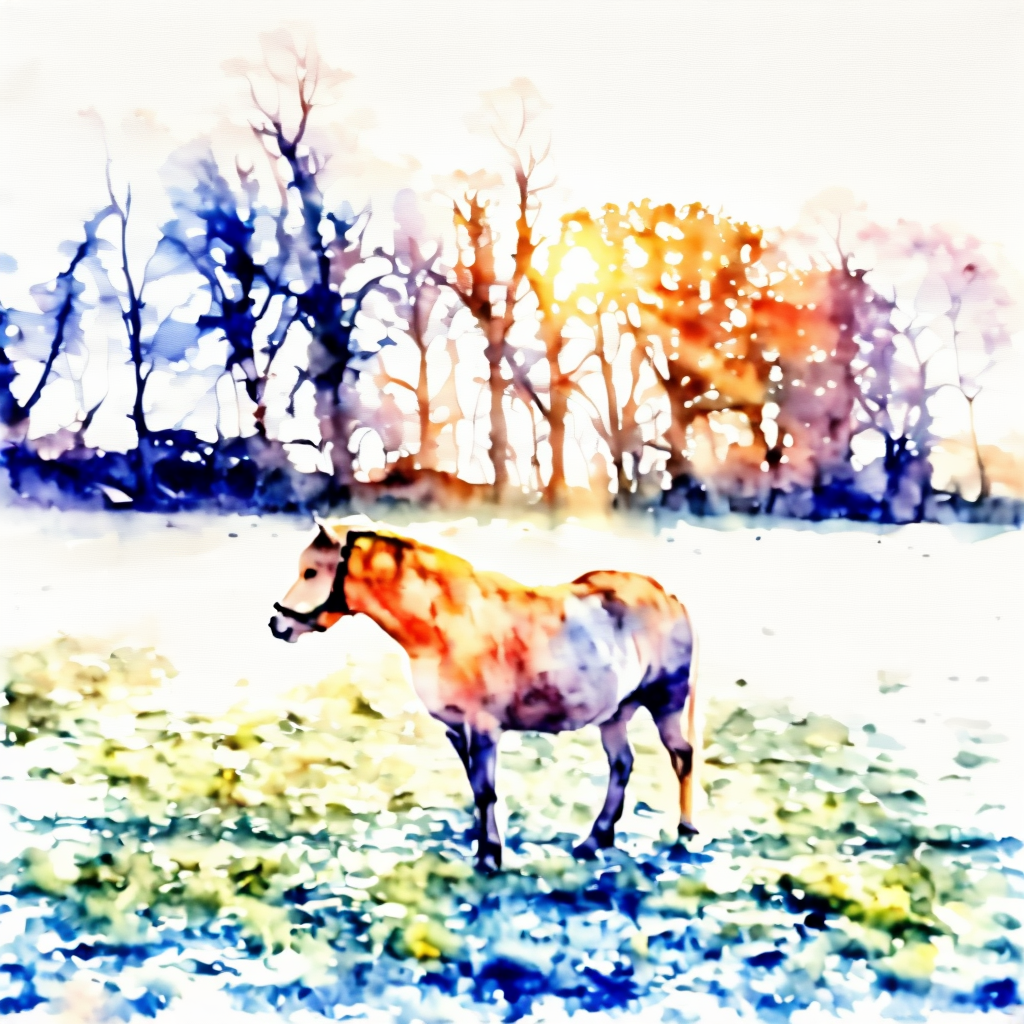}
{A white horse stands in a field with a sunset in the background.}
{(b) Understanding miss}
{\alignsmile}
{\alignsad}
\alignentrysep
{Add a small wooden cabin to the left part of the image, near the tree, blending naturally with the landscape.}
{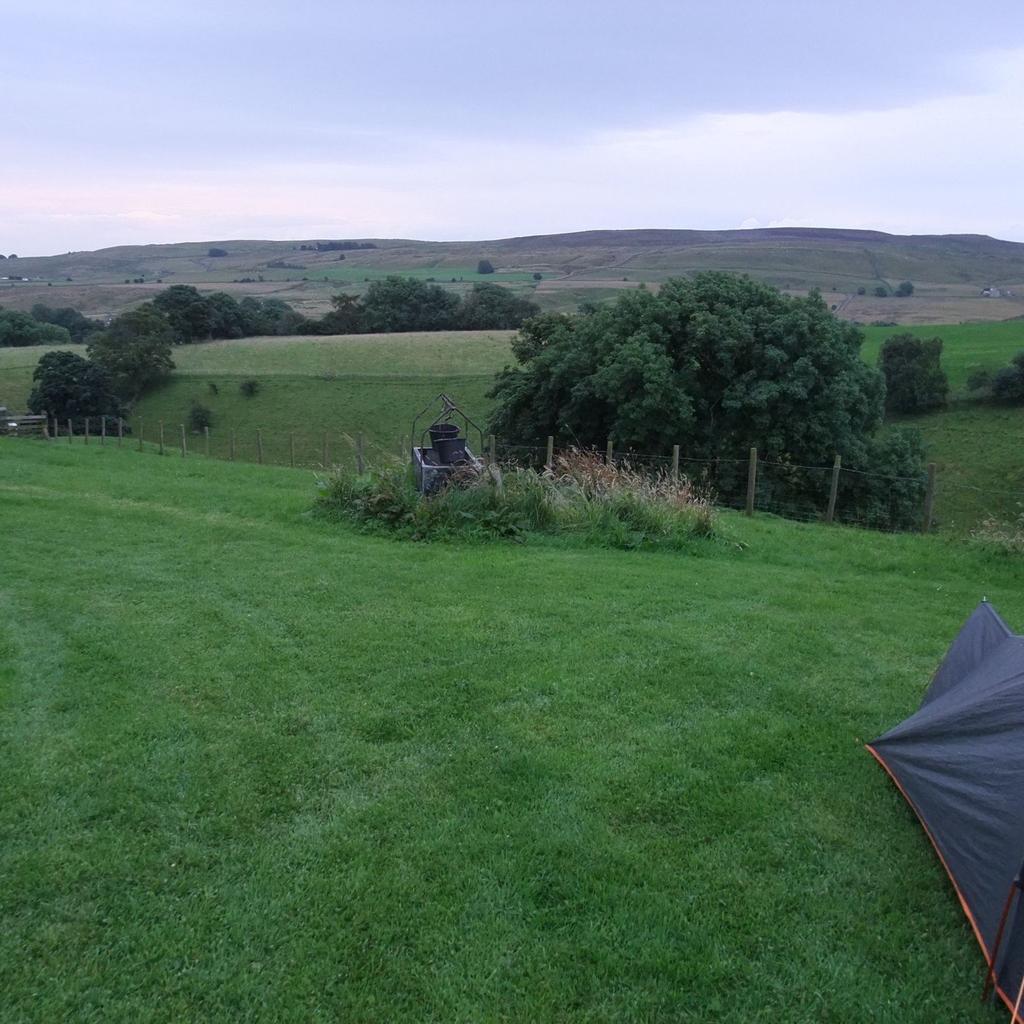}
{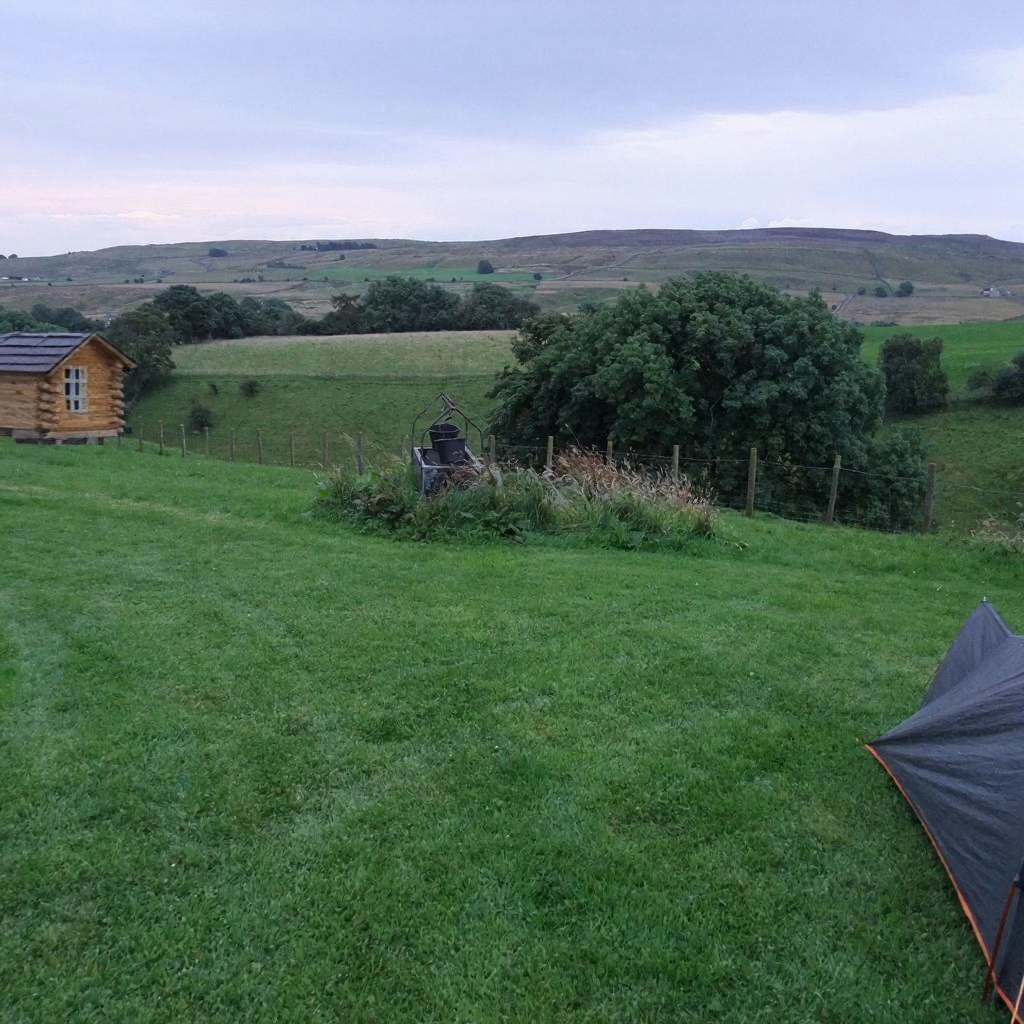}
{A small wooden cabin is added to the left part of the image, near the tree, blending naturally with the landscape.}
{(c) Information gap}
{\alignsmile}
{\aligngapface}
\end{minipage}}
\caption{\textbf{Shared-target alignment diagnostics.} Representative diagnostics for BAGEL compare the understanding pathway and generation pathway under the same source image and editing instruction. Row (a) shows a generation failure: the caption correctly specifies the requested bicycle, but the generated image omits it. Row (b) shows an understanding failure: the generated image follows the watercolor-style edit, but the caption does not specify the style change. Row (c) shows an information-coverage mismatch: the caption captures the cabin addition but provides a much sparser target expression than the generated visual output.}
\label{fig:align_diagnostic_examples}
\endgroup
\end{figure}

To further characterize this misalignment, we analyze representative failure cases in Fig.~\ref{fig:align_diagnostic_examples}. We observe three recurring patterns. The first is \emph{generation miss}, where the model correctly describes the intended edit but fails to realize it visually. The second is \emph{understanding miss}, where the generated image follows the instruction but the textual understanding output overlooks the realized change. The third is \emph{insufficient information coverage}, where both outputs capture the coarse intent but diverge in fine-grained objects, attributes, spatial relations, or preservation constraints. These cases suggest that the problem is not merely an isolated weakness of either understanding or generation. Rather, the two capabilities lack an explicit mechanism that anchors them to a common target state.

Existing unified models typically optimize understanding and generation as separate output objectives. The understanding pathway learns to produce text, while the generation pathway learns to produce images. Although both pathways are conditioned on the same input, their outputs are not explicitly constrained to represent the same target visual result. As a consequence, the textual output and the image output may each appear plausible in isolation, but correspond to different interpretations of the intended target. This motivates a post-training framework that aligns understanding and generation through a shared target, rather than treating them as two loosely coupled tasks.

To this end, we propose \emph{STBridge}, a shared-target alignment framework for UMM post-training. STBridge treats the desired target visual result as a common semantic anchor between understanding and generation. The understanding pathway expresses this target in language, while the generation pathway realizes it visually. Under this formulation, target captioning and image editing are no longer optimized as independent tasks. Instead, they become two modal expressions of the same target state, creating an explicit and trainable relation between what the model describes and what it generates.

We instantiate this formulation in the image-editing setting. Each training sample consists of a source image, a user instruction, and a target image, denoted as (\(I_s, x, I_t\)). The source image \(I_s\) and instruction \(x\) provide the same input condition for both pathways, while the target image \(I_t\) defines the shared target visual state. The understanding pathway learns to predict a textual description of \(I_t\), which we call the \emph{target caption}. This caption describes the intended edit, the resulting visual content, and the information that should be preserved from the source image. The generation pathway learns to produce the corresponding target image. Since both outputs are aligned to the same \(I_t\), the model is encouraged to coordinate what it says with what it generates.

STBridge is trained through a three-stage post-training procedure. First, \textbf{Shared Target Alignment} uses supervised fine-tuning to establish the correspondence between the target caption and the target image. Second, \textbf{Target Expression Optimization} applies reinforcement learning to improve the predicted target caption, encouraging it to capture the intended target state more accurately and completely. Third, \textbf{Target Realization Optimization} further applies reinforcement learning to the generation branch, encouraging the edited image to faithfully realize the refined target expression. We also organize editing and understanding data into a unified editing format, allowing multiple data sources to jointly support shared-target learning. Through this design, STBridge progressively improves the consistency between target expression and visual realization.

Our contributions are summarized as follows:
\begin{itemize}
\item We identify the understanding-generation alignment gap in unified multimodal models. Rather than evaluating understanding and generation in isolation, we examine whether the two capabilities produce consistent outputs under the same user intent, and diagnose the gap through both global scene-level consistency and fine-grained alignment metrics.

\item We propose \emph{STBridge}, a shared-target alignment framework that treats understanding and generation as two modal expressions of the same target visual state. This formulation establishes an explicit semantic anchor between target captioning and image editing.

\item We design a three-stage post-training procedure for shared-target learning. Supervised fine-tuning first establishes the correspondence between target captions and target images, while sequential reinforcement learning further refines target expression and visual realization.

\item Experiments show that STBridge brings consistent improvements across visual understanding, image generation, and image editing benchmarks, with gains on BLINK (+5.15), CVBench-2D/3D (+1.19/+1.92), OddGridBench (+5.14), MMVP (+2.70), WISE (+1.20), ImgEdit (+0.46), and RISE (+21.00). Alignment analysis further confirms that STBridge substantially narrows the gap between what the model describes and what it generates, demonstrating the effectiveness of shared-target alignment for UMM post-training.
\end{itemize}

\section{Related Work}

\paragraph{Unified multimodal models.}
Unified multimodal models aim to support visual understanding and visual generation within a single modeling framework. Early efforts study unified sequence-to-sequence or autoregressive modeling across tasks and modalities, such as OFA~\cite{Wang_2022_OFA}, Unified-IO 2~\cite{Lu_2023_UnifiedIO2}, LaVIT~\cite{Jin_2023_LaVIT}, VL-GPT~\cite{Zhu_2023_VLGPT}, DreamLLM~\cite{Dong_2023_DreamLLM}, Emu2~\cite{Sun_2023_Emu2}, and SEED-X~\cite{Ge_2024_SEEDX}. Recent UMMs explore different forms of native unification, including early-fusion token-based modeling~\cite{Chameleon_2024}, hybrid autoregressive and diffusion modeling~\cite{Xie_2024_Showo,Xie_2025_Showo2,Zhou_2024_Transfusion}, next-token prediction over tokenized multimodal sequences~\cite{Wang_2024_Emu3}, unified diffusion-language foundations~\cite{Wu_2024_VILAU,Tong_2024_MetaMorph,Chen_2025_BLIP3o}, and decoupled visual encoders within a unified transformer~\cite{Wu_2024_Janus,Chen_2025_JanusPro}. More recent systems further strengthen native understanding-and-generation capabilities: BAGEL scales decoder-only unified pretraining on large interleaved multimodal data~\cite{deng2025emerging}; InternVL-U extends the InternVL family toward unified understanding, reasoning, generation, and editing~\cite{Tian_2026_InternVLU}; the TUNA series studies unified visual representations and pixel embeddings for native multimodal understanding and generation~\cite{Liu_2025_TUNA,Liu_2026_Tuna2}; SenseNova-U1 introduces a unified architecture for multimodal understanding and generation~\cite{Diao_2026_SenseNovaU1}; and OneCAT studies decoder-only autoregressive unification~\cite{Li_2025_OneCAT}. These works make unified architectures increasingly concrete.

\paragraph{Post-training unified multimodal models.}
Existing post-training methods for UMMs mostly follow two routes. The first route mixes diverse understanding and generation data, often together with supervised fine-tuning, preference optimization, or RL objectives~\cite{tian2026unigen,Mao_2025_UniRL,Li_2026_LaViDaR1,Nie_2026_InterleavedGRPO,Mo_2026_LVRPO}. This can improve unified models, but the training recipe becomes sensitive to task composition and data balancing. The second route uses one capability to supervise the other. Reconstruction or auxiliary generation targets use generation-style objectives to improve visual understanding~\cite{Su_2026_UniMRG,Yan_2025_UMMAE}, while understanding-derived signals can guide or reward generation~\cite{Xie_2025_RecA,Pan_2026_GvU,Liu_2026_UNO}. These methods reveal useful links between understanding and generation, but the two capabilities are usually optimized as separate outputs or in a directional manner. STBridge differs by establishing a shared target through an editing-style interface and training it with a three-stage post-training procedure, so that target expression and visual realization are optimized along the same information flow.

\paragraph{Rewards for visual generation and editing.}
Reinforcement learning and reward modeling have become increasingly important for visual generation and editing. Flow-GRPO~\cite{Liu_2025_FlowGRPO} adapts online RL to flow matching models, and MixGRPO~\cite{Li_2025_MixGRPO} improves flow-based GRPO efficiency with mixed ODE-SDE training. For image editing, EditReward trains a human-aligned reward model~\cite{Wu_2025_EditReward}, while ThinkRL-Edit and Edit-R1 study reasoning- or verifier-based rewards for editing RL~\cite{Li_2026_ThinkRLEdit,Guo_2026_EditR1}. These works provide useful optimization and evaluation signals for visual outputs. STBridge uses rewards to optimize the shared-target alignment, first improving target expression and then improving visual realization.

\section{Data Construction}
\label{sec:data_construction}

We construct the training data for STBridge in two components: supervised fine-tuning (SFT) data and reinforcement learning (RL) data. Both components are organized under the editing interface \((I_s,x,I_t)\), where \(I_s\) denotes the source image, \(x\) the instruction, and \(I_t\) the target visual result. The SFT data supports the alignment stage, while the RL data supports the subsequent optimization stage. Fig.~\ref{fig:data_composition} summarizes their data type composition, with related fine-grained subtypes grouped into coarse categories for readability; the detailed construction is described in the following subsections.

\begin{center}
\centering
\begingroup
\captionsetup{type=figure}
\setlength{\abovecaptionskip}{4pt}
\setlength{\belowcaptionskip}{0pt}
\makebox[\textwidth][c]{\includegraphics[width=\textwidth]{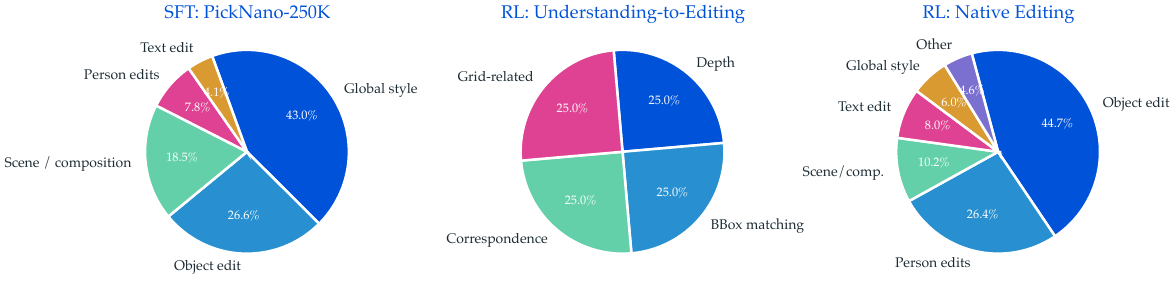}}
\caption{\textbf{Training data composition.} The SFT pie summarizes the PickNano-250K SFT split. The RL pies summarize the converted visual-understanding data and native image-editing data after grouping related fine-grained subtypes into coarse categories.}
\label{fig:data_composition}
\endgroup
\end{center}

\subsection{Training Data for the SFT Stage}
The SFT data is constructed to provide paired supervision for target expression and visual realization in the alignment stage. Starting from an editing example \((I_s,x,I_t)\), we augment the source-instruction pair with a target expression \(z^*=c_t^*\), where \(c_t^*\) is a target caption that makes the target visual result explicit before image generation.

We build this data by directly using the PickNano-250K SFT split~\cite{qian2026pico}, which contains 257,730 examples, without additional filtering, resampling, or taxonomy-based stratification. Each example provides a source image \(I_s\), an editing instruction \(x\), and a target image \(I_t\).

As shown in Fig.~\ref{fig:sft_annotation_pipeline}, we annotate each example with a target caption using Qwen3.5-35B-A3B~\cite{Qwen_2025_Qwen3}. Specifically, we feed the target image \(I_t\) to the VLM and ask it to produce a detailed visual description of the edited result as the target caption \(c_t^*\). The resulting SFT example is represented as \((I_s,x,c_t^*,I_t)\), which supports the SFT objective of predicting \(z^*=c_t^*\) from \((I_s,x)\) and then generating \(I_t\) conditioned on this target expression. A concrete example is shown in Fig.~\ref{fig:sft_data_demo}.

\begin{center}
\centering
\begingroup
\captionsetup{type=figure}
\setlength{\abovecaptionskip}{3pt}
\setlength{\belowcaptionskip}{0pt}
\newcommand{\sftflowfont}{\fontsize{6.5pt}{6.9pt}\selectfont}
\begin{minipage}{\textwidth}
\centering
\begin{tikzpicture}[
    x=1cm,y=1cm,
    every node/.style={font=\sftflowfont},
    box/.style={draw=youtuBlue!25,fill=youtuLightBlue!55,rounded corners=1.5pt,line width=0.3pt,align=center,inner xsep=3pt,inner ysep=3pt,text width=2.05cm,minimum height=1.02cm},
    data/.style={draw=youtuBlue!30,fill=white,rounded corners=1.5pt,line width=0.3pt,align=center,inner xsep=3pt,inner ysep=3pt,text width=2.05cm,minimum height=1.02cm},
    model/.style={draw=youtuBlue!45,fill=youtuBlue!8,rounded corners=1.5pt,line width=0.35pt,align=center,inner xsep=3pt,inner ysep=3pt,text width=2.05cm,minimum height=1.02cm},
    output/.style={draw=youtuBlue!35,fill=youtuLightBlue!75,rounded corners=1.5pt,line width=0.35pt,align=center,inner xsep=3pt,inner ysep=3pt,text width=2.05cm,minimum height=1.02cm},
    tuple/.style={draw=youtuBlue!45,fill=youtuLightBlue!85,rounded corners=1.5pt,line width=0.35pt,align=center,inner xsep=3pt,inner ysep=3pt,text width=2.05cm,minimum height=1.02cm},
    arrow/.style={->,>=stealth,draw=youtuBlue!70,line width=0.45pt,shorten >=1.5pt,shorten <=1.5pt},
    branch/.style={->,>=stealth,draw=youtuBlue!55,line width=0.4pt,shorten >=1.5pt,shorten <=1.5pt}
]
\node[box] (src) at (-6.65,0) {\textbf{PickNano SFT split}\\source, instruction, target};
\node[box] (subset) at (-3.99,0) {\textbf{250K examples}\\direct usage\\no extra filtering};
\node[data] (targetimg) at (-1.33,0) {\textbf{Target image}\\\(I_t\)};
\node[model] (capvlm) at (1.33,0) {\textbf{Qwen3.5-35B-A3B}\\VLM description};
\node[output] (caption) at (3.99,0) {\textbf{Target Caption}\\\(c_t^*\)};
\node[tuple] (final) at (6.65,0) {\textbf{SFT tuple}\\\((I_s,x,c_t^*,I_t)\)};

\draw[arrow] (src) -- (subset);
\draw[arrow] (subset) -- (targetimg);
\draw[arrow] (targetimg) -- (capvlm);
\draw[arrow] (capvlm) -- (caption);
\draw[arrow] (caption) -- (final);
\end{tikzpicture}
\caption{\textbf{SFT annotation pipeline.} We directly use the 250K PickNano SFT split and annotate each example with a target caption.}
\label{fig:sft_annotation_pipeline}
\end{minipage}
\endgroup
\end{center}

\begin{center}
\centering
\begingroup
\captionsetup{type=figure}
\setlength{\abovecaptionskip}{3pt}
\setlength{\belowcaptionskip}{0pt}
\newcommand{\sftheadfont}{\fontsize{6.1pt}{6.1pt}\selectfont}
\newcommand{\sftinstrfont}{\fontsize{5.8pt}{6.25pt}\selectfont}
\newcommand{\sftboxfont}{\fontsize{5.0pt}{5.25pt}\selectfont}
\definecolor{sftLogoMagenta}{RGB}{223,66,146}
\newcommand{\sftkey}[1]{\textcolor{sftLogoMagenta}{\textbf{#1}}}
\newcommand{\sftimageheight}{0.220\textheight}
\newcommand{\sftimageboxheight}{0.232\textheight}
\newcommand{\sftheader}[1]{\centering{\sftheadfont\textbf{\textcolor{youtuBlue}{#1}}}\par\vspace{2pt}}
\newcommand{\sftimagebox}[1]{%
\begin{tcolorbox}[colback=white,colframe=youtuBlue!18,boxrule=0.25pt,arc=1.2pt,left=1.2pt,right=1.2pt,top=1.2pt,bottom=1.2pt,width=\linewidth,height=\sftimageboxheight,valign=center,before skip=0pt,after skip=0pt]
\centering\includegraphics[width=\linewidth,height=\sftimageheight,keepaspectratio]{#1}
\end{tcolorbox}}
\newcommand{\sfttextbox}[1]{%
\begin{tcolorbox}[colback=youtuLightBlue!38,colframe=youtuBlue!18,boxrule=0.25pt,arc=1.2pt,left=1.6pt,right=1.6pt,top=1.6pt,bottom=1.6pt,width=\linewidth,before skip=0pt,after skip=0pt]
{\raggedright\sftboxfont\setlength{\parindent}{0pt}\setlength{\parskip}{0pt}#1\par}
\end{tcolorbox}}
\makebox[\textwidth][c]{%
\begin{minipage}{\textwidth}
\begin{tcolorbox}[colback=youtuLightBlue!35,colframe=youtuBlue!18,boxrule=0.25pt,arc=1.2pt,left=4pt,right=4pt,top=3pt,bottom=3pt,width=\linewidth,before skip=0pt,after skip=0pt]
{\raggedright\sftinstrfont\setlength{\parindent}{0pt}\setlength{\parskip}{0pt}\textbf{Instruction.} Remove the black baseball cap from the player, naturally generating realistic hair that matches the player's apparent age and facial features, ensuring the lighting, color, and texture of the hair and exposed head seamlessly integrate with the existing image's overall composition and realism.\par}
\end{tcolorbox}
\vspace{3pt}
\noindent
\begin{minipage}[t]{0.490\linewidth}
\sftheader{Source Image}
\sftimagebox{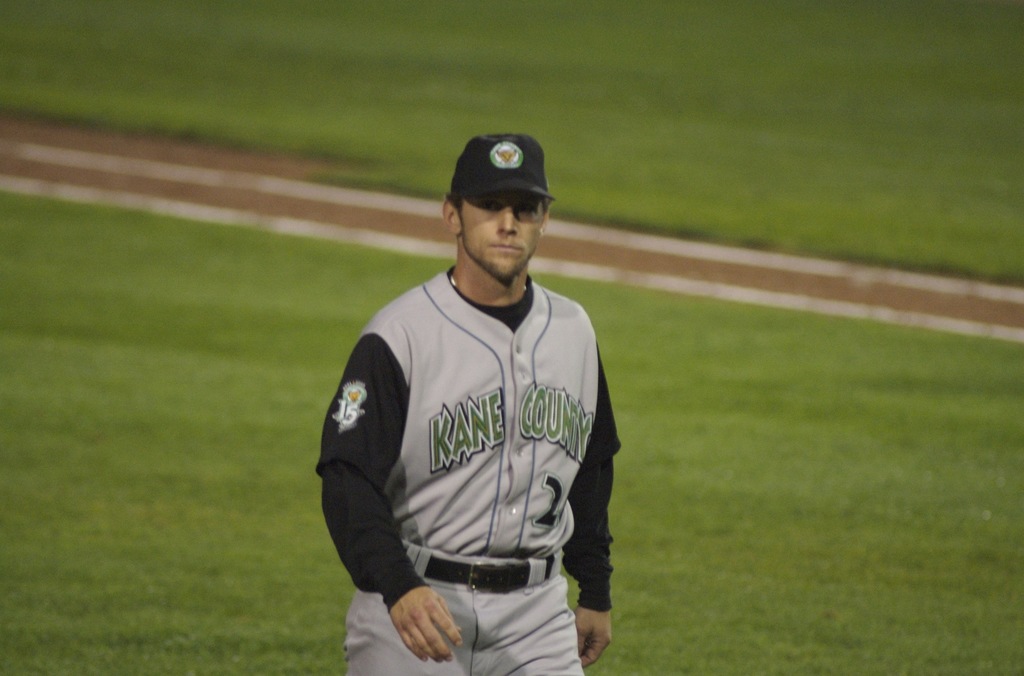}
\end{minipage}\hfill
\begin{minipage}[t]{0.490\linewidth}
\sftheader{Target Image}
\sftimagebox{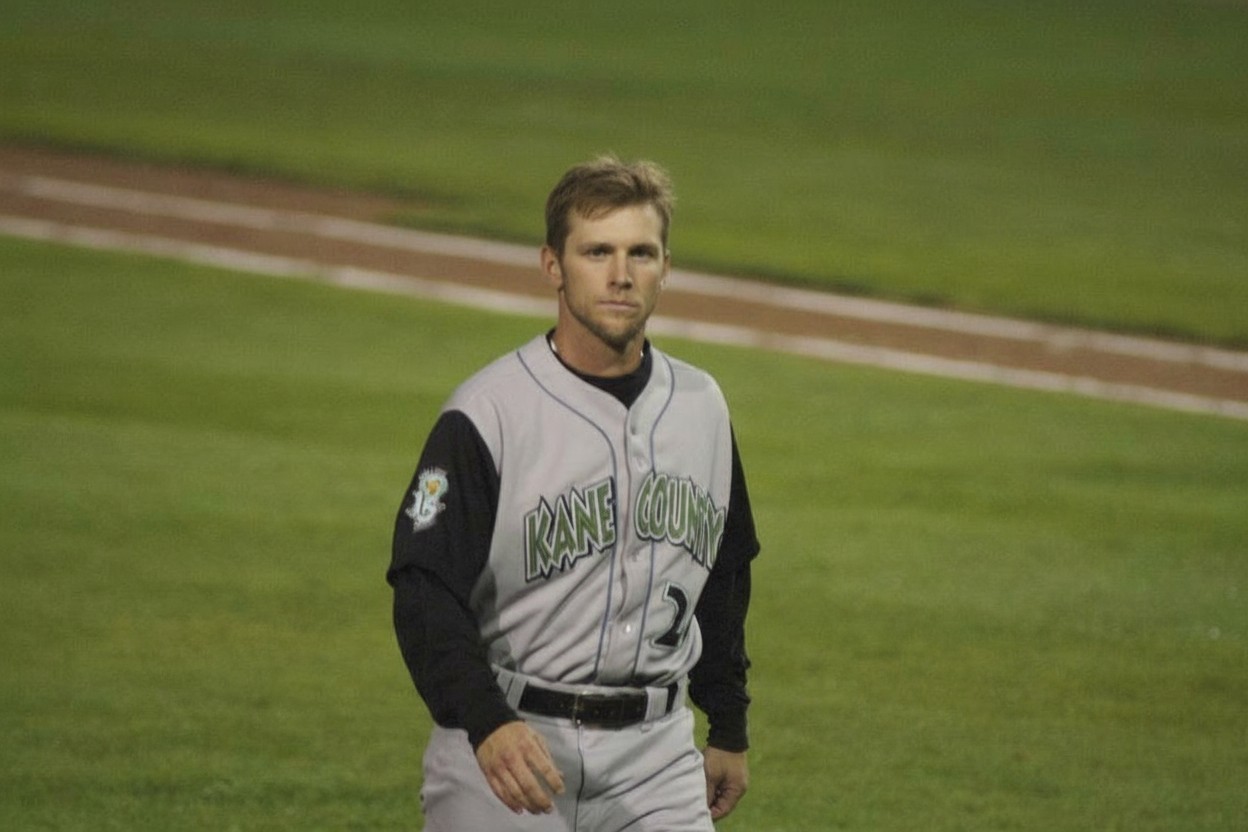}
\end{minipage}\hfill
\vspace{4pt}
\noindent
\begin{minipage}[t]{\linewidth}
\sftheader{Target Caption}
\sfttextbox{The overall scene is an outdoor baseball field captured under artificial stadium lighting, characterized by a vast expanse of green grass and a diagonal strip of reddish-brown dirt running from the upper left to the middle right. The viewpoint is slightly elevated and positioned in front of the subject, creating a medium shot that frames the player from the thighs up against the blurred background of the playing surface. \sftkey{In the center foreground, a male baseball player stands facing forward with a neutral facial expression and short, light brown hair.} \sftkey{He is wearing a grey button-up jersey with black long sleeves underneath; the jersey features the text "KANE COUNTY" in green lettering with black outlines across the chest and the number "2" visible on the lower right area.} A circular patch with a white background and dark graphic is located on his right upper arm. His arms hang loosely along his body, and he wears a black belt threaded through the loops of his grey pants. The player is positioned in front of the field, with the dirt path running horizontally behind his torso, separating the foreground grass from the background grass. \sftkey{The lighting casts soft shadows on his face and uniform, highlighting the texture of the fabric and the contours of his face, while the background remains out of focus to emphasize the subject.}}
\end{minipage}
\end{minipage}}
\caption{\textbf{SFT training example.} A concrete SFT example pairs the source-target image pair and instruction with a target caption that describes the target visual result. The presented images are selected from the open-source Pico-Banana-400K dataset~\cite{qian2026pico}.}
\label{fig:sft_data_demo}
\endgroup
\end{center}

\subsection{Training Data for the RL Stage}
For RL post-training, we organize all data sources under the same editing-style tuple \((I_s,x,I_t)\). Here, \(I_s\) is the source image, \(x\) is an instruction, and \(I_t\) is the reference target visual result used for offline data preparation but not provided to the model during RL training. This design keeps the data construction consistent with the shared-target interface established by SFT, and allows heterogeneous data sources to be represented as transformations from a source image and instruction to a target visual result.

Native image-editing data directly provides source-instruction-target transformations, which are essential for training the model's editing capabilities. However, such data primarily captures the final editing transformations and may not sufficiently provide the rich visual understanding signals that are crucial for accurate target expression. Visual-understanding data offers complementary information that editing data cannot provide, such as explicit spatial reasoning (depth, distance, relative positioning), precise correspondence matching, attribute recognition, relation understanding, and local preservation constraints. These understanding abilities are fundamental for interpreting the editing instruction and predicting what should be changed versus what should remain unchanged. To incorporate these essential understanding signals into RL training and enable both understanding and generation pathways to learn from the same data, we organize visual-understanding data as source-instruction-target transformations under the shared editing-style target interface. Given an image \(I\), a question \(q\), and its answer \(a\), we treat the image as the source image, rewrite the question as an editing-style instruction \(x=\tau(q)\), and let the answer determine an answer-revealing target visual result \(I_t^{ans}=\rho(I,q,a)\):
\begin{equation}
    (I,q,a) \Rightarrow (I_s=I,\ x=\tau(q),\ I_t=I_t^{ans}=\rho(I,q,a)).
\end{equation}
The target visual result makes the answer visible or verifiable, e.g., by highlighting, cropping, correspondence marking, or spatial transformation. Fig.~\ref{fig:data_interface_examples} illustrates this representation. In the depth example, the derived instruction asks the model to highlight the point closer to the camera in green. In the grounding example, it asks the model to crop the selected object. In the correspondence example, it asks the model to mark the matching point across two views. In the grid reasoning example, it asks the model to apply the requested small spatial translation while keeping the other cells fixed. This visual-understanding subset covers depth and distance reasoning, spatial reasoning, grounding, correspondence, attribute recognition, and relation understanding.
This conversion is not intended to turn visual understanding into ordinary image editing. Instead, it makes the answer-defining visual evidence explicit as a target visual result, so that both the target-expression pathway and the visual-realization pathway can be trained around the same shared target.

\begin{center}
\centering
\begingroup
\captionsetup{type=figure}
\setlength{\abovecaptionskip}{3pt}
\setlength{\belowcaptionskip}{0pt}
\newcommand{\dataheadfont}{\fontsize{5.8pt}{5.8pt}\selectfont}
\newcommand{\datainstrfont}{\fontsize{6.4pt}{6.7pt}\selectfont}
\newcommand{\dataimageheight}{0.122\textheight}
\newcommand{\datainstrheight}{0.045\textheight}
\newcommand{\dataimgbox}[3]{%
\begin{minipage}[t]{#1\linewidth}
\centering
{\dataheadfont\textbf{\textcolor{youtuBlue}{#2}}}\par\vspace{0.3pt}
\begin{minipage}[c][\dataimageheight][c]{\linewidth}
\centering
\includegraphics[width=\linewidth,height=\dataimageheight,keepaspectratio]{#3}
\end{minipage}
\end{minipage}}
\newcommand{\dataentry}[5]{%
\begin{minipage}[t]{0.48\linewidth}
\centering
\begin{tcolorbox}[colback=white,colframe=gray!30,boxrule=0.4pt,arc=1.0pt,left=1.5pt,right=1.5pt,top=1.5pt,bottom=1.5pt,width=0.94\linewidth,before skip=2pt,after skip=2pt]
\centering
{\dataheadfont\textbf{\textcolor{youtuBlue}{Question-derived Instruction}}}
\vspace{0.3pt}
\begin{tcolorbox}[colback=youtuLightBlue!45,colframe=youtuLightBlue!60,boxrule=0.3pt,arc=1.2pt,left=1.2pt,right=1.2pt,top=1.2pt,bottom=1.2pt,width=\linewidth,height=\datainstrheight,valign=center,before skip=0pt,after skip=0pt]
{\datainstrfont\centering\setlength{\parindent}{0pt}\setlength{\parskip}{0pt}#2\par}
\end{tcolorbox}
\par\vspace{1.5pt}
\dataimgbox{#4}{Source Image}{#1}\hfill
\dataimgbox{#5}{Target Image}{#3}
\end{tcolorbox}
\end{minipage}}
\makebox[\textwidth][c]{%
\begin{minipage}{\textwidth}
\centering
\dataentry
{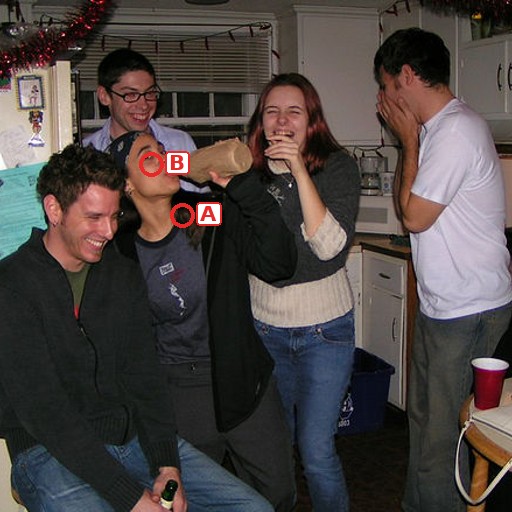}
{Choose between point A and point B. \textcolor{magenta}{\textbf{Highlight the closer point in green and the farther point in gray.}}}
{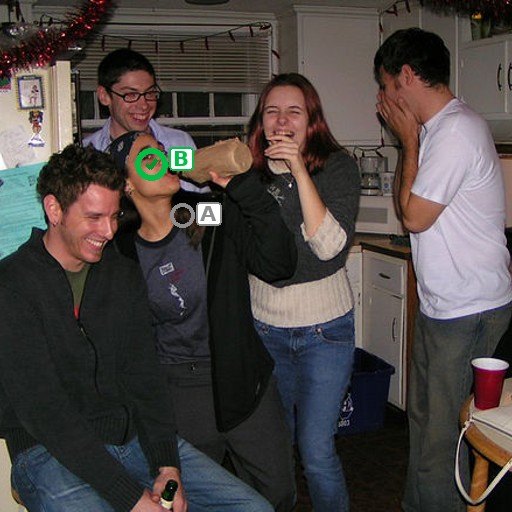}
{0.485}
{0.485}
\hspace{8pt}
\dataentry
{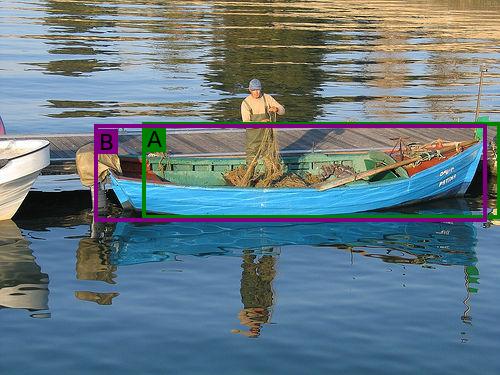}
{Choose the bounding box that best matches the category ``boat'', then \textcolor{magenta}{\textbf{crop it as a tight target image.}}}
{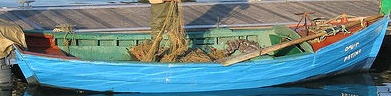}
{0.485}
{0.485}
\par\vspace{6pt}
\dataentry
{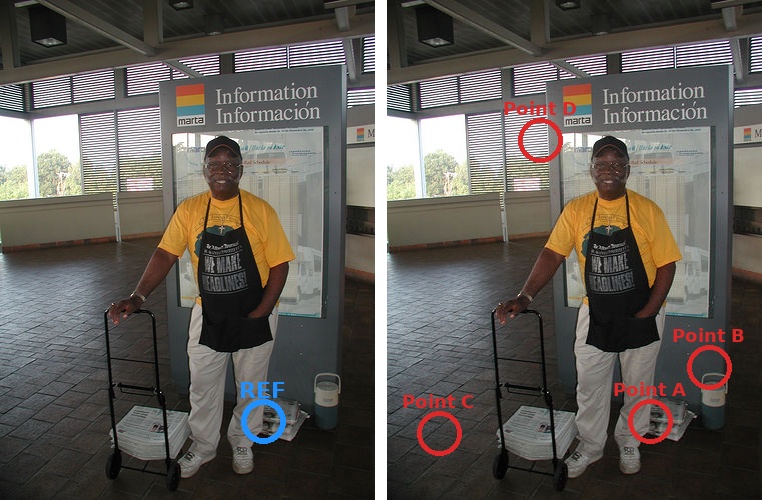}
{Find the labeled point on the right panel that matches the blue REF location on the left. \textcolor{magenta}{\textbf{Highlight it in green and add a green check mark.}}}
{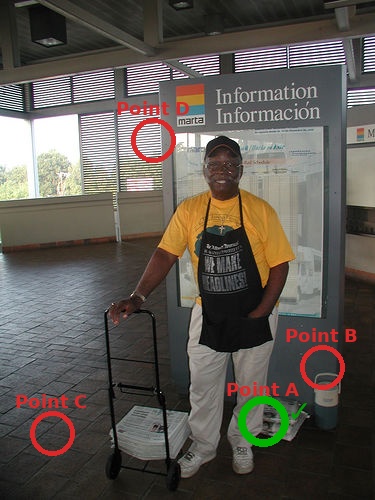}
{0.62}
{0.34}
\hspace{8pt}
\dataentry
{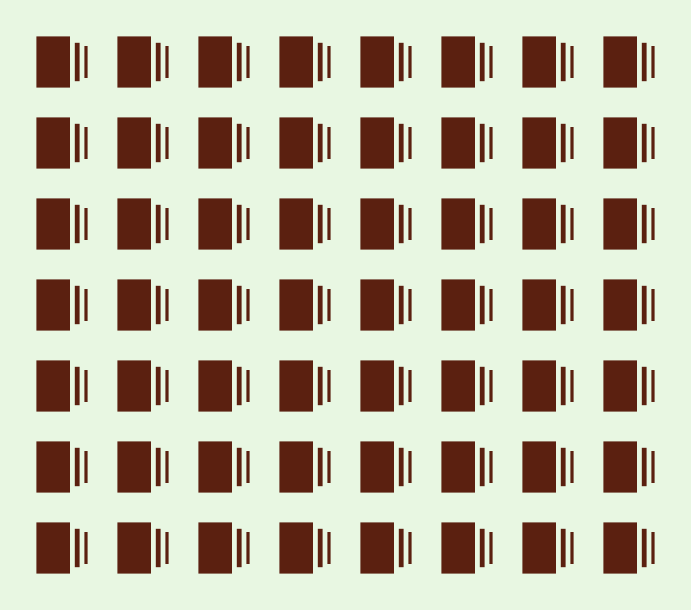}
{\textcolor{magenta}{\textbf{Translate only the mark at Row~2, Column~8 by (+5, -8) image pixels.}} Keep all other cells fixed.}
{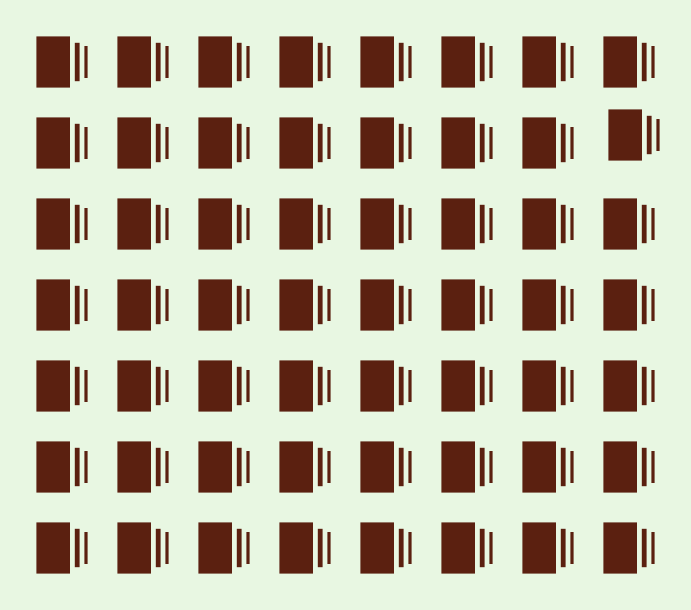}
{0.485}
{0.485}
\end{minipage}}
\caption{\textbf{Understanding-to-editing interface.} We represent visual-understanding examples under the shared editing-style target interface. Each example shows a source-instruction-target transformation derived from a visual-understanding question, demonstrating highlighting, cropping, correspondence marking, and spatial transformation operations. This shared interface enables both understanding and generation pathways to learn from the same data, facilitating information sharing between the dual-pathway model. The presented images are selected from the open-source Flickr30k dataset~\cite{young2014image}.}
\label{fig:data_interface_examples}
\endgroup
\end{center}

For native image-editing data, we construct examples from the ThinkEdit-RL multi-turn editing subset introduced in EditThinker~\cite{Li_2025_EditThinker}. Since the source dataset contains diverse editing operations, we perform stratified sampling by edit type so that the selected data preserves balanced coverage across editing skills. We then use the dataset-provided annotation scores to filter the multi-turn trajectories: we keep examples whose first-round edit score is greater than or equal to 3 and less than 7. This criterion selects cases that provide appropriately challenging editing data. Each retained example is represented in the same tuple \((I_s,x,I_t)\), where \(I_s\) is the source image for the selected turn, \(x\) is the corresponding editing instruction, and \(I_t\) is the improved edited result. The resulting data covers local and global edits, including object modification, attribute editing, background change, multi-entity editing, and content-preserving transformation.

Combining these two sources yields the final mixed training set for the RL stage, consisting of about 15.3K examples: 8K visual-understanding examples represented as source-instruction-target transformations and 7,349 native image-editing examples. Additionally, the full mixed pool is used in both stages of the subsequent sequential RL procedure: we do not reserve visual-understanding examples for understanding-pathway RL or native image-editing examples for generation-pathway RL.

\section{Method}
\label{sec:method}

\begin{figure}[!htbp]
\centering
\setlength{\abovecaptionskip}{4pt}
\setlength{\belowcaptionskip}{0pt}
\makebox[\textwidth][c]{\includegraphics[width=\textwidth]{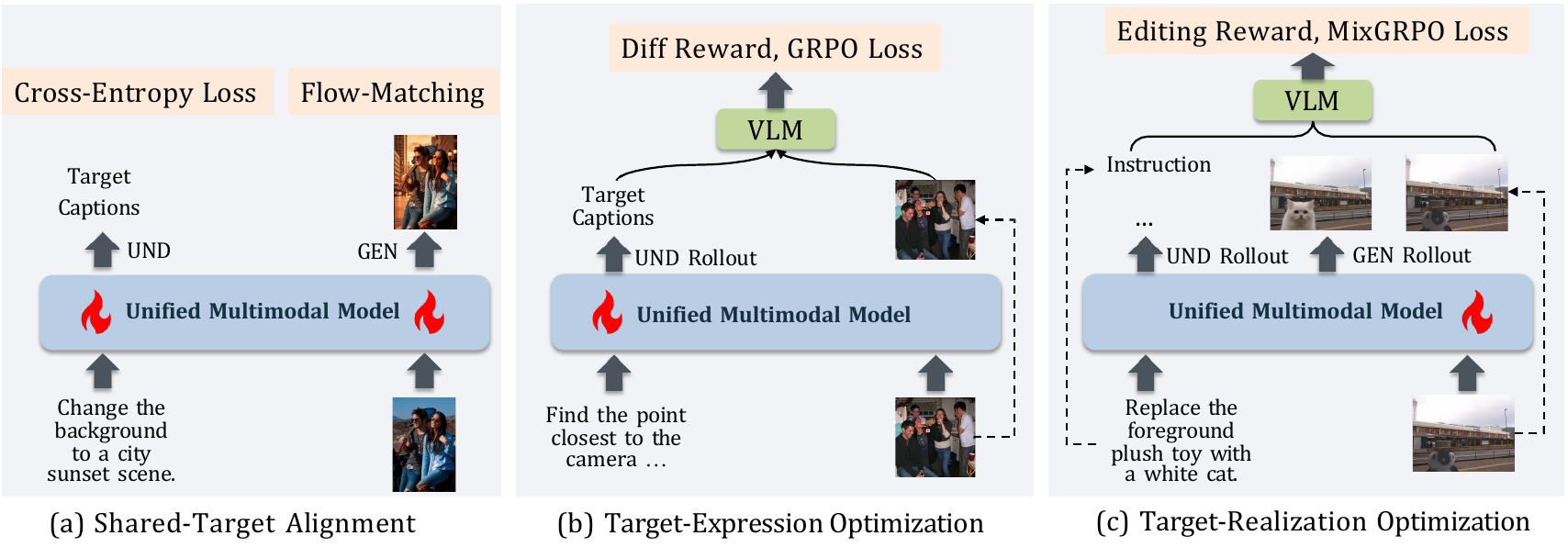}}
\caption{\textbf{STBridge pipeline.} Overview of the STBridge three-stage post-training procedure. \textbf{(a)} Shared-target SFT establishes the shared-target alignment by supervising the model to predict a textual target expression and then realize the same shared target as the edited image. \textbf{(b)} Target-expression RL optimizes the textual expression using a Diff Reward without image rollouts. \textbf{(c)} Visual-realization RL freezes the optimized target-expression policy and improves image generation conditioned on the predicted expression. The presented images are selected from the open-source Pico-Banana-400K~\cite{qian2026pico} and Flickr30k~\cite{young2014image} datasets.}
\label{fig:method_pipeline}
\end{figure}

\subsection{Overview}

STBridge uses a three-stage post-training procedure to align textual expression and visual realization through a shared target. As shown in Fig.~\ref{fig:method_pipeline}, the framework consists of three stages: \textbf{Shared-Target Alignment}, \textbf{Target-Expression Optimization}, and \textbf{Target-Realization Optimization}. These stages progressively align the model's predicted target expression with its generated edited image, ensuring that both modalities consistently refer to the same target visual result. We describe the three stages below.

\subsection{Shared Target Alignment}

The first stage establishes the shared-target alignment, rather than serving as a simple supervised warm-up. Since STBridge aims to align textual expression and visual realization around the same target visual result, the model must first learn the correspondence between these two modalities. For each SFT example \((I_s,x,c_t^*,I_t)\), the target caption \(c_t^*\) is annotated from the target image \(I_t\), and thus serves as the textual expression of the target visual result. The target image \(I_t\) provides the corresponding visual realization. Together, \(c_t^*\) and \(I_t\) offer paired supervision for the same shared target, enabling the model to learn how to express the target in text and how to realize it as an image under the same input condition.

This stage plays two roles in STBridge. First, it binds textual target expression and visual target result to the same shared target, preventing them from being learned as two unrelated outputs. Second, it provides an optimizable basis for the following RL stages. Without this stage, the following RL stages would lack a stable target-expression basis and could not optimize the same shared information flow coherently.

Concretely, this stage uses two supervised objectives. The first objective trains the model to predict the target caption \(c_t^*\) from the source image \(I_s\), instruction \(x\), and an understanding-pathway system prompt \(p_{\mathrm{sys}}\) that asks the model to output the target caption, so that it learns to express the target visual result in text. We use a standard next-token cross-entropy loss over the target-caption tokens:
\begin{equation}
\mathcal{L}_{\mathrm{cap}}
=
-\sum_{j=1}^{|c_t^*|}
\log p_\theta\left(c_{t,j}^* \mid I_s, x, p_{\mathrm{sys}}, c_{t,1:j-1}^*\right).
\end{equation}
The second objective trains the model to generate the target image \(I_t\) conditioned on the target caption, so that it learns to realize the same shared target visually. We supervise this visual realization with a flow matching loss in the image latent space:
\begin{equation}
\mathcal{L}_{\mathrm{img}}
=
\mathcal{L}_{\mathrm{FM}}\left(I_t \mid I_s, x, c_t^*\right).
\end{equation}

The final objective for this stage is:
\begin{equation}
\mathcal{L}_{\mathrm{SFT}}
=
\mathcal{L}_{\mathrm{cap}}
+
\mathcal{L}_{\mathrm{img}}.
\end{equation}
Through this paired supervision, the model obtains an initial shared-target alignment that connects the textual expression of the target caption with the visual realization of the target image. The subsequent RL stages further refine this alignment by first improving the accuracy of the target expression and then improving the fidelity of its visual realization.

\subsection{Target Expression Optimization}

After Shared-Target Alignment, the model has learned an initial shared-target alignment between the target caption and the target image. However, the target caption produced after SFT may still contain fine-grained inaccuracies, such as missing the edited entity, confusing changed and preserved content, or failing to capture local attributes and spatial relations. Therefore, this stage aims to refine the textual expression in the shared-target alignment, so that the predicted target caption can more accurately refer to the target visual result.

We use the 15.3K mixed RL training data constructed under the editing-style interface for this stage. Each example follows the shared-target triplet \((I_s,x,I_t)\), where \(I_s\) is the source image, \(x\) is the instruction, and \(I_t\) is the target image. Given \((I_s,x)\), the model samples a group of candidate target captions \(\{\hat{c}_t^{(k)}\}_{k=1}^{K}\). No edited image is rolled out in this stage; instead, the objective is to refine the target caption before visual realization is optimized. In other words, Target-Expression Optimization only optimizes the understanding pathway.

To refine the target caption, we need a reward that evaluates whether the caption correctly describes the target visual result. A direct solution is to measure the overall image-text consistency between the candidate caption \(\hat{c}_t\) and the target image \(I_t\), but such global matching is not suitable for this stage. Since image editing often changes only a local region or limited content, large portions of the target image \(I_t\), such as the background, main scene structure, and non-edited objects, may remain the same as the source image \(I_s\). As a result, caption-image similarity can be dominated by these unchanged regions. Even if a candidate caption does not accurately describe the actual edit, it may still receive a high score as long as it describes most static content in the target image. Therefore, for this stage, the reward should emphasize whether the target caption correctly reflects the edit-relevant target state with respect to the source image.

\begin{figure}[!htbp]
\centering
\setlength{\abovecaptionskip}{4pt}
\setlength{\belowcaptionskip}{0pt}
\makebox[\textwidth][c]{\includegraphics[width=\textwidth]{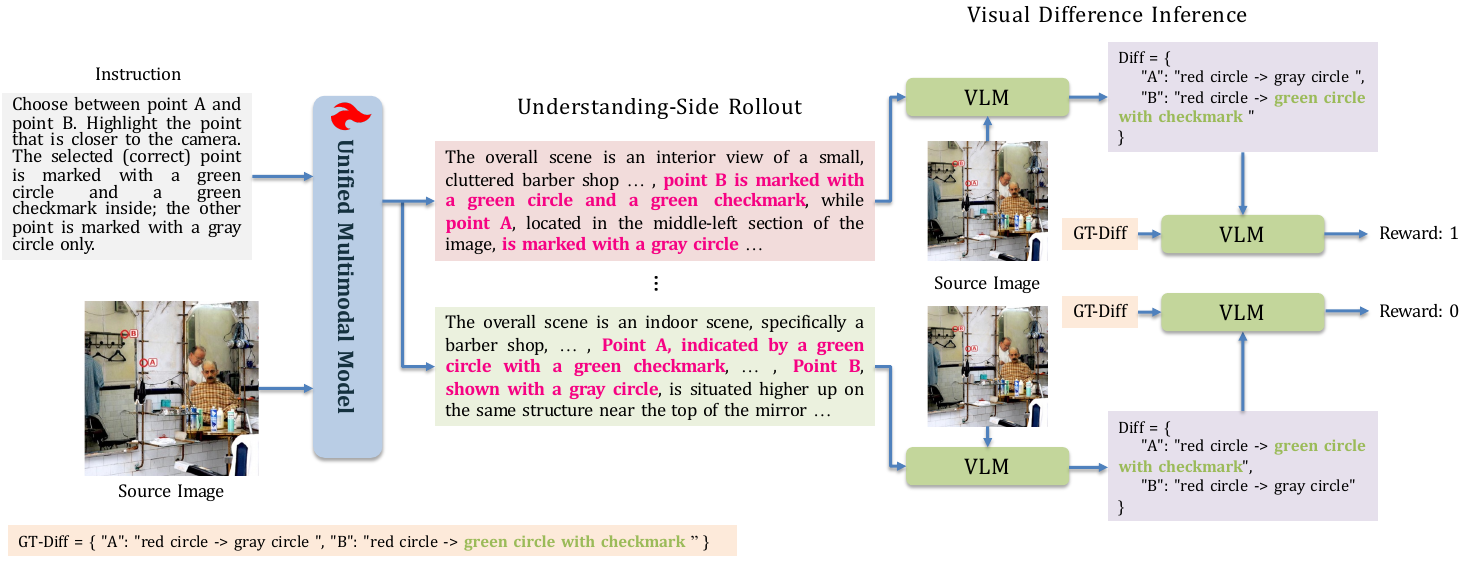}}
\caption{\textbf{Diff Reward.} Illustration of the Diff Reward used in Target-Expression Optimization. For each sampled target caption, a VLM infers the visual difference between the source image and the caption-implied target visual result. The inferred difference is compared with the annotated reference difference between the source image and the target image, and the resulting alignment score is used as the reward. This reward evaluates whether the textual target expression captures the correct edit-relevant difference without rolling out edited images. The presented images are selected from the open-source Flickr30k dataset~\cite{young2014image}.}
\label{fig:diff_reward}
\end{figure}

Based on this observation, we shift the reward from global caption-image matching to difference alignment and introduce the Diff Reward. As shown in Fig.~\ref{fig:diff_reward}, for each candidate caption \(\hat{c}_t^{(k)}\), we first ask a VLM to infer the target difference implied by this caption given the source image \(I_s\):
\begin{equation}
\widehat{\Delta}^{(k)}
=
\mathrm{Diff}_{\mathrm{VLM}}\left(I_s,\hat{c}_t^{(k)}\right).
\end{equation}
The predicted difference describes the edit implied by the candidate caption, including the object, location, and attribute to be changed, as well as the content that should remain unchanged.

We then compare the predicted difference \(\widehat{\Delta}^{(k)}\) with the offline annotated reference difference \(\Delta^*\). Here, \(\Delta^*\) is constructed from the source image \(I_s\) and the target image \(I_t\), and represents the true change required by the shared target with respect to the source image. The Diff Reward is defined as:
\begin{equation}
R_{\mathrm{diff}}^{(k)}
=
\mathrm{Align}\left(\widehat{\Delta}^{(k)}, \Delta^*\right),
\end{equation}
where \(\mathrm{Align}(\cdot,\cdot)\) returns a scalar score in \([0,1]\).

A candidate caption receives a high reward if it correctly expresses the target change and accurately distinguishes edited content from preserved content. It receives a low reward if it misses the target change, modifies the wrong object, or confuses changed and preserved regions. Therefore, Diff Reward still serves the learning of target captions, but avoids directly relying on global caption-image similarity that can be dominated by unchanged regions. Instead, it provides supervision through whether the candidate caption correctly expresses the target difference from \(I_s\) to \(I_t\). In this way, the learned target caption can describe the overall target image while better covering edit-relevant changes and preservation constraints.

We optimize the target-caption policy with GRPO. For the same source image and instruction, the model samples multiple candidate captions and computes group-relative advantages according to \(R_{\mathrm{diff}}\), increasing the likelihood of captions that better express the target difference. During this stage, the image-generation branch remains frozen. After optimization, the model obtains a more reliable target-caption policy, which provides a more accurate textual target expression for the following Target-Realization Optimization stage.

\subsection{Target Realization Optimization}

After Target-Expression Optimization, the model obtains a more reliable target-caption policy. At this point, the textual-expression pathway of the shared-target alignment has been optimized, and the goal of the third stage shifts to visual realization: with the target-caption policy frozen, we optimize the generation branch so that it can produce an edited image under the predicted target-caption condition. This further aligns visual realization with the refined target expression.

Specifically, this stage uses the same 15.3K mixed RL training set constructed in Sec.~\ref{sec:data_construction}, which consists of 8K converted visual-understanding examples and 7,349 native image-editing examples. For each sample, given the source image \(I_s\) and instruction \(x\), the frozen target-caption policy first predicts a target caption \(\hat{c}_t\). The generation branch then conditions on \(I_s\), \(x\), and \(\hat{c}_t\) to sample a group of edited-image rollouts \(\{\hat{I}_t^{(k)}\}_{k=1}^{K}\). Different from the previous stage, this stage no longer updates the target-caption policy, but only updates the generation branch. Therefore, \(\hat{c}_t\) is the target expression predicted by the frozen policy for the current input, and it serves as the conditioning signal for image generation.

To optimize visual realization, we use a reward that evaluates whether the generated image realizes the target intent under this predicted target-caption condition. We therefore adopt a realization-oriented editing reward that jointly considers edit success, instruction-detail fidelity, source preservation, visual quality, and overall preference. These dimensions measure whether the generated image performs the intended edit, follows fine-grained instruction details, preserves non-target content, and remains visually plausible.

\begin{figure}[!htbp]
\centering
\setlength{\abovecaptionskip}{4pt}
\setlength{\belowcaptionskip}{0pt}
\makebox[\textwidth][c]{\includegraphics[width=\textwidth]{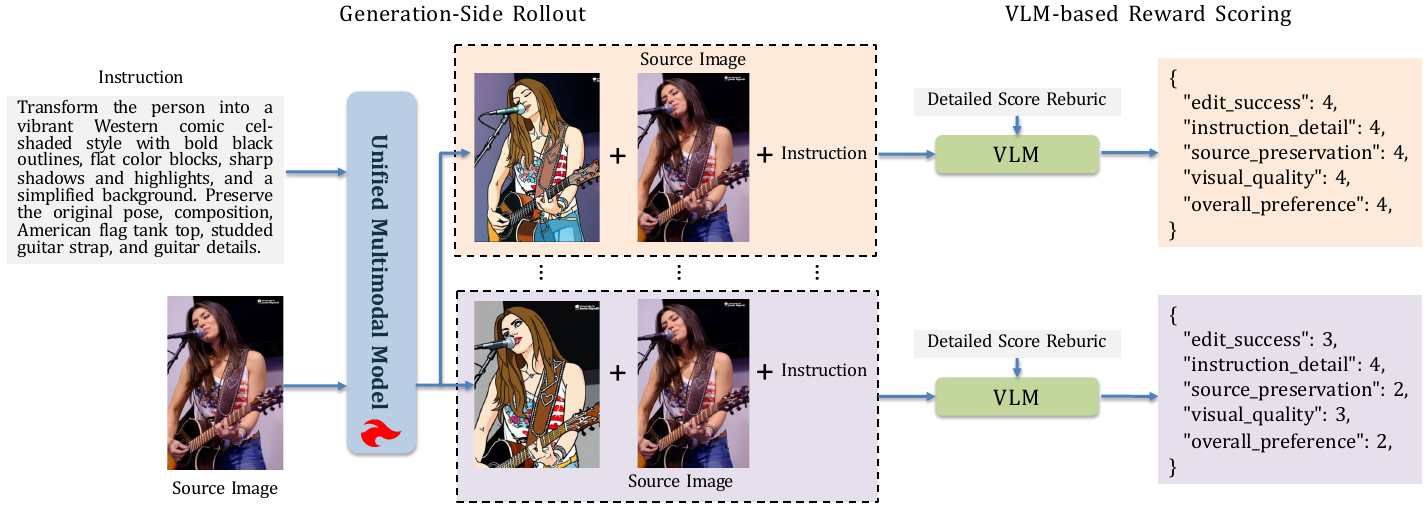}}
\caption{\textbf{Realization-oriented editing reward.} Illustration of the reward used in Target-Realization Optimization. The optimized target-expression policy is frozen, and its predicted target caption is used as the condition for image generation. A VLM judge compares each source--generated pair under the instruction and returns five rubric scores: edit success, instruction-detail fidelity, source preservation, visual quality, and overall preference. The final reward combines these scores with a gate based on edit success and preservation and is used for MixGRPO updates. The presented images are selected from the open-source EditThinker dataset~\cite{Li_2025_EditThinker}.}
\label{fig:edit_reward}
\end{figure}

As shown in Fig.~\ref{fig:edit_reward}, for each generated image \(\hat{I}_t^{(k)}\), we ask a VLM judge to compare the source--generated pair under the instruction and return five rubric scores:
\begin{equation}
    s_{\mathrm{edit}},\;
    s_{\mathrm{detail}},\;
    s_{\mathrm{preserve}},\;
    s_{\mathrm{quality}},\;
    s_{\mathrm{pref}}
    \in [0,4].
\end{equation}
These scores correspond to edit success, instruction-detail fidelity, source preservation, visual quality, and overall preference, respectively. We first aggregate them into a normalized base realization score:
\begin{equation}
    b =
    \frac{
    s_{\mathrm{edit}}
    + s_{\mathrm{detail}}
    + s_{\mathrm{preserve}}
    + s_{\mathrm{quality}}
    + s_{\mathrm{pref}}
    }{20}.
\end{equation}
To emphasize the two necessary conditions for a valid edit, we further define a gate based on edit success and source preservation:
\begin{equation}
    g =
    \alpha
    + \beta
    \frac{s_{\mathrm{edit}} + s_{\mathrm{preserve}}}{8}.
\end{equation}
Here, \(\alpha\) provides a base reward floor, and \(\beta\) controls the strength of the edit-and-preservation modulation. In our experiments, we set \(\alpha=0.1\) and \(\beta=0.9\), so that \(g\in[0.1,1.0]\).
The final visual-realization reward is defined as:
\begin{equation}
    R_G = b \cdot g .
\end{equation}
This reward assigns high values only when the generated image performs the requested edit, preserves non-target content, follows the instruction details, and remains visually plausible.

Finally, we optimize the generation branch with MixGRPO. For the same source image and instruction, the model samples multiple image rollouts and computes group-relative advantages according to \(R_G\), increasing the likelihood of images that better realize the target intent under the predicted target-caption condition. Throughout this stage, the target-caption policy remains frozen. Through this stage, STBridge aligns the refined textual target expression with high-quality visual realization.

\section{Experiments}

\subsection{Implementation Details}

We implement STBridge based on the BAGEL codebase~\cite{deng2025emerging} and initialize the model from the BAGEL checkpoint. The Shared-Target Alignment stage is trained with distributed FSDP on 8 nodes with 8 GPUs per node, for a total of 64 GPUs. We train for 500 optimization steps with a per-GPU batch size of 1, resulting in a global batch size of 64. We use a learning rate of \(2\times10^{-5}\), a cosine learning-rate schedule with a minimum learning rate of \(1\times10^{-6}\), and EMA decay of 0.995. The global and per-sample token limits are both set to 35,000.
Both Target-Expression Optimization and Target-Realization Optimization use the same 8-node, 64-GPU FSDP setup. Following the sequential freezing strategy described in Sec.~\ref{sec:method}, only the active branch is updated in each stage: Target-Expression Optimization updates the target-caption policy while keeping the image-generation branch frozen, whereas Target-Realization Optimization freezes the target-caption policy and updates only the generation branch. Both stages are trained on the same 15.3K mixed RL training set organized under the editing-style interface, with images resized to \(512\times512\) and a maximum token budget of 32,768.
Target-Expression Optimization is trained for 300 steps with GRPO, and Target-Realization Optimization is trained for 150 steps with MixGRPO~\cite{Li_2025_MixGRPO}. Both stages use a learning rate of \(5\times10^{-6}\). For both RL stages, rewards are computed online using Qwen3.5-122B-A10B served by vLLM.

\subsection{Evaluation Benchmarks}

We evaluate our model across image understanding, generation, and editing tasks. For image understanding, we use BLINK~\cite{fu2024blink}, CVBench~\cite{tong2024cambrian}, VisPuzzle~\cite{gu2025thinkmorph}, MM-Vet~\cite{yu2023mm}, MMVP~\cite{tong2024eyes}, and OddGridBench~\cite{weng2026oddgridbench} to measure fine-grained visual perception, spatial and depth reasoning, puzzle-style reasoning, holistic multimodal understanding, visual robustness, and grid-based anomaly localization. For image generation, we evaluate on WISE~\cite{niu2025wise}, which measures whether the model can generate images that satisfy world knowledge--informed semantic instructions across 25 subdomains spanning cultural common sense, spatio-temporal reasoning, and natural science. For image editing, we use ImgEdit~\cite{ye2026imgedit} for general instruction-guided editing and RISE~\cite{zhao2026envisioning} for knowledge-guided complex editing, where the model must apply the requested modification under richer semantic constraints while preserving source-image content that should remain unchanged.

\subsection{Quantitative Results}

We denote the model after Shared-Target Alignment as STBridge-SFT, and the model after the full sequential RL pipeline as STBridge-RL. In figures where space is limited, \emph{Ours} refers to STBridge-RL unless otherwise specified.

\paragraph{Image understanding.}
Tabs.~\ref{tab:blink}, \ref{tab:cvbench}, and \ref{tab:oddgridbench} report the results on visual-understanding benchmarks. Compared with BAGEL, STBridge-SFT improves the average performance on BLINK by +2.97 and OddGridBench by +2.64, while also improving VisPuzzle, MM-Vet, and MMVP by +0.50, +0.60, and +2.03, respectively. These results suggest that establishing the shared-target channel can benefit visual understanding, although the supervision is introduced through an editing-style target interface. This indicates that aligning target caption and target image under the same shared target may provide useful training signals beyond image editing itself.

STBridge-RL further improves over BAGEL by +5.15 on BLINK, +1.19/+1.92 on CVBench 2D/3D, +5.14 on OddGridBench, and +0.75/+1.30/+2.70 on VisPuzzle, MM-Vet, and MMVP. After the shared-target channel is established, the sequential RL stages further refine the shared information flow by calibrating target expression and aligning visual realization. This suggests that optimizing the shared target from both expression and realization sides can also benefit visual-understanding performance.

\clearpage
\setcounter{topnumber}{3}
\setcounter{totalnumber}{3}
\renewcommand{\topfraction}{0.95}
\renewcommand{\textfraction}{0.05}
\begin{table}[!t]
\centering
\caption{Results on the BLINK benchmark. Higher scores are better.}
\label{tab:blink}
\definecolor{youtuhighlight}{RGB}{40,144,208}
\newcommand{\blinkbest}[1]{\cellcolor{youtuhighlight!18}\textbf{#1}}
\newcommand{\blinkgain}[1]{\textcolor{youtuhighlight}{\textbf{#1}}}
\resizebox{\textwidth}{!}{
\begin{tabular}{l c c c c c c c c c c c c c c c}
\toprule
Model & Visual Sim. & Count. & Rel. Depth & Jigsaw & Art Style & Func. Corr. & Sem. Corr. & Spatial Rel. & Obj. Loc. & Visual Corr. & Multi-view & Rel. Reflect. & Forensic & IQ Test & Avg. \\
\midrule
Bagel & 89.63 & 74.17 & 86.29 & \blinkbest{83.33} & \blinkbest{76.92} & 30.77 & 36.69 & 81.82 & 66.39 & 73.84 & 51.88 & 32.09 & 37.88 & 30.67 & 60.88 \\
STBridge-SFT & 89.63 & \blinkbest{75.83} & 86.29 & \blinkbest{83.33} & \blinkbest{76.92} & 33.85 & 46.04 & 83.22 & 72.95 & 77.91 & \blinkbest{56.39} & \blinkbest{35.07} & \blinkbest{43.18} & \blinkbest{33.33} & 63.85 \\
STBridge-RL & \blinkbest{92.59} & 73.33 & \blinkbest{92.74} & 82.67 & \blinkbest{76.92} & \blinkbest{42.31} & \blinkbest{51.08} & \blinkbest{83.92} & \blinkbest{87.70} & \blinkbest{84.88} & 55.64 & 32.84 & 37.88 & 30.00 & \blinkbest{66.03} \\
\midrule
Gain & \blinkgain{+2.96} & -0.84 & \blinkgain{+6.45} & -0.66 & 0.00 & \blinkgain{+11.54} & \blinkgain{+14.39} & \blinkgain{+2.10} & \blinkgain{+21.31} & \blinkgain{+11.04} & \blinkgain{+3.76} & \blinkgain{+0.75} & 0.00 & -0.67 & \blinkgain{+5.15} \\
\bottomrule
\end{tabular}
}
\end{table}

\begin{table}[!t]
\centering
\caption{Results on CVBench and additional visual-understanding benchmarks. Higher scores are better.}
\label{tab:cvbench}
\small
\setlength{\tabcolsep}{5pt}
\definecolor{cvbenchhighlight}{RGB}{40,144,208}
\newcommand{\cvbenchbest}[1]{\cellcolor{cvbenchhighlight!18}\textbf{#1}}
\newcommand{\cvbenchgain}[1]{\textcolor[RGB]{40,144,208}{\textbf{#1}}}
\resizebox{\textwidth}{!}{%
\begin{tabular}{l c c c c c c c c c}
\toprule
Model & \multicolumn{3}{c}{CVBench 2D} & \multicolumn{3}{c}{CVBench 3D} & VisPuzzle & MM-Vet & MMVP \\
\cmidrule(lr){2-4}\cmidrule(lr){5-7}\cmidrule(lr){8-10}
 & Count & Relation & Accuracy & Depth & Distance & Accuracy & Score & Score & Score \\
\midrule
Bagel & \cvbenchbest{71.83} & 86.77 & 78.01 & 91.50 & 70.67 & 81.08 & 34.75 & 63.30 & 69.30 \\
STBridge-SFT & 70.18 & 88.92 & 77.98 & 91.17 & 72.33 & 81.75 & 35.25 & 63.90 & 71.33 \\
STBridge-RL & 70.56 & \cvbenchbest{91.08} & \cvbenchbest{79.20} & \cvbenchbest{93.00} & \cvbenchbest{73.00} & \cvbenchbest{83.00} & \cvbenchbest{35.50} & \cvbenchbest{64.60} & \cvbenchbest{72.00} \\
\midrule
Gain & -1.27 & \cvbenchgain{+4.31} & \cvbenchgain{+1.19} & \cvbenchgain{+1.50} & \cvbenchgain{+2.33} & \cvbenchgain{+1.92} & \cvbenchgain{+0.75} & \cvbenchgain{+1.30} & \cvbenchgain{+2.70} \\
\bottomrule
\end{tabular}%
}
\end{table}

\begin{table}[!t]
\centering
\caption{Results on the OddGridBench benchmark. Higher scores are better.}
\label{tab:oddgridbench}
\small
\setlength{\tabcolsep}{8pt}
\definecolor{oddgridhighlight}{RGB}{40,144,208}
\newcommand{\oddgridbest}[1]{\cellcolor{oddgridhighlight!18}\textbf{#1}}
\newcommand{\oddgridgain}[1]{\textcolor[RGB]{40,144,208}{\textbf{#1}}}
\begin{tabular*}{\textwidth}{@{\extracolsep{\fill}}l c c c c c}
\toprule
Model & Color & Position & Rotation & Size & All \\
\midrule
Bagel & 65.60 & 40.20 & 44.70 & 40.10 & 36.57 \\
STBridge-SFT & 69.60 & 42.90 & 48.10 & 43.10 & 39.21 \\
STBridge-RL & \oddgridbest{73.01} & \oddgridbest{45.27} & \oddgridbest{50.85} & \oddgridbest{46.18} & \oddgridbest{41.71} \\
\midrule
Gain & \oddgridgain{+7.41} & \oddgridgain{+5.07} & \oddgridgain{+6.15} & \oddgridgain{+6.08} & \oddgridgain{+5.14} \\
\bottomrule
\end{tabular*}
\end{table}

\paragraph{Image generation.} Tab.~\ref{tab:wise} evaluates whether generation can follow world knowledge--informed prompts. Following the official WISE evaluation configuration, we deploy Qwen3.5-35B-A3B~\cite{Qwen_2025_Qwen3} with vLLM as the evaluation judge. Compared with BAGEL, STBridge-SFT improves the overall WISE score from 62.80\% to 63.30\%, and STBridge-RL further improves it to 64.00\%. The gains are especially notable on Space and Biology, where STBridge-RL improves over BAGEL by +10.33 and +21.50, respectively, with additional improvements on Physics (+5.00) and Chemistry (+2.17). These results are consistent with the motivation of STBridge. WISE differs from generic text-to-image evaluation because many prompts require the model to interpret knowledge, relation, or spatial constraints before composing the target image. The improvements on Space, Biology, Physics, and Chemistry suggest that shared-target alignment may be helpful when generation depends on structured visual-semantic interpretation.

\begin{table}[!htbp]
\centering
\caption{Results on the WISE benchmark (\%). Higher scores are better.}
\label{tab:wise}
\small
\setlength{\tabcolsep}{8pt}
\definecolor{wisehighlight}{RGB}{40,144,208}
\newcommand{\wisebest}[1]{\cellcolor{wisehighlight!18}\textbf{#1}}
\newcommand{\wisegain}[1]{\textcolor[RGB]{40,144,208}{\textbf{#1}}}
\begin{tabular*}{\textwidth}{@{\extracolsep{\fill}}l c c c c c c c}
\toprule
Model & Cultural & Time & Space & Biology & Physics & Chemistry & Overall \\
\midrule
BAGEL & \wisebest{78.00} & \wisebest{63.33} & 56.67 & 37.50 & 55.00 & 50.83 & 62.80 \\
STBridge-SFT & 68.00 & 63.33 & 65.85 & 55.70 & 59.63 & 51.75 & 63.30 \\
STBridge-RL & 69.00 & 63.00 & \wisebest{67.00} & \wisebest{59.00} & \wisebest{60.00} & \wisebest{53.00} & \wisebest{64.00} \\
\midrule
Gain & -9.00 & -0.33 & \wisegain{+10.33} & \wisegain{+21.50} & \wisegain{+5.00} & \wisegain{+2.17} & \wisegain{+1.20} \\
\bottomrule
\end{tabular*}
\end{table}

\begin{table}[!htbp]
\centering
\caption{Results on the ImgEdit benchmark. Higher scores are better.}
\label{tab:imgedit}
\definecolor{imgedithighlight}{RGB}{40,144,208}
\newcommand{\imgeditbest}[1]{\cellcolor{imgedithighlight!18}\textbf{#1}}
\newcommand{\imgeditgain}[1]{\textcolor[RGB]{40,144,208}{\textbf{#1}}}
\resizebox{\textwidth}{!}{
\begin{tabular}{l c c c c c c c c c c}
\toprule
Model & Extract & Replace & Remove & Adjust & Style & Background & Add & Compose & Action & Overall \\
\midrule
Bagel & 1.58 & 3.85 & 3.16 & 3.59 & \imgeditbest{4.51} & 3.39 & 3.81 & \imgeditbest{2.67} & \imgeditbest{4.25} & 3.42 \\
STBridge-SFT & 2.09 & 3.88 & 3.69 & \imgeditbest{3.83} & 3.88 & 3.83 & 3.92 & 2.48 & 4.12 & 3.57 \\
STBridge-RL & \imgeditbest{3.34} & \imgeditbest{4.19} & \imgeditbest{3.96} & 3.76 & 4.39 & \imgeditbest{3.97} & \imgeditbest{3.97} & 2.36 & 4.10 & \imgeditbest{3.88} \\
\midrule
Gain & \imgeditgain{+1.76} & \imgeditgain{+0.34} & \imgeditgain{+0.80} & \imgeditgain{+0.17} & -0.12 & \imgeditgain{+0.58} & \imgeditgain{+0.16} & -0.31 & -0.15 & \imgeditgain{+0.46} \\
\bottomrule
\end{tabular}
}
\end{table}

\begin{table}[!htbp]
\centering
\caption{Results on the RISE benchmark. Higher scores are better.}
\label{tab:rise}
\small
\setlength{\tabcolsep}{6pt}
\newcommand{\risegain}[1]{\textcolor[RGB]{40,144,208}{\textbf{#1}}}
\begin{tabular*}{\textwidth}{@{\extracolsep{\fill}}>{\raggedright\arraybackslash}p{0.34\textwidth} c c c c c}
\toprule
Model & Temporal (\%) & Causal (\%) & Spatial (\%) & Logical (\%) & Overall (\%) \\
\midrule
GPT-Image-1.5 & 57.60 & 62.20 & 62.00 & 21.20 & 51.40 \\
GPT-Image-2 & 45.90 & 66.70 & 50.00 & 34.10 & 49.40 \\
Nano Banana-pro & 43.50 & 63.30 & 48.00 & 37.60 & 48.30 \\
Gemini-2.5-Flash-Image (Nano Banana) & 29.40 & 48.90 & 37.00 & 18.80 & 33.90 \\
GPT-Image-1 & 36.50 & 34.40 & 37.00 & 10.60 & 30.00 \\
GPT-Image-1-mini & 25.90 & 31.10 & 33.00 & 9.40 & 25.30 \\
Qwen-Image-Edit-2511 & 21.20 & 18.90 & 31.00 & 4.70 & 19.40 \\
\hdashline
Bagel & 11.80 & 27.80 & 21.00 & 1.20 & 15.80 \\
STBridge-SFT & 37.00 & 43.20 & 49.00 & 1.10 & 32.90 \\
STBridge-RL & 41.00 & 47.60 & 53.60 & 1.90 & 36.80 \\
\midrule
Gain & \risegain{+29.20} & \risegain{+19.80} & \risegain{+32.60} & \risegain{+0.70} & \risegain{+21.00} \\
\bottomrule
\end{tabular*}
\end{table}

\FloatBarrier

\paragraph{Image editing.}
Image editing provides a direct evaluation setting for examining editing performance under source-instruction constraints. As shown in Tabs.~\ref{tab:imgedit} and \ref{tab:rise}, STBridge improves over BAGEL on both standard and knowledge-guided editing settings. On ImgEdit, STBridge-SFT raises the overall score from 3.42 to 3.57, and STBridge-RL further improves it to 3.88.

On RISE, where the task requires more complex knowledge-guided editing, STBridge-SFT improves the overall score from 15.80\% to 32.90\%, and STBridge-RL further reaches 36.80\%. These results further suggest the effectiveness of STBridge: by aligning target expression and visual realization through a shared target, the model improves not only standard image editing performance, but also more challenging cases that require interpreting structured editing constraints before producing the final edited image.

\subsection{Qualitative Results}

\begin{figure}[p]
\centering
\begingroup
\vspace*{-0.82in}
\setlength{\abovecaptionskip}{2pt}
\setlength{\belowcaptionskip}{0pt}
\newcommand{\understandpanelheight}{0.176\textheight}
\newcommand{\understandpromptheight}{0.021\textheight}
\newcommand{\understandpromptfont}{\fontsize{6.9pt}{6.4pt}\selectfont}
\newcommand{\understandtextfont}{\fontsize{5.8pt}{6.15pt}\selectfont}
\definecolor{understandWrongRed}{RGB}{198,69,65}
\definecolor{understandGoodGreen}{RGB}{63,208,170}
\newcommand{\understandgoodface}{\raisebox{-0.15ex}{\tikz[scale=0.115,baseline=-0.55ex]{\fill[understandGoodGreen!16] (0,0) circle (1); \draw[understandGoodGreen,line width=0.45pt] (0,0) circle (1); \fill[understandGoodGreen] (-0.35,0.28) circle (0.10); \fill[understandGoodGreen] (0.35,0.28) circle (0.10); \draw[understandGoodGreen,line width=0.45pt] (-0.48,-0.20) .. controls (0,-0.62) .. (0.48,-0.20);}}}
\newcommand{\understandbadface}{\raisebox{-0.15ex}{\tikz[scale=0.115,baseline=-0.55ex]{\fill[understandWrongRed!14] (0,0) circle (1); \draw[understandWrongRed,line width=0.45pt] (0,0) circle (1); \draw[understandWrongRed,line width=0.45pt] (-0.48,0.35) -- (-0.22,0.15); \draw[understandWrongRed,line width=0.45pt] (-0.22,0.35) -- (-0.48,0.15); \draw[understandWrongRed,line width=0.45pt] (0.22,0.35) -- (0.48,0.15); \draw[understandWrongRed,line width=0.45pt] (0.48,0.35) -- (0.22,0.15); \draw[understandWrongRed,line width=0.45pt] (-0.48,-0.45) .. controls (0,-0.10) .. (0.48,-0.45);}}}
\newcommand{\understandbad}[1]{\understandbadface\,\textcolor{understandWrongRed}{\textbf{#1}}}
\newcommand{\understandours}[1]{\understandgoodface\,\textcolor{understandGoodGreen}{\textbf{#1}}}
\newcommand{\understandcompareentry}[7]{%
\begin{minipage}[t]{0.497\linewidth}
\begin{minipage}[t][\understandpromptheight][c]{\linewidth}
\centering\understandpromptfont\textbf{Question:} \emph{#1}
\end{minipage}\par\vspace{0.8pt}
\begin{minipage}[t]{0.505\linewidth}
\vspace{0pt}
\centering
\includegraphics[width=\linewidth,height=\understandpanelheight,keepaspectratio]{#2}
\end{minipage}\hfill
\begin{minipage}[t]{0.475\linewidth}
\vspace{0pt}
\begin{tcolorbox}[colback=youtuLightBlue,colframe=youtuBlue!25,boxrule=0.25pt,arc=1.2pt,left=1.4pt,right=1.4pt,top=1.1pt,bottom=1.1pt,width=\linewidth,fit to height=\understandpanelheight,fit basedim=5.55pt,fit skip=0.82,fit algorithm=fontsize]
{\raggedright\understandtextfont\setlength{\parindent}{0pt}\setlength{\parskip}{0pt}%
\textbf{Task.} #3\par\vspace{0.8pt}%
\textbf{Choices.} #4\par\vspace{0.8pt}%
\textbf{GT:} \understandours{#5}\par\vspace{0.8pt}%
\textbf{BAGEL:} \understandbad{#6}\par\vspace{0.8pt}%
\textbf{Ours:} \understandours{#7}\par}
\end{tcolorbox}
\end{minipage}
\end{minipage}}
\makebox[\textwidth][c]{%
\begin{minipage}[t]{1.12\textwidth}
\noindent
\understandcompareentry
{Depth: which object is closer to the camera?}
{}
{Compare the red-box shelves with the blue-box mirror.}
{A. shelves; B. mirror}
{B}
{A}
{B}
\hfill
\understandcompareentry
{Distance: which object is closer to the books?}
{}
{Compare the blue-box door and green-box lamp relative to the red-box books.}
{A. door; B. lamp}
{B}
{A}
{B}
\par\vspace{7.0pt}
\noindent
\understandcompareentry
{Relation: where is the potted plant relative to the horse?}
{}
{Infer the relative position of the red-box potted plant with respect to the horse.}
{A. left; B. right}
{B}
{A}
{B}
\hfill
\understandcompareentry
{Jigsaw: recover the shuffled 2$\times$2 image.}
{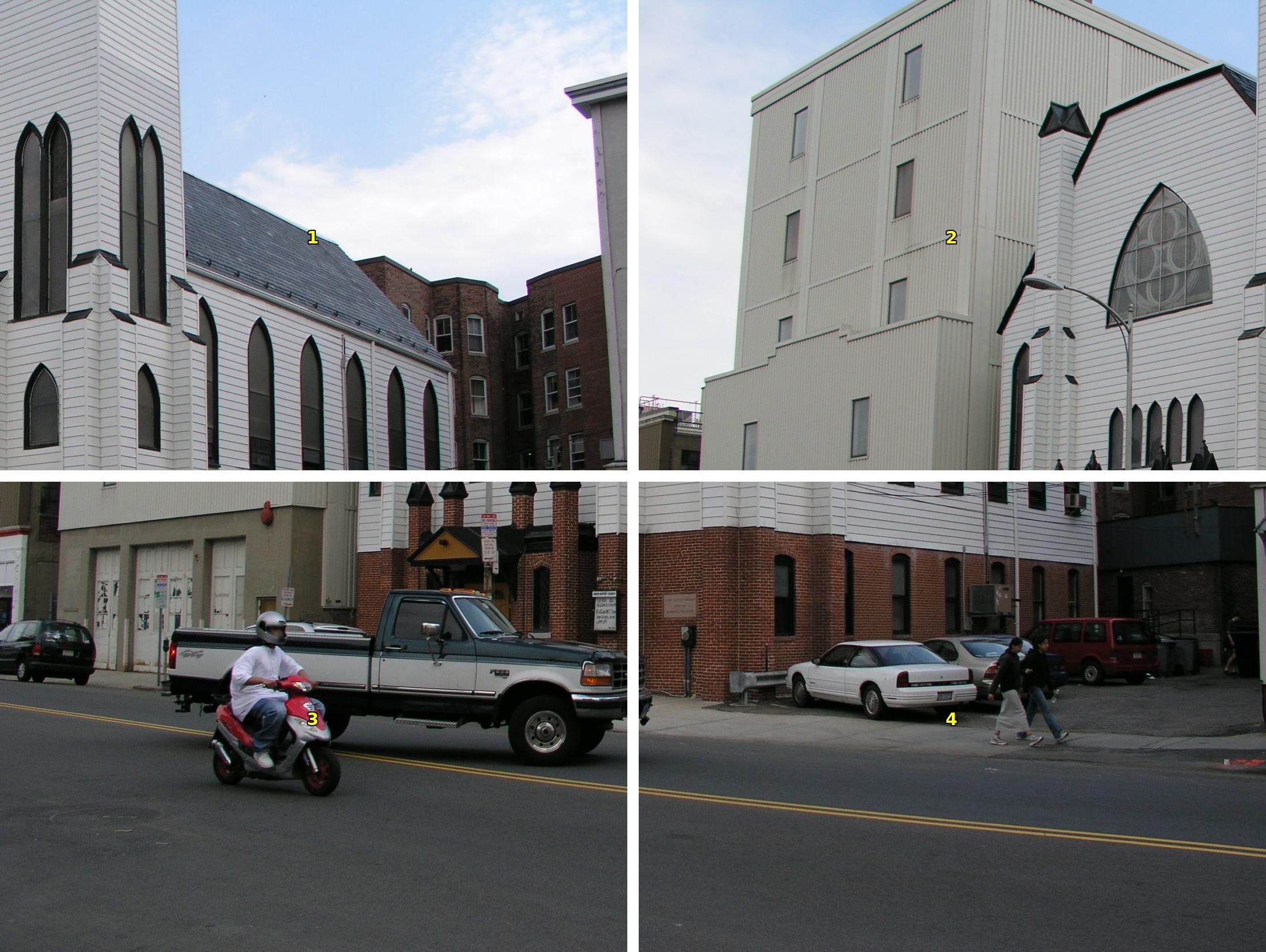}
{Choose the arrangement that reconstructs the natural image.}
{A--D candidate arrangements}
{B}
{C}
{B}
\par\vspace{7.0pt}
\noindent
\understandcompareentry
{OddGrid: find the color outlier.}
{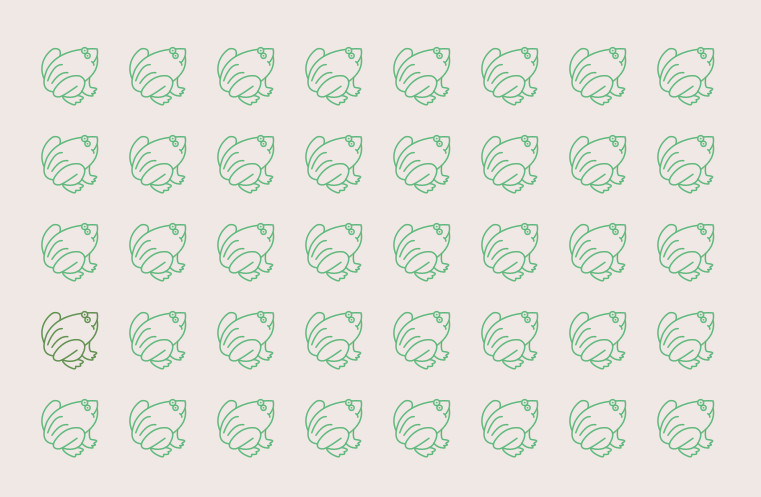}
{Locate the only color outlier in the 5$\times$8 grid.}
{Grid item: frog; oddity: color}
{Row 4, Column 1}
{Row 3, Column 8}
{Row 4, Column 1}
\hfill
\understandcompareentry
{OddGrid: find the rotation outlier.}
{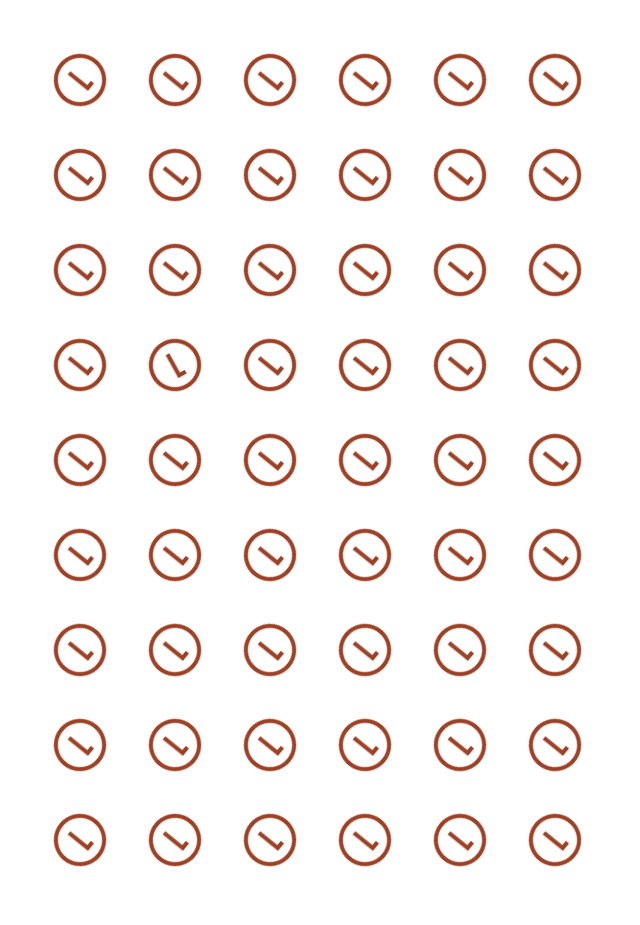}
{Locate the only rotation outlier in the 9$\times$6 grid.}
{Grid item: number 1; oddity: rotation}
{Row 4, Column 2}
{Row 4, Column 3}
{Row 4, Column 2}
\end{minipage}}
\caption{\textbf{Visual-understanding comparisons.} Qualitative BAGEL-vs-Ours comparisons on visual-understanding tasks are arranged in a three-row, two-column grid. Each panel shows the visual input and a compact answer block with the ground-truth answer, the BAGEL baseline prediction, and the Ours prediction. Green smile icons mark correct answers, while red frown icons mark incorrect BAGEL answers. All test images shown above are selected from the open-source benchmarks BLINK~\cite{fu2024blink} and OddGridBench~\cite{weng2026oddgridbench}.}
\label{fig:understand_compare_layout}
\endgroup
\end{figure}

\paragraph{Visual understanding.}
Figs.~\ref{fig:understand_qualitative_layout} and~\ref{fig:understand_compare_layout} visualize the effect of STBridge on visual understanding. The examples cover fine-grained similarity, depth and distance reasoning, spatial relation recognition, jigsaw reconstruction, and OddGrid-style anomaly localization. In these cases, STBridge-RL produces the correct answer while the BAGEL baseline often fails on the key visual distinction, suggesting that the shared-target formulation helps the model focus on answer-defining evidence rather than generic image content.

\begin{figure}[p]
\centering
\begingroup
\vspace*{-1.12in}
\setlength{\abovecaptionskip}{2pt}
\setlength{\belowcaptionskip}{0pt}
\newcommand{\imgeditimageheight}{0.124\textheight}
\newcommand{\imgeditpromptheight}{0.027\textheight}
\newcommand{\imgeditpromptfont}{\fontsize{5.9pt}{6.15pt}\selectfont}
\newcommand{\imgeditlabelfont}{\fontsize{5.8pt}{6.0pt}\selectfont}
\definecolor{imgeditBagelRed}{RGB}{198,69,65}
\definecolor{imgeditOursGreen}{RGB}{63,208,170}
\newcommand{\imgeditmodel}[3]{%
\begin{minipage}[t]{0.318\linewidth}
\centering
{\imgeditlabelfont\textcolor{#3}{\textbf{#1}}}\par\vspace{0.8pt}
\includegraphics[width=\linewidth,height=\imgeditimageheight,keepaspectratio]{#2}
\end{minipage}}
\newcommand{\imgeditmodelsep}{\hfill}
\newcommand{\imgeditpanel}[5]{%
\begin{minipage}[t]{0.486\linewidth}
\begin{minipage}[t][\imgeditpromptheight][c]{\linewidth}
\centering\imgeditpromptfont\textbf{Instruction:} \emph{#1}
\end{minipage}\par\vspace{1.1pt}
\makebox[\linewidth][c]{%
\imgeditmodel{Source}{#2}{black}\imgeditmodelsep
\imgeditmodel{BAGEL}{#3}{imgeditBagelRed}\imgeditmodelsep
\imgeditmodel{Ours}{#4}{imgeditOursGreen}}
\end{minipage}}
\newcommand{\imgeditpanelsep}{\hfill}
\begin{minipage}[t]{\textwidth}
\noindent
\imgeditpanel
{Extract the human figure in the image.}
{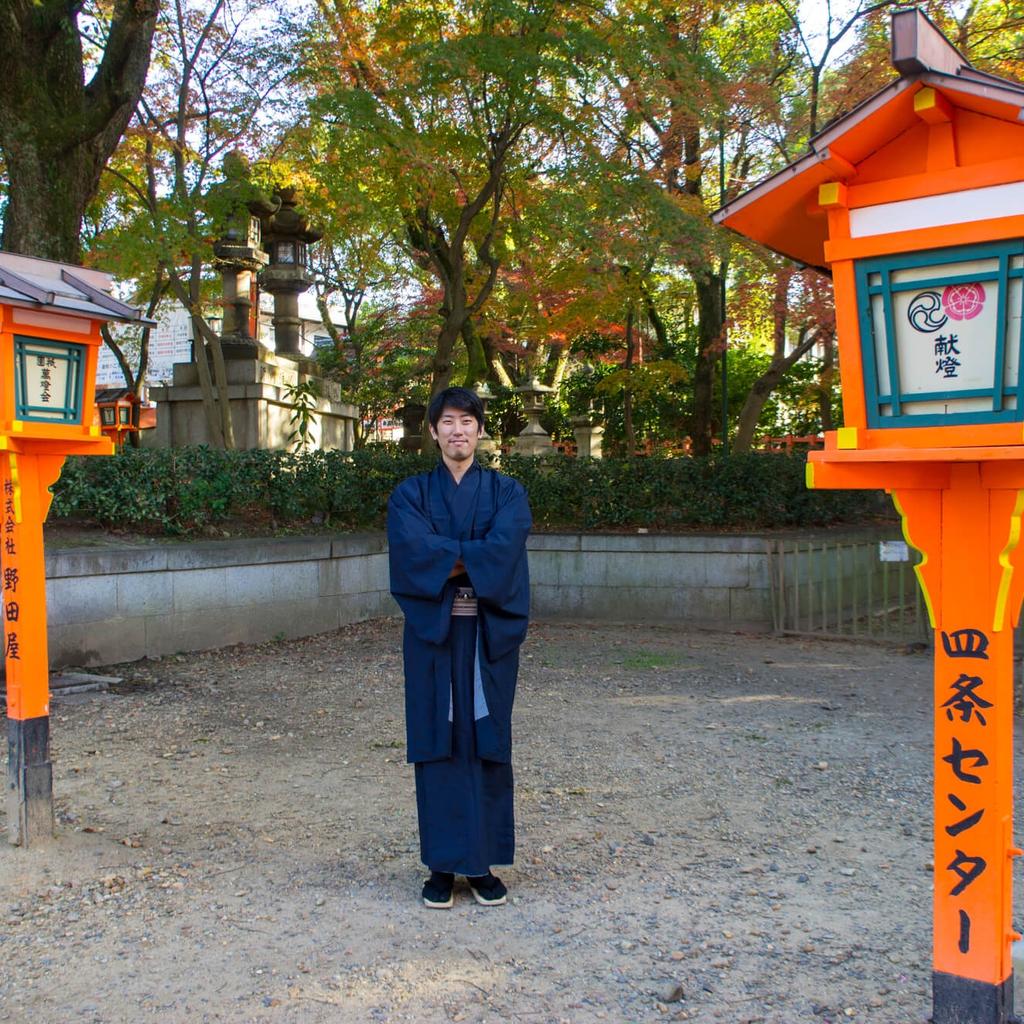}
{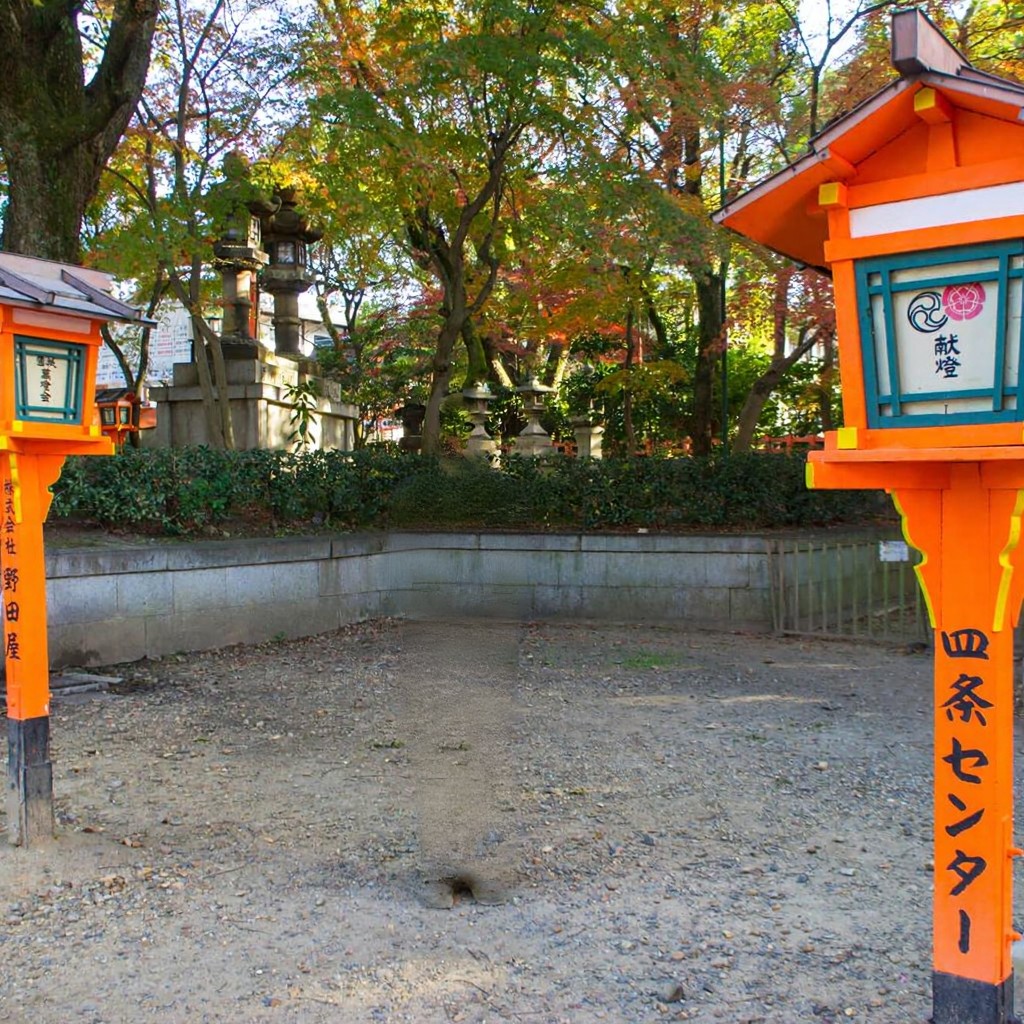}
{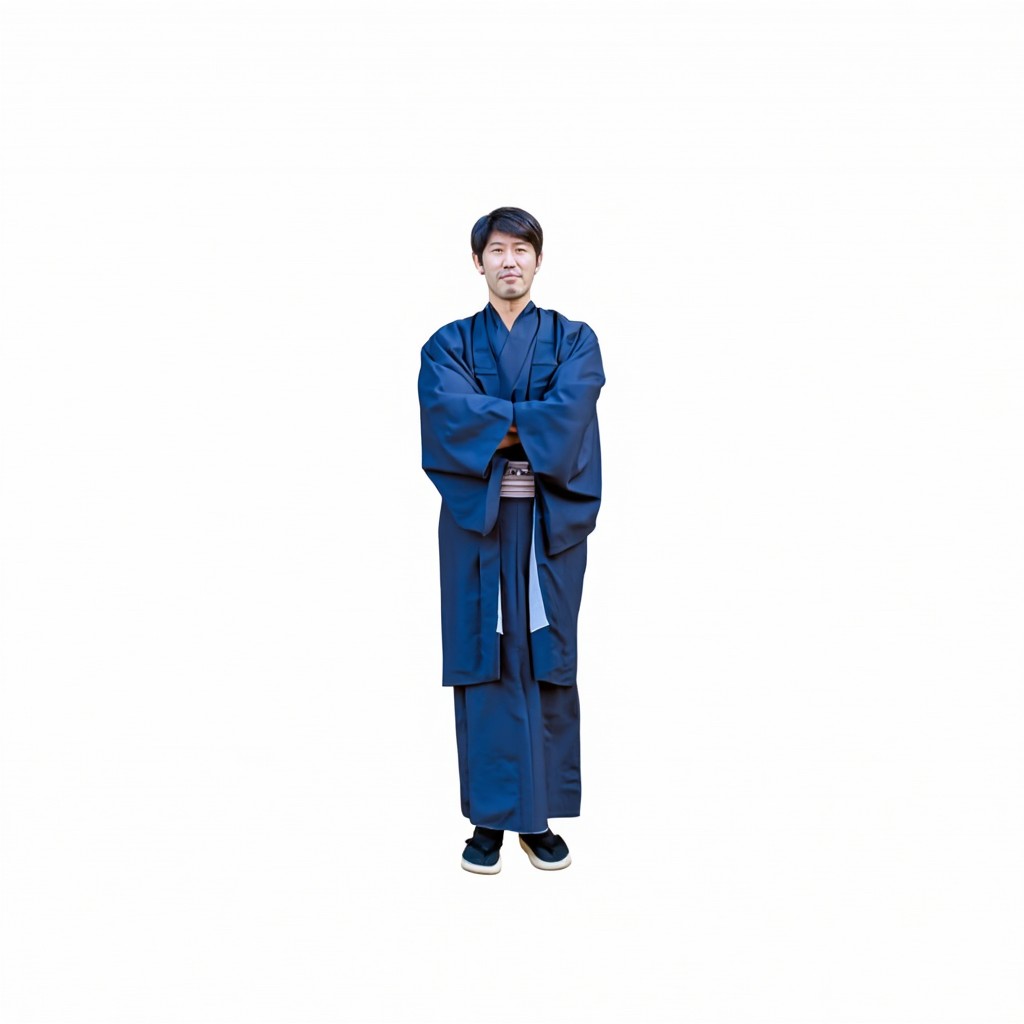}
{}
\imgeditpanelsep
\imgeditpanel
{Replace the dog with a vintage bicycle in the same outdoor setting.}
{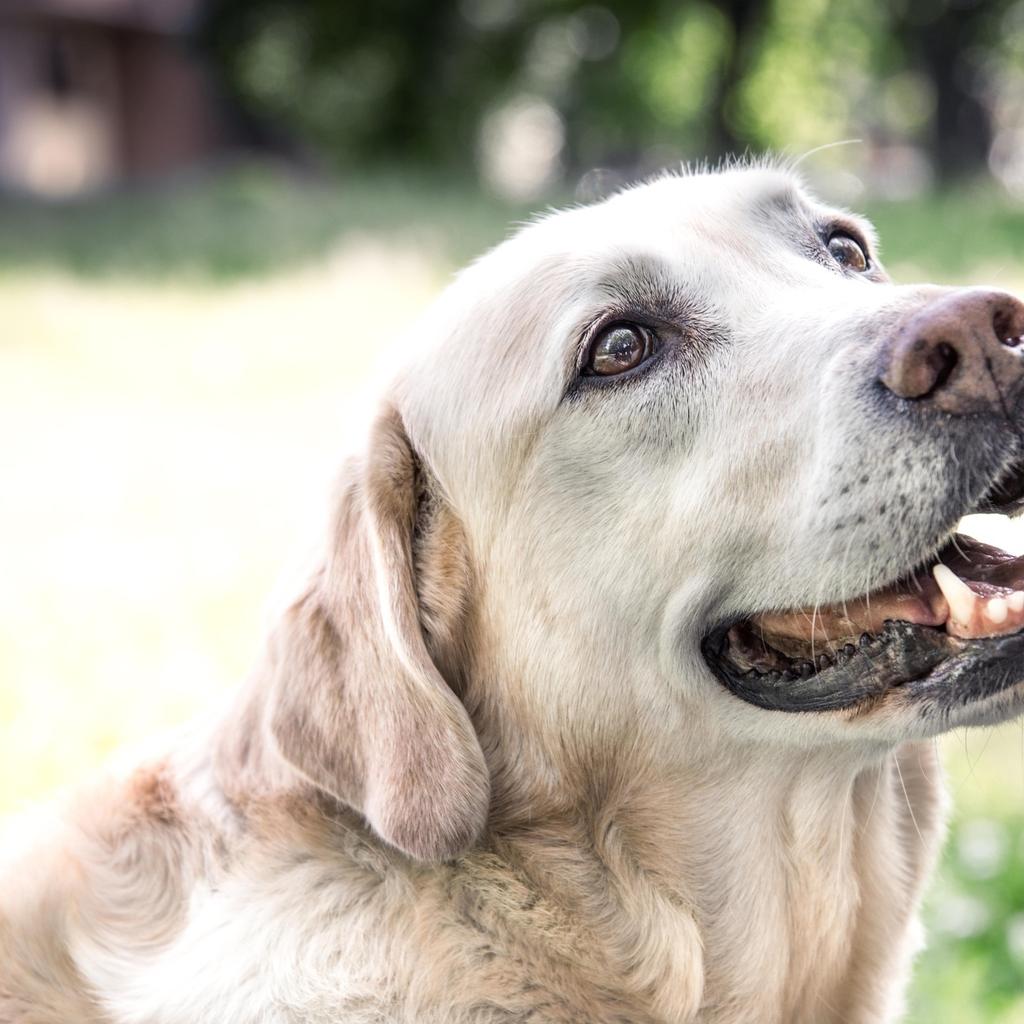}
{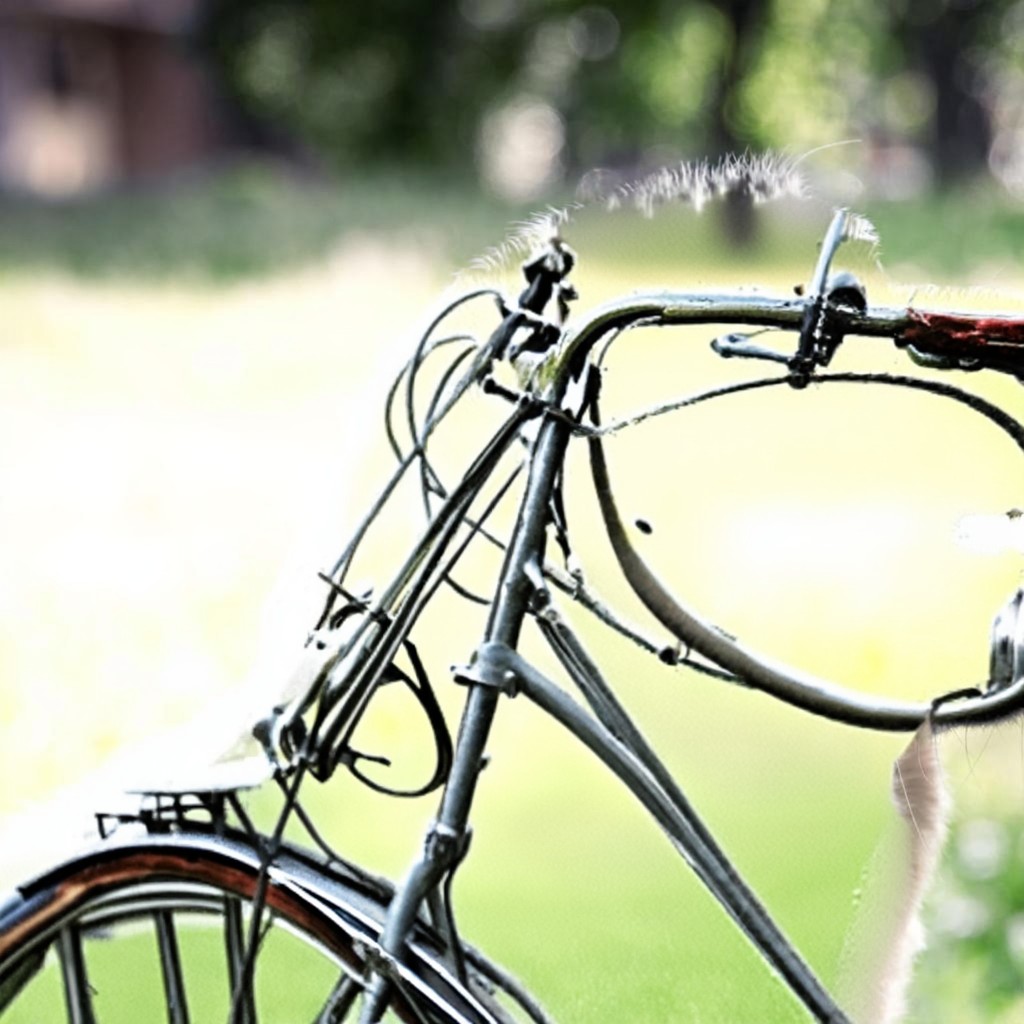}
{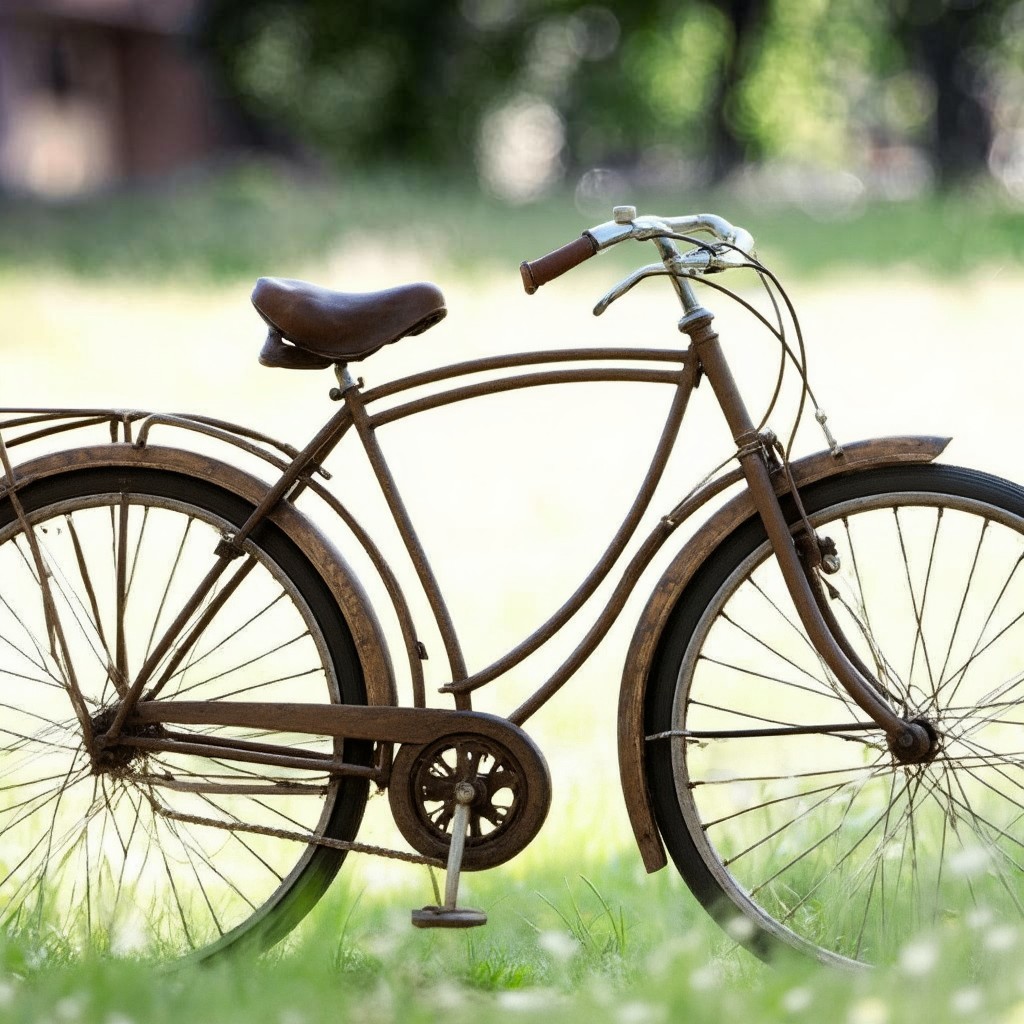}
{}
\par\vspace{5.2pt}
\noindent
\imgeditpanel
{Remove the rainbow and cloud face paint from the image.}
{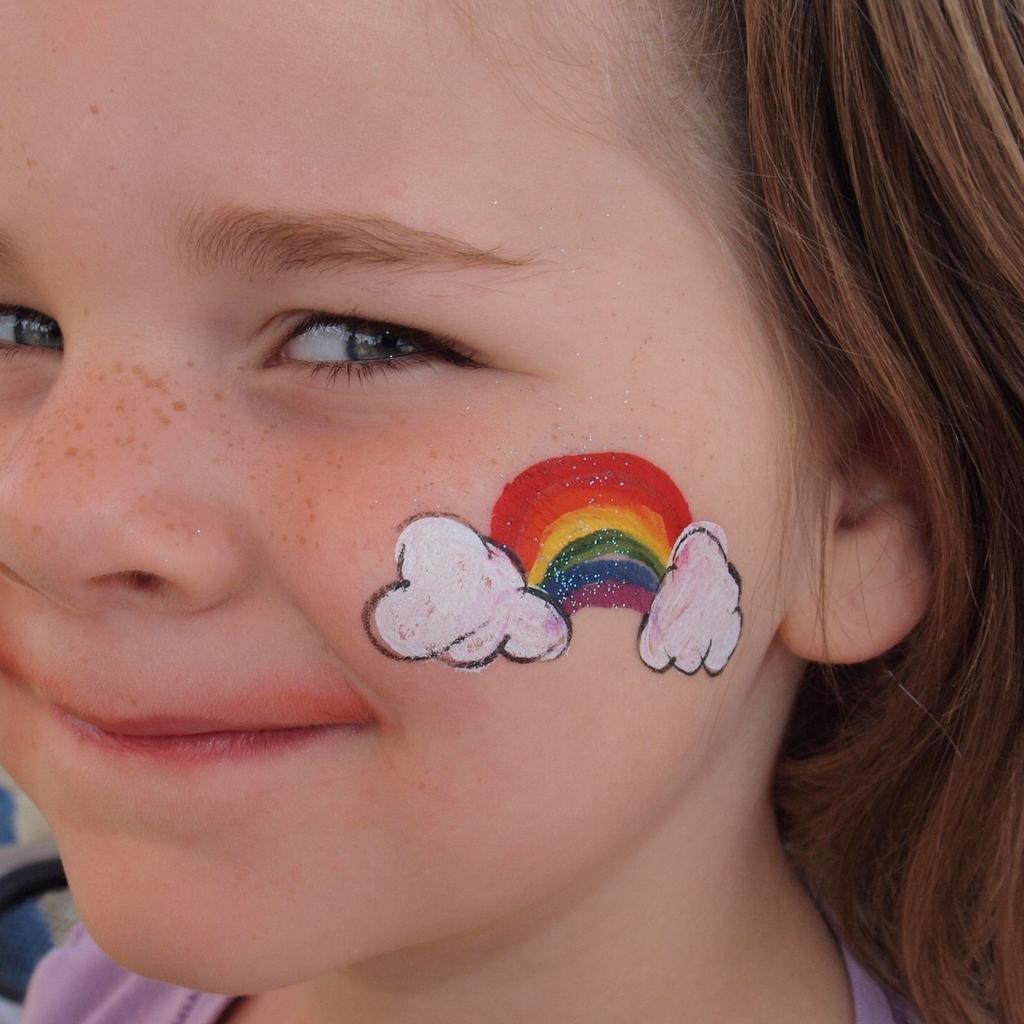}
{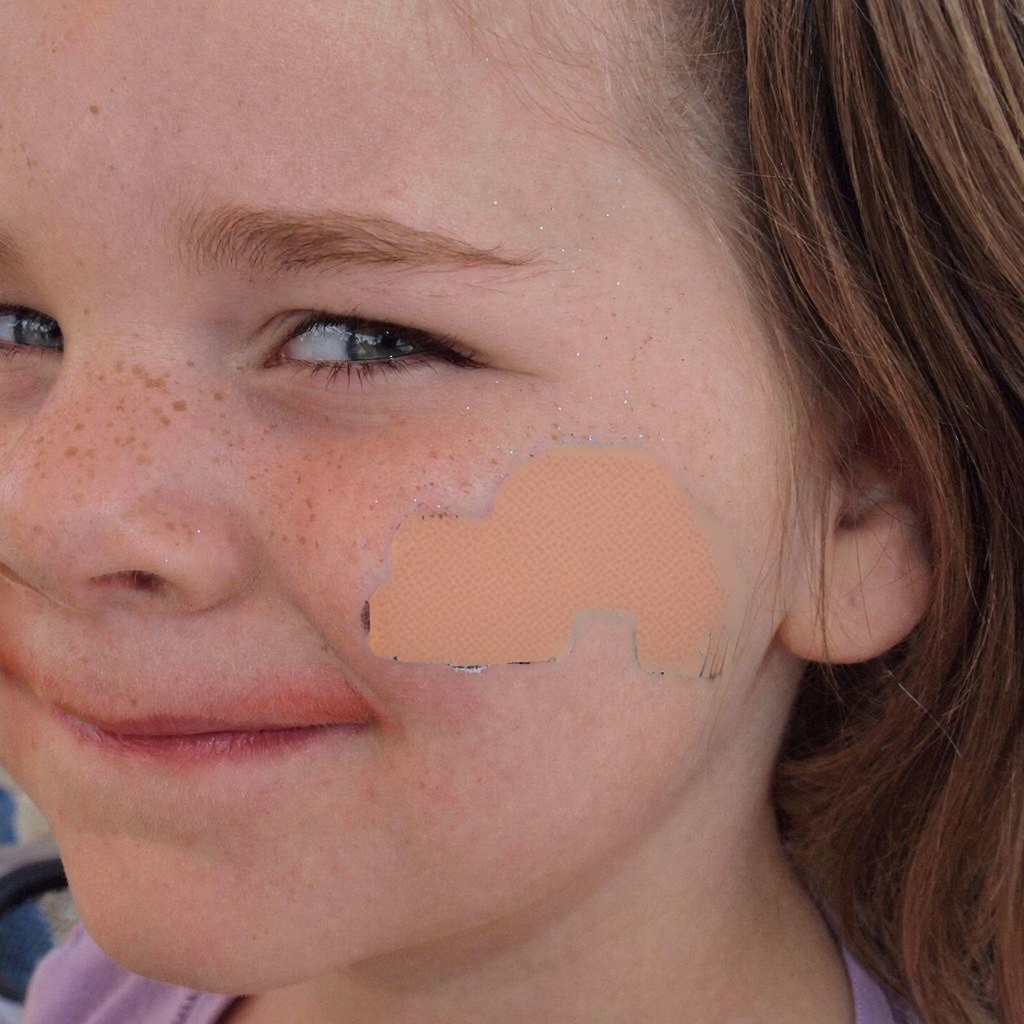}
{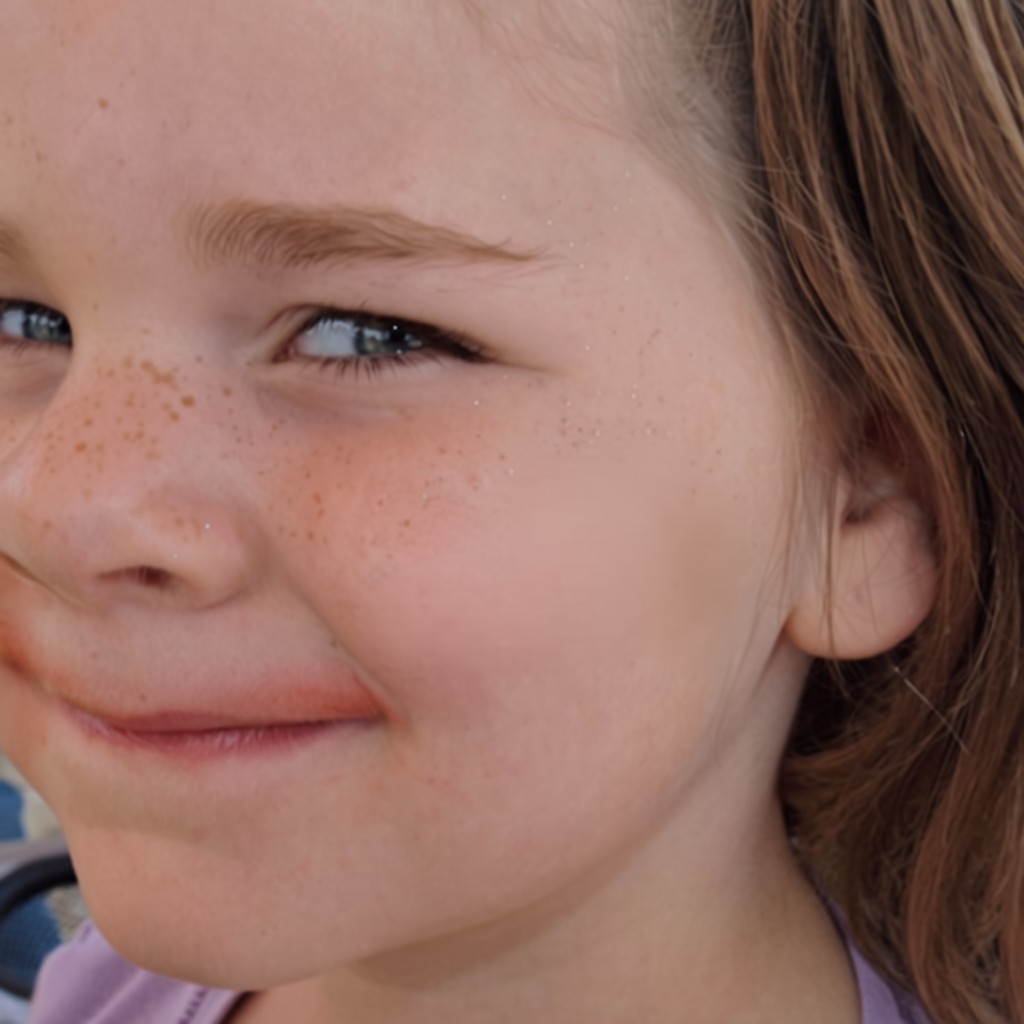}
{}
\imgeditpanelsep
\imgeditpanel
{Change the priest's coat color to dark brown.}
{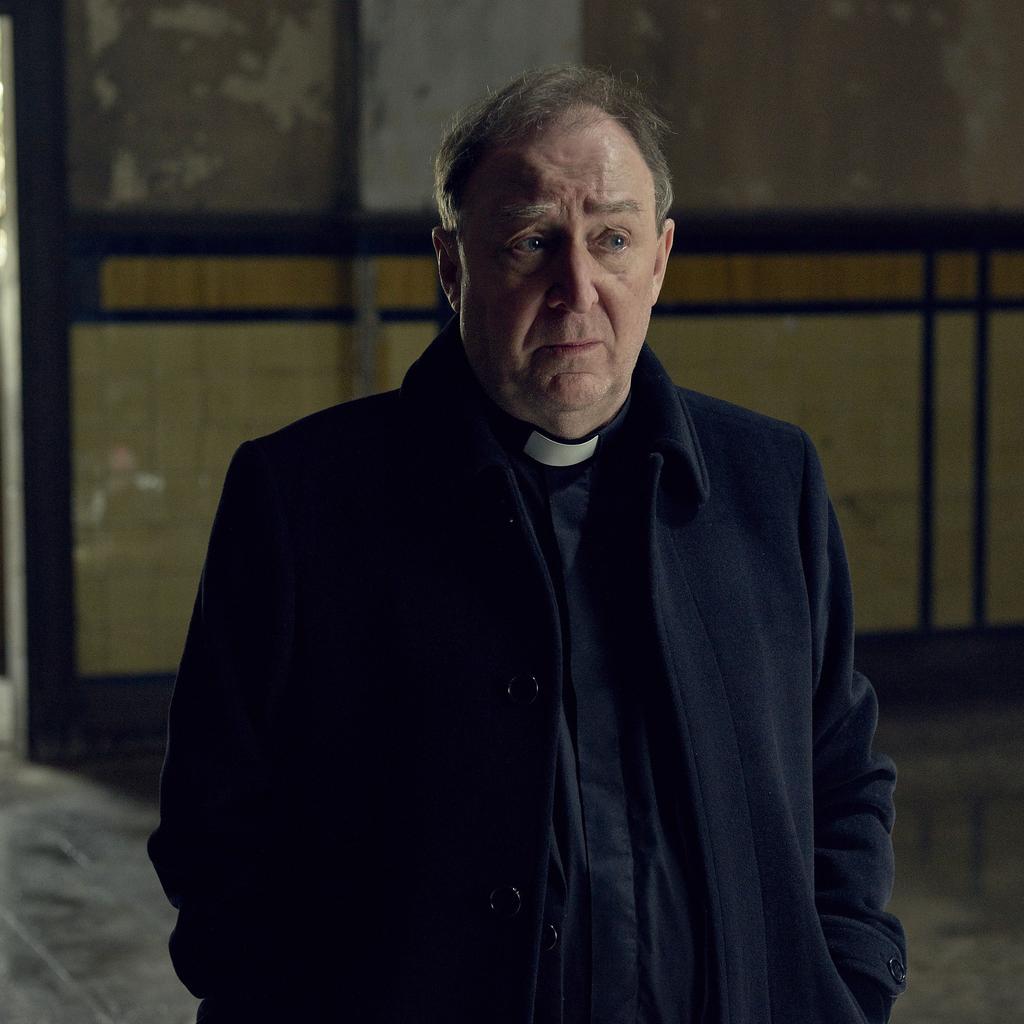}
{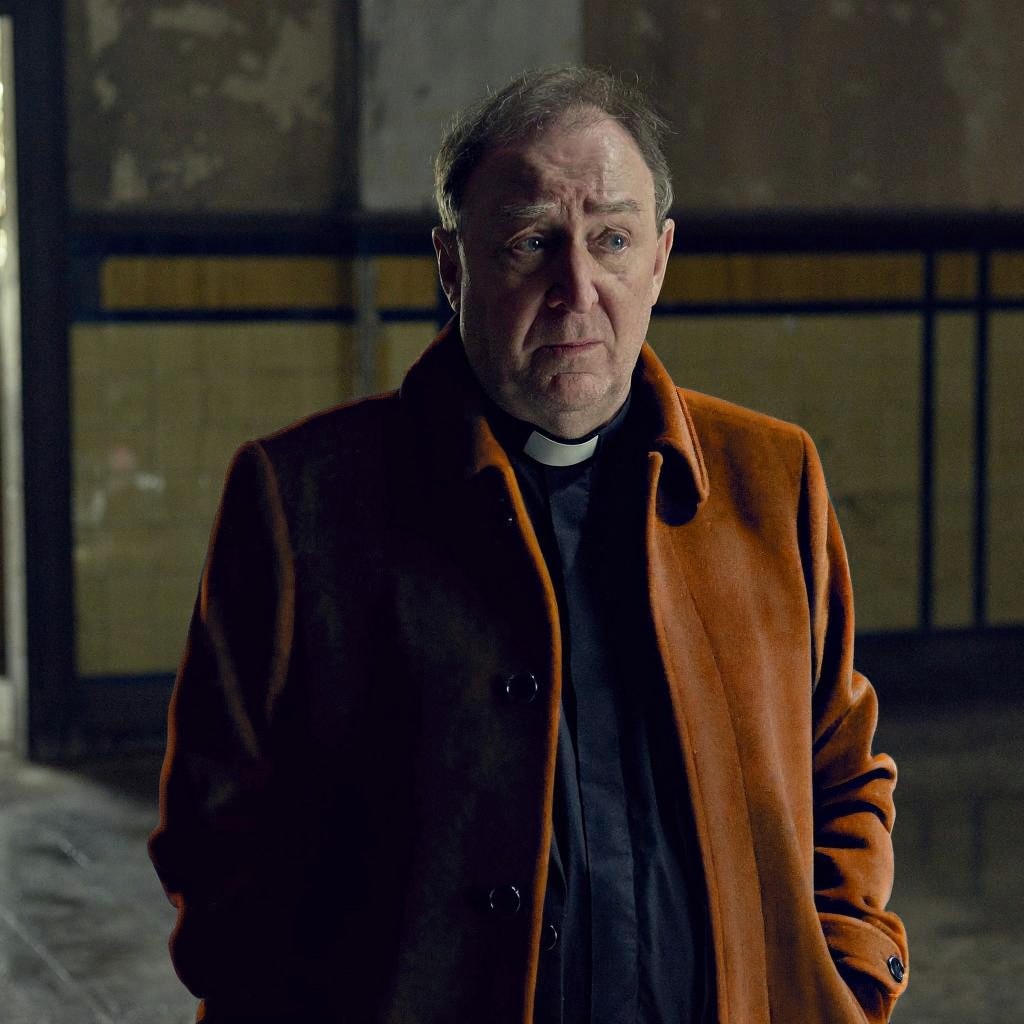}
{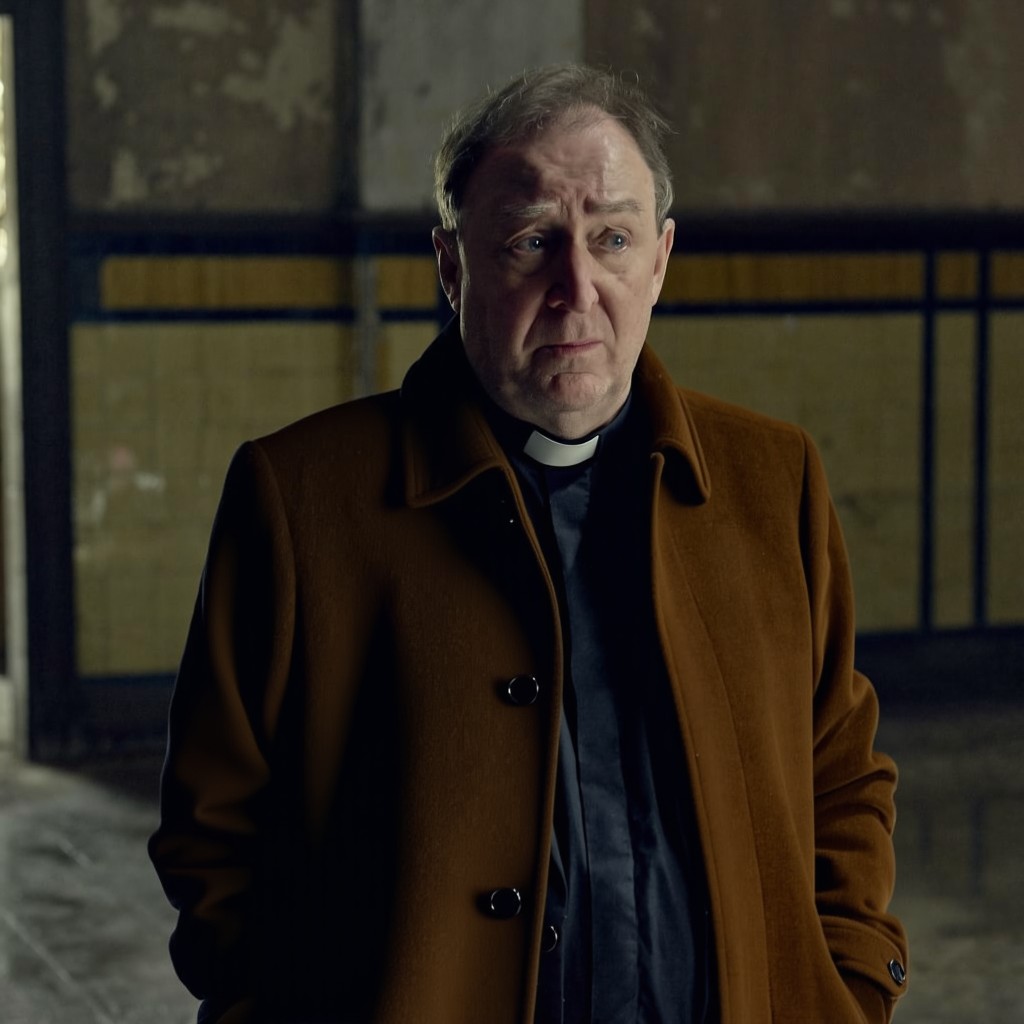}
{}
\par\vspace{5.2pt}
\noindent
\imgeditpanel
{Transfer the image into a Lego-brick stop-motion diorama style.}
{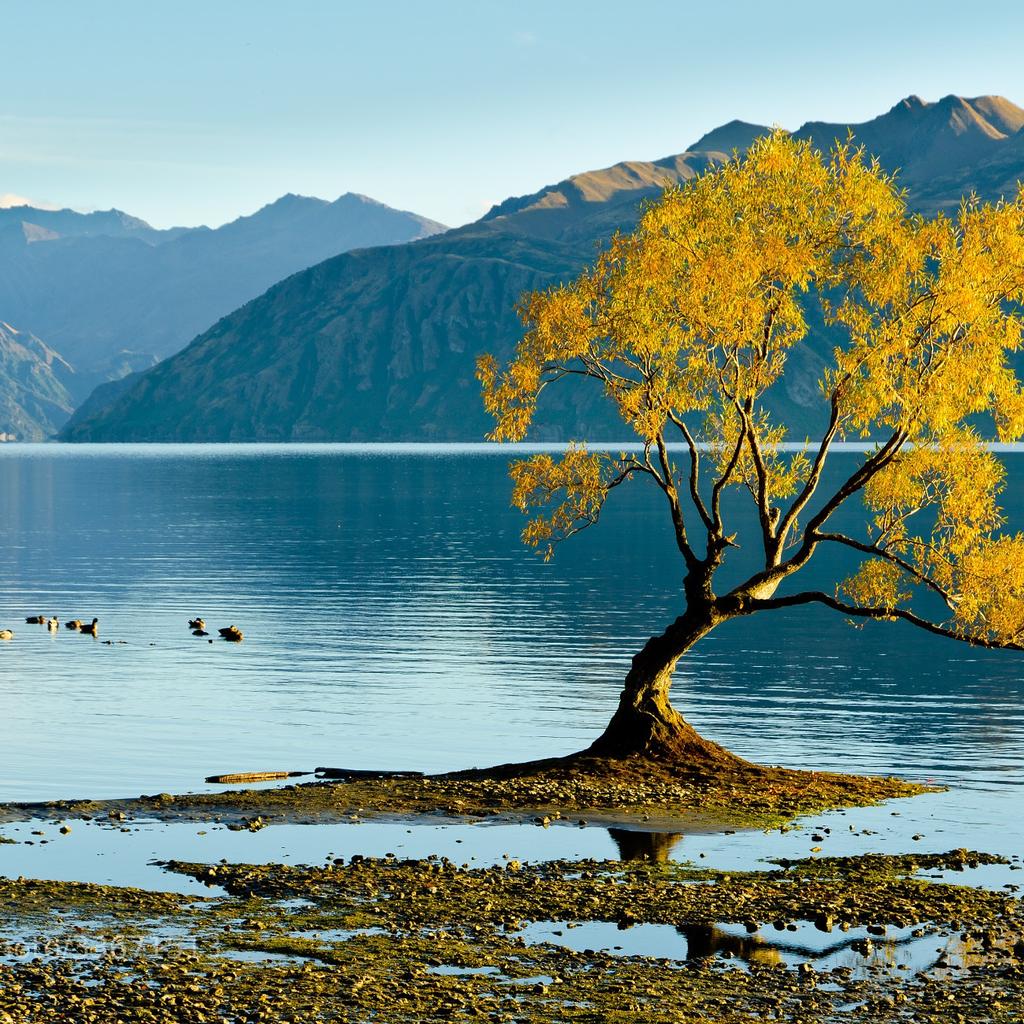}
{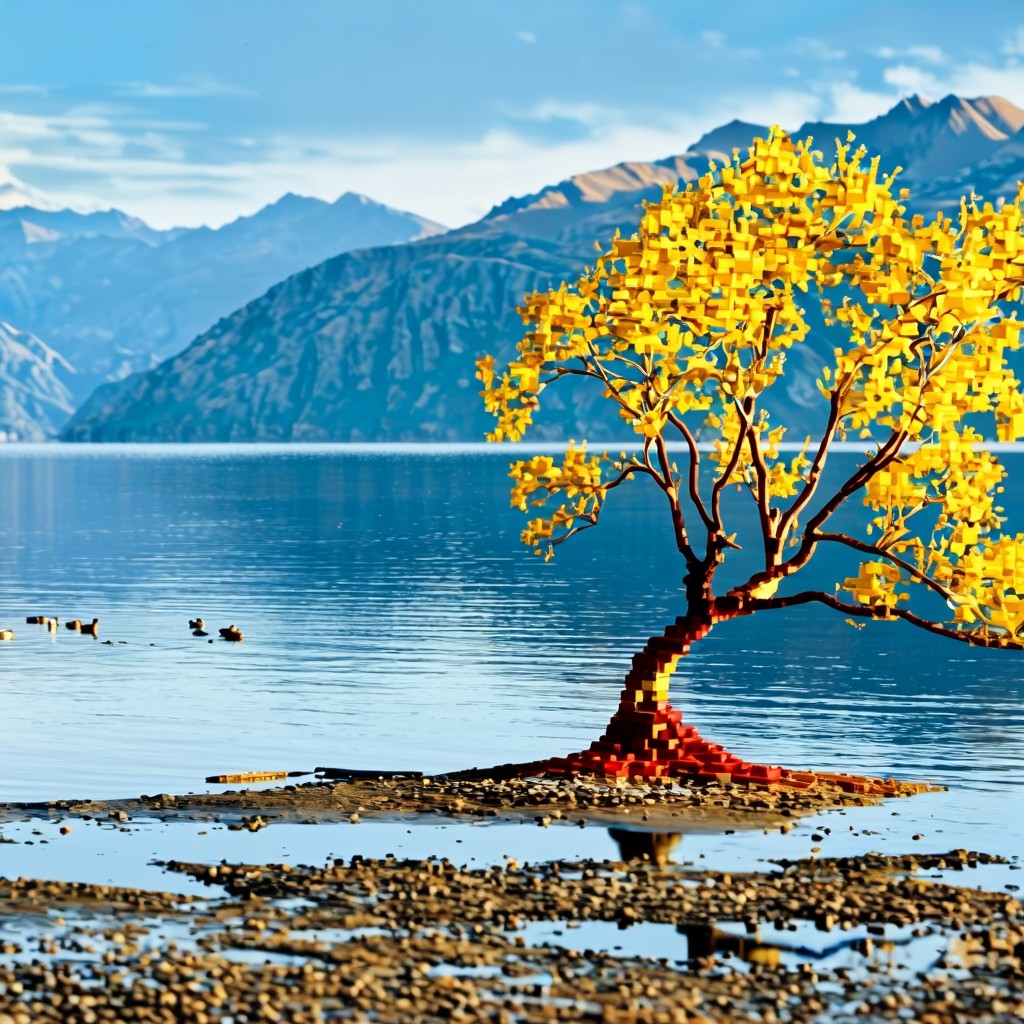}
{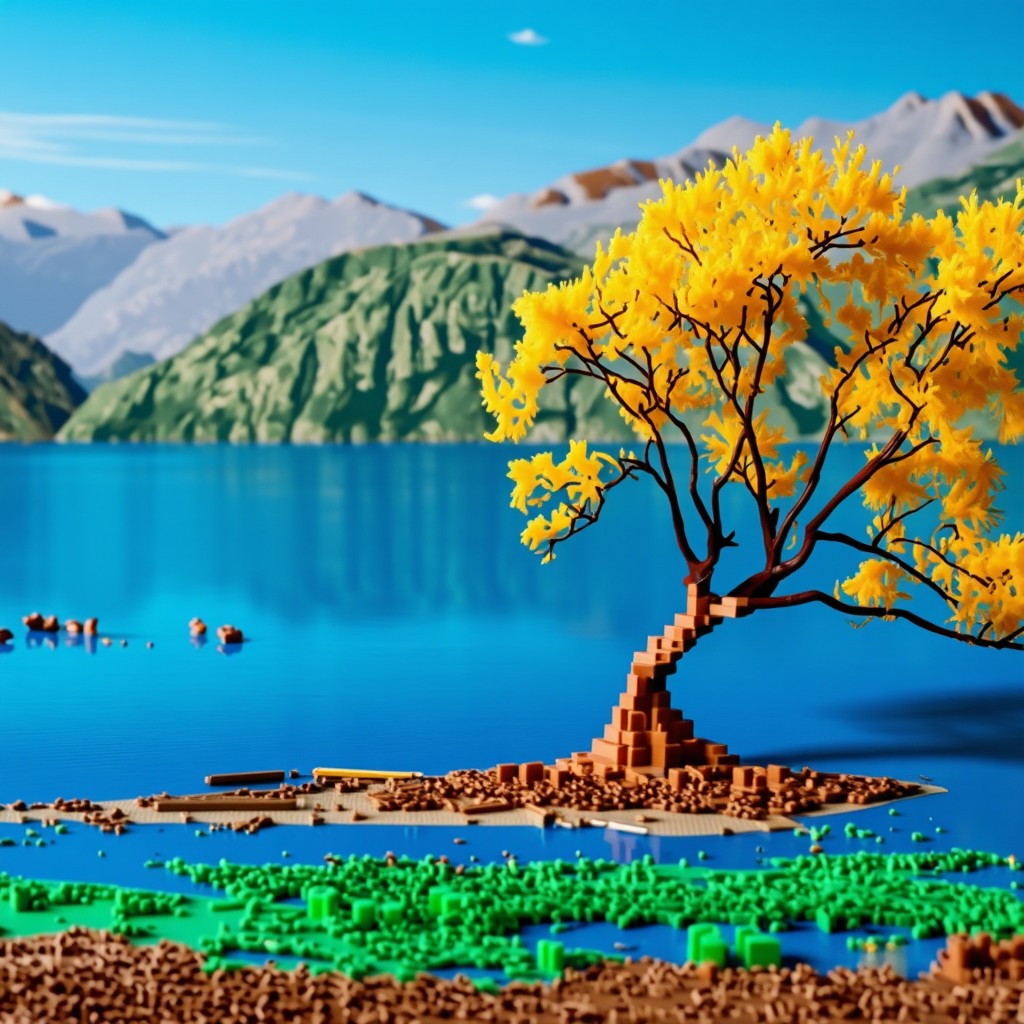}
{}
\imgeditpanelsep
\imgeditpanel
{Change the trees to palm trees and make the ground a sandy beach.}
{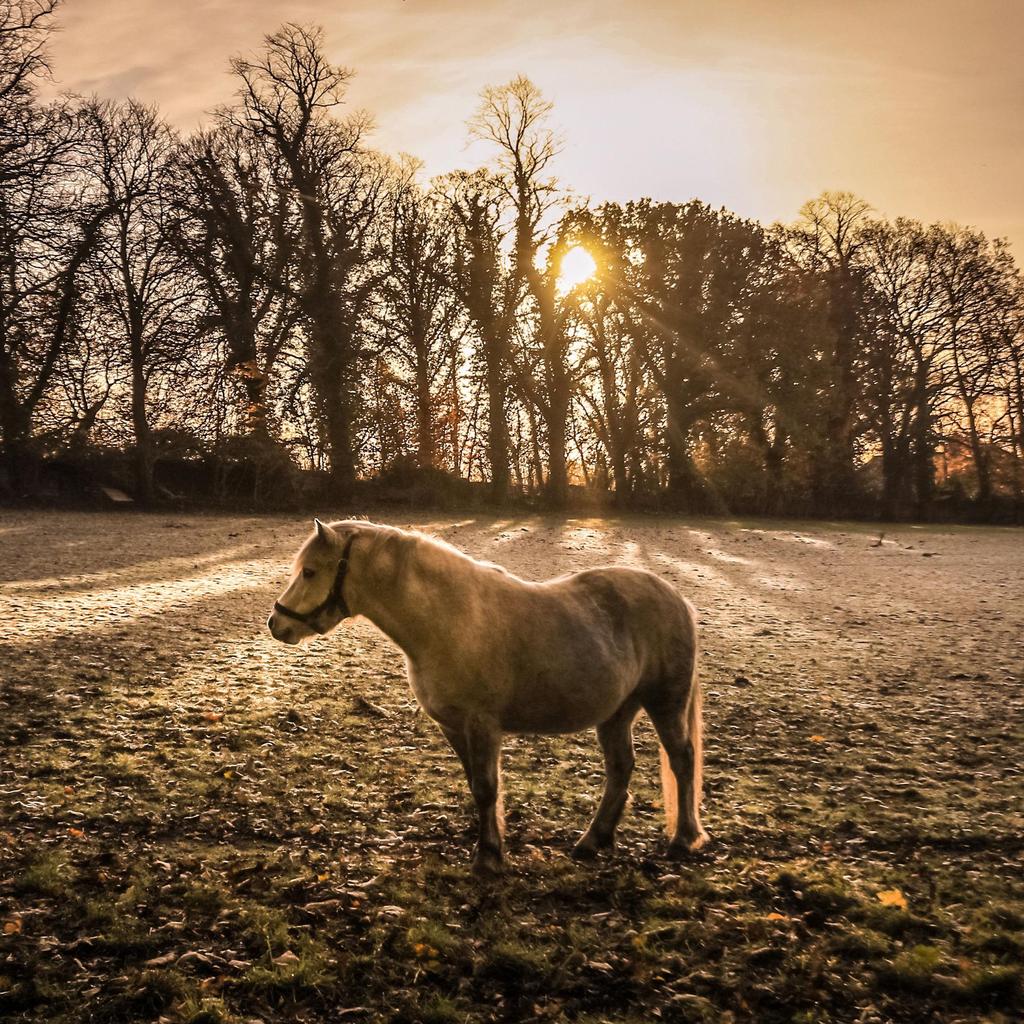}
{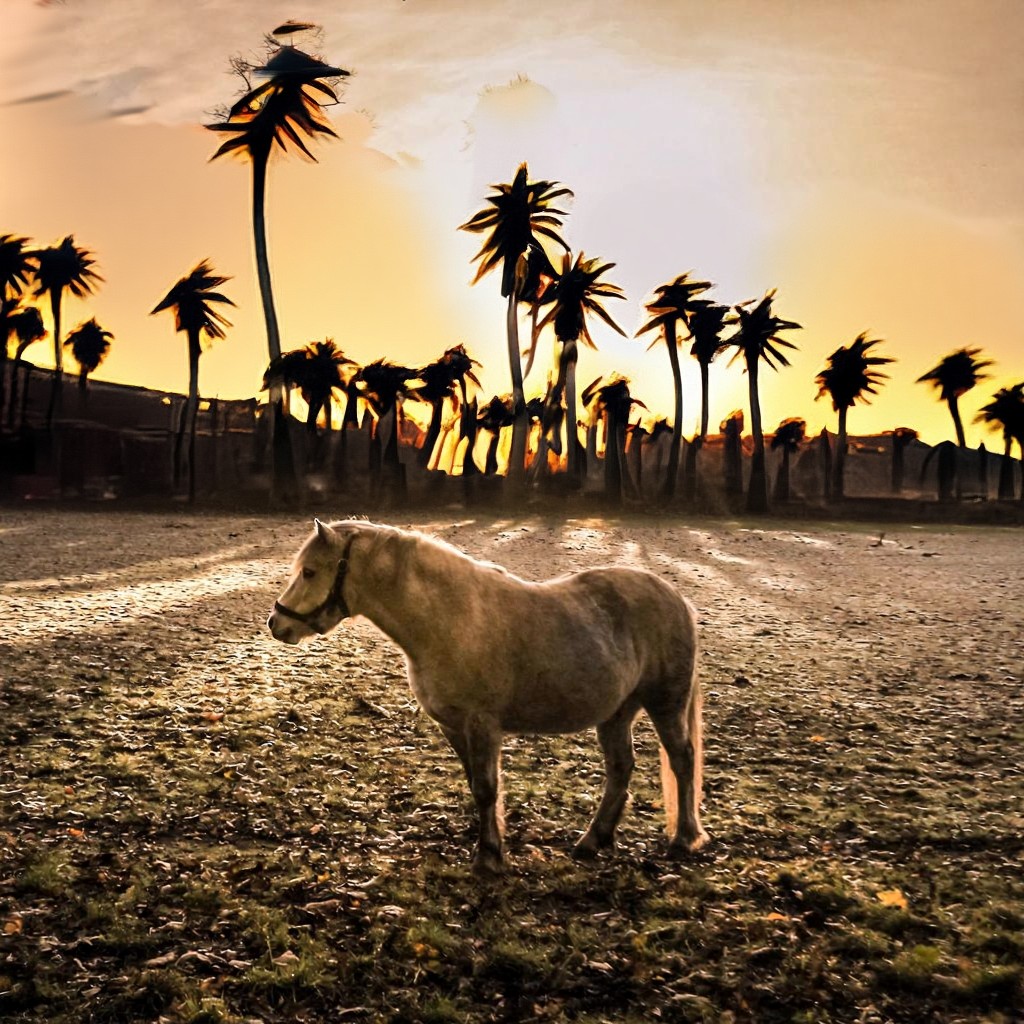}
{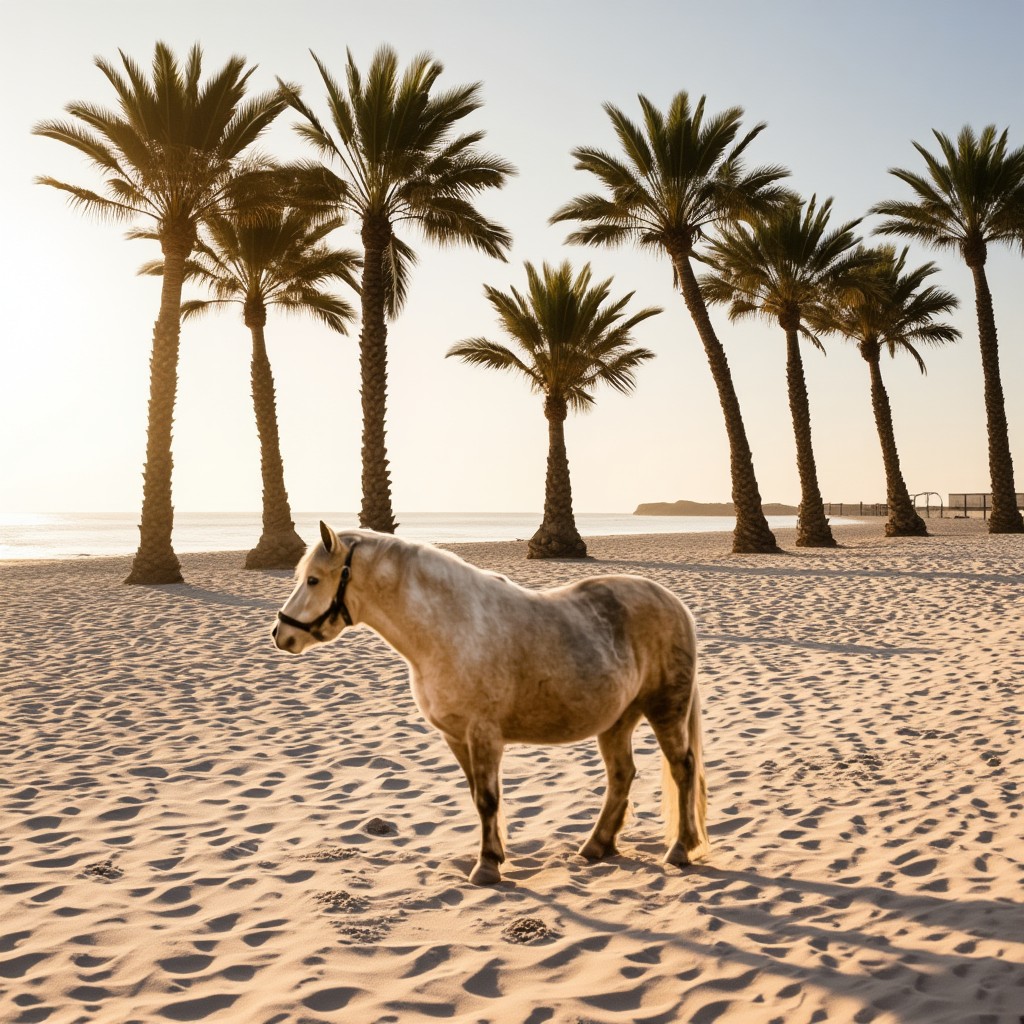}
{}
\par\vspace{5.2pt}
\noindent
\imgeditpanel
{Add a small brown dog inside the open trunk of the antique car.}
{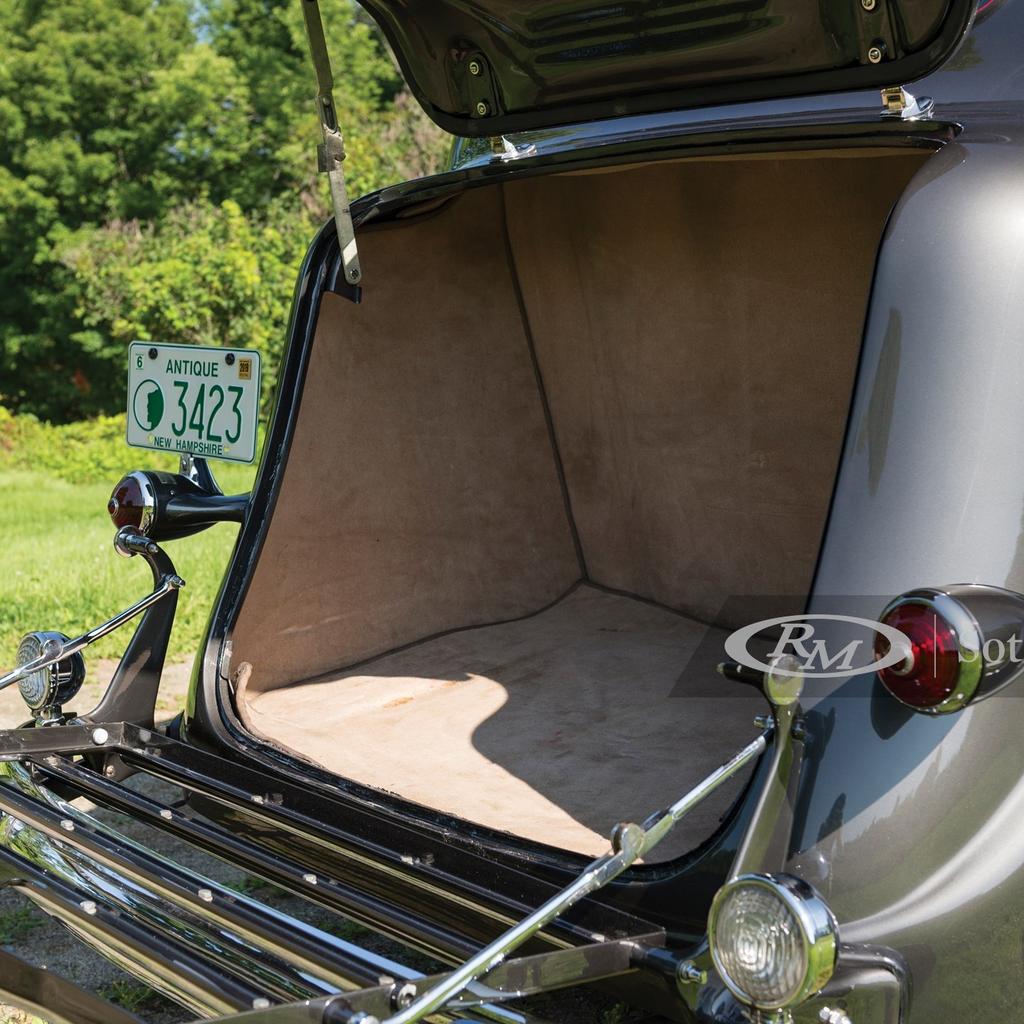}
{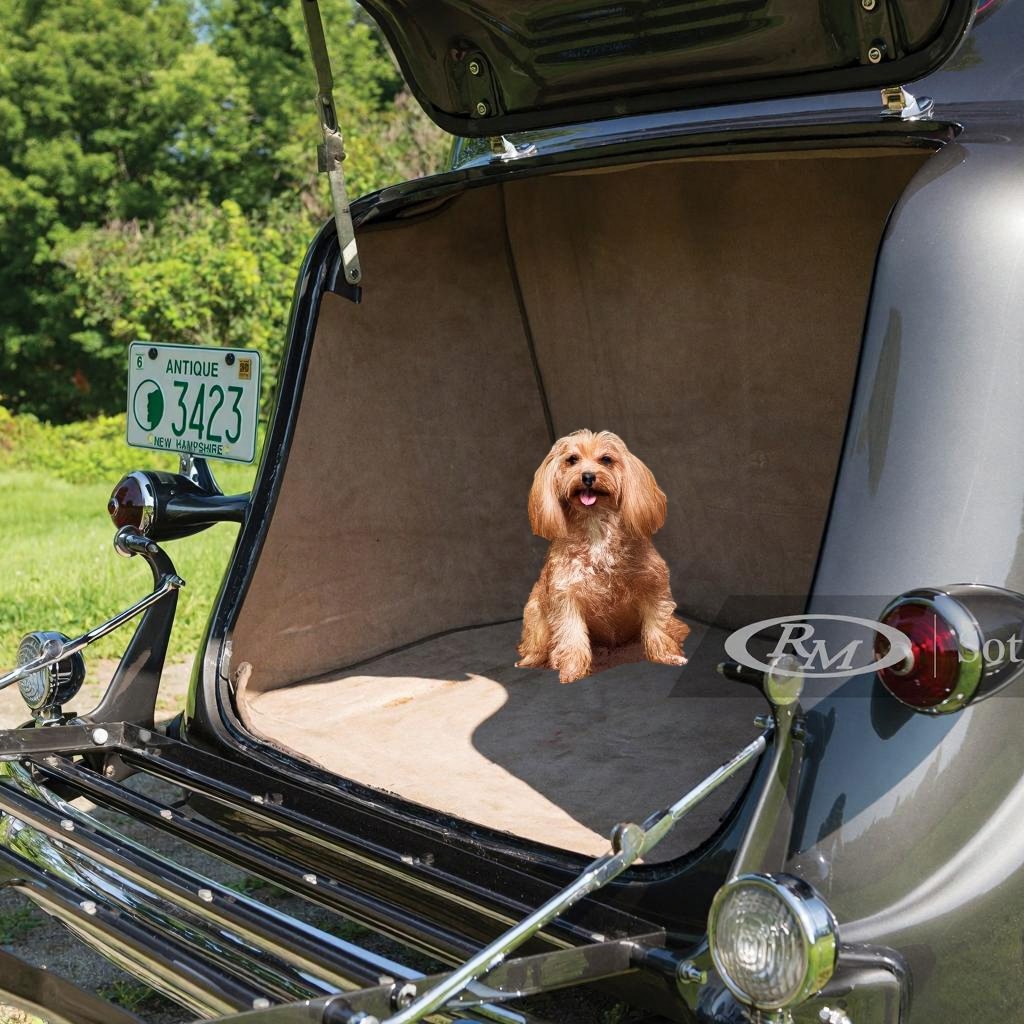}
{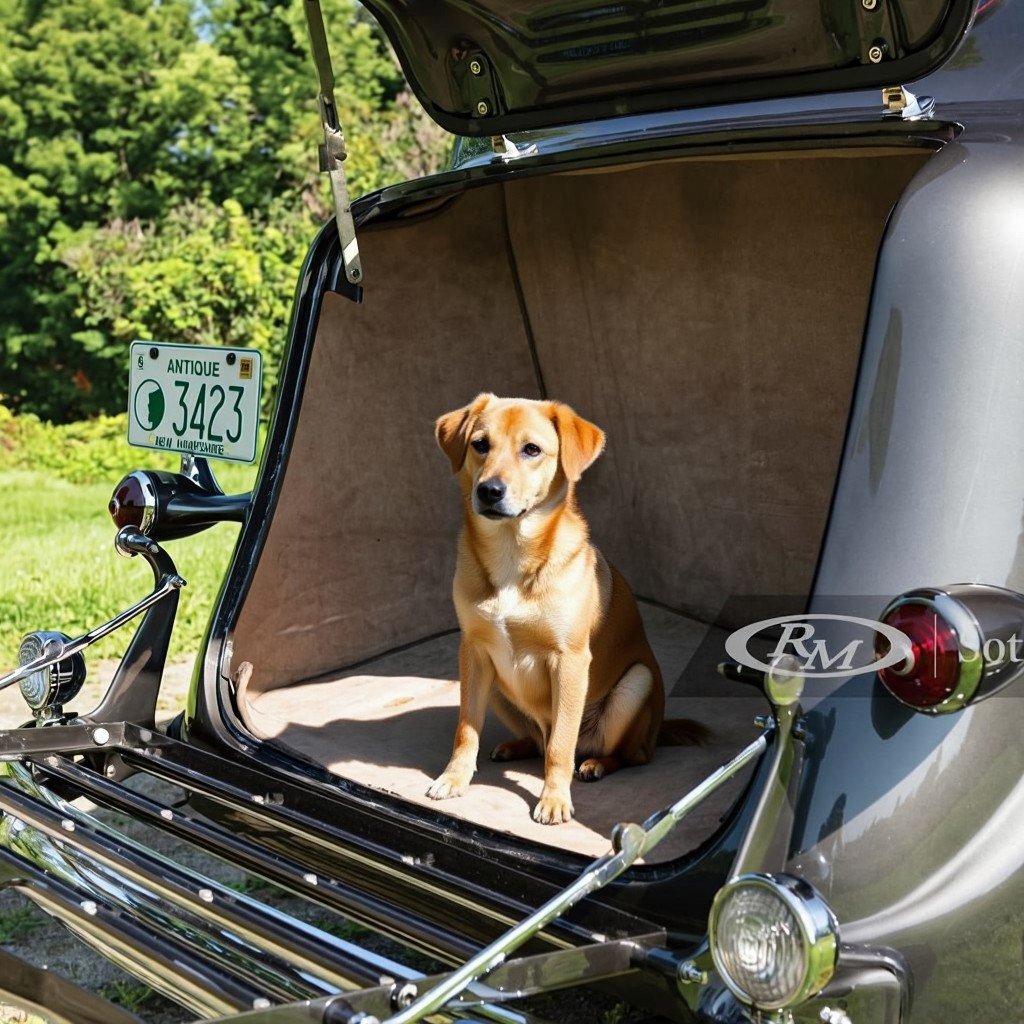}
{}
\imgeditpanelsep
\imgeditpanel
{Make the person lift his head slightly.}
{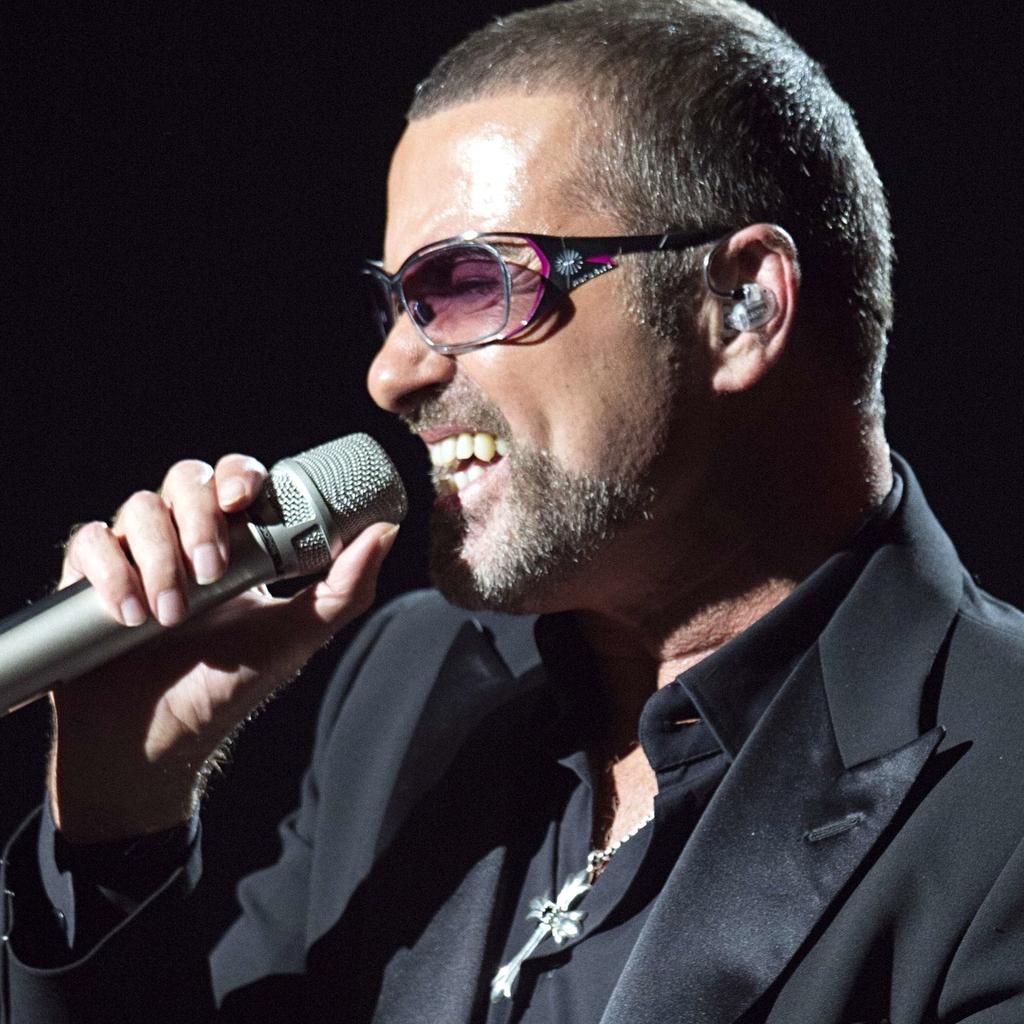}
{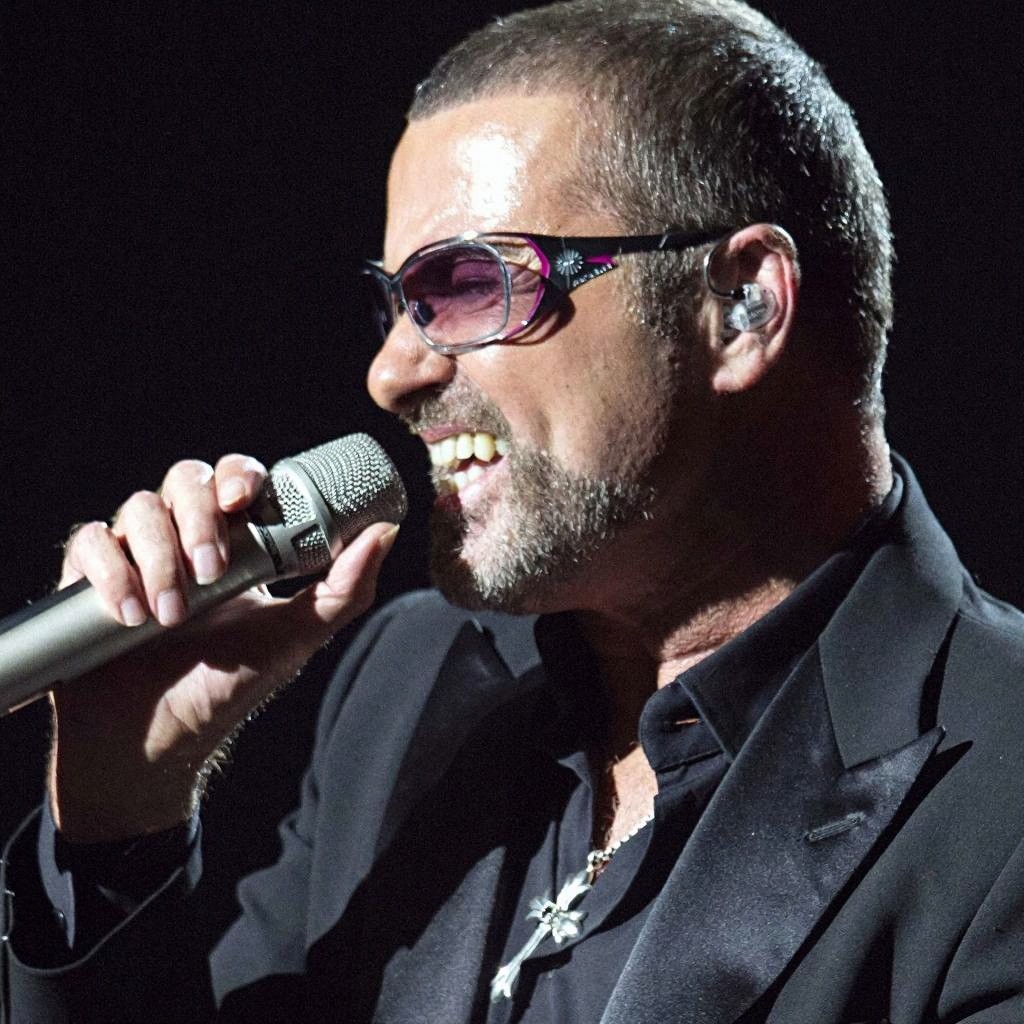}
{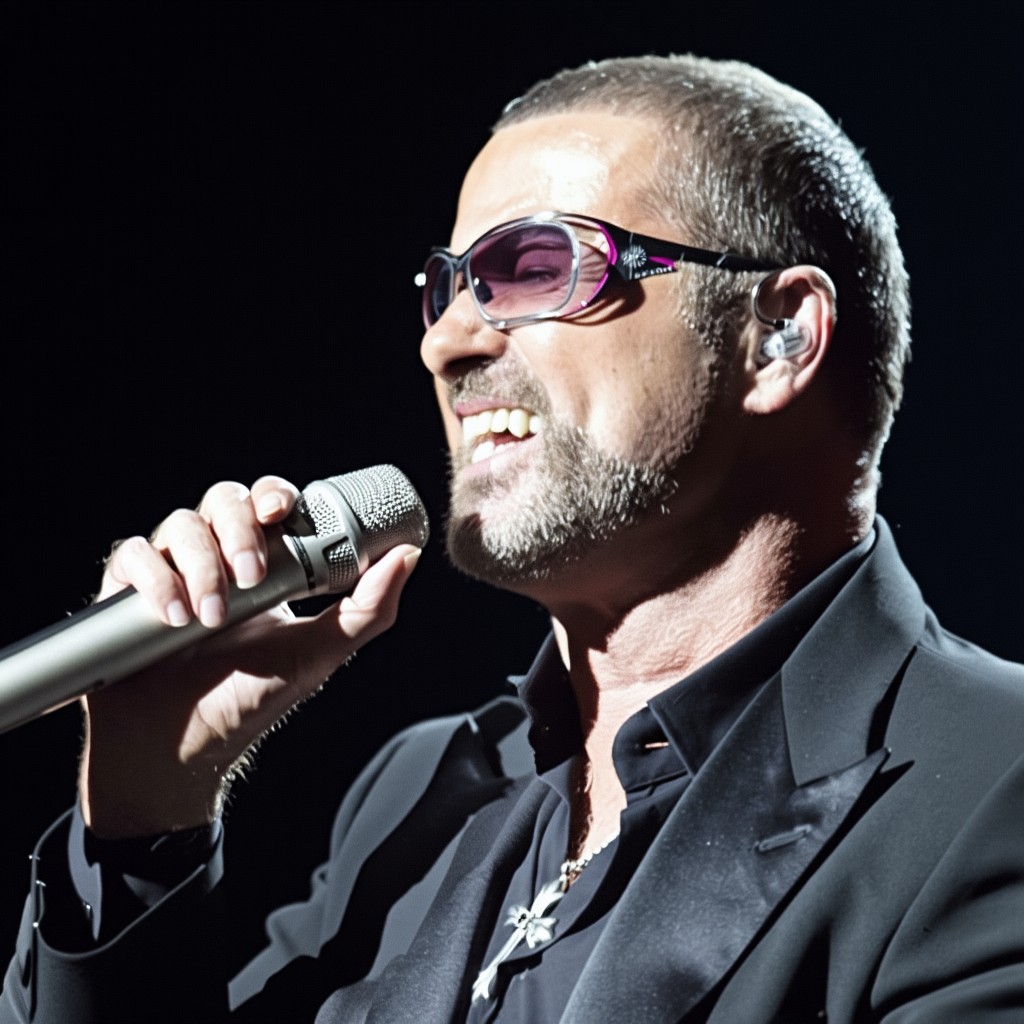}
{}
\end{minipage}
\caption{\textbf{ImgEdit comparisons.} Qualitative ImgEdit benchmark comparisons are arranged in a four-row, two-column grid. Each panel places the editing instruction on top and compares the source image, the BAGEL baseline output, and the Ours output in parallel. All test images shown above are selected from the open-source benchmark ImgEdit~\cite{ye2026imgedit}.}
\label{fig:imgedit_compare_layout}
\endgroup
\end{figure}

\paragraph{Generation and editing.}
Figs.~\ref{fig:layout_mosaic} and~\ref{fig:wise_qualitative_layout} show representative generation results, where STBridge renders diverse subjects, layouts, and knowledge-conditioned scenes from explicit target expressions. Figs.~\ref{fig:imgedit_compare_layout} and~\ref{fig:rise_qualitative_layout} further present instruction-guided and knowledge-guided editing cases. Compared with BAGEL, STBridge-RL better preserves source-image content while applying the requested changes, including object replacement, removal, style transfer, background modification, and spatially or causally constrained edits.

\clearpage
\clearpage

\clearpage

\subsection{Additional Analysis}

\paragraph{Understanding-generation alignment analysis.}
To directly examine whether the understanding-side textual expression is aligned with the generation-side visual output, we conduct an understanding-generation alignment analysis on 400 image-editing pairs. For each pair, we execute the model twice under the two sides of the editing interface. On the understanding side, the model takes the source image, the editing instruction, and a system prompt that asks it to describe the target edited result, and outputs a target caption. On the generation side, the model takes the same source image and instruction and directly generates the edited image. We then use Gemini-3.1-Flash-Image-Preview as the VLM judge to evaluate the consistency between the predicted target caption and the generated edited image.

The evaluation contains one global dimension and five fine-grained dimensions. The global dimension measures whether the caption matches the generated image in scene type, lighting, viewpoint, and composition. The fine-grained dimensions evaluate entity presence, entity attributes, entity-part attributes, spatial relations, and inter-attribute relations. We report the average of the five fine-grained dimensions as \emph{Fine-grained Avg.}, and compute \emph{Overall} by averaging the global score and the fine-grained average.

\begin{table}[!htbp]
\centering
\caption{Understanding-generation alignment analysis. We evaluate the consistency between the understanding-side target caption and the generation-side edited image using Gemini-3.1-Flash-Image-Preview as the VLM judge. Higher scores indicate better alignment.}
\label{tab:ug_alignment}
\small
\setlength{\tabcolsep}{7pt}
\definecolor{alignhighlight}{RGB}{40,144,208}
\newcommand{\alignbest}[1]{\cellcolor{alignhighlight!18}\textbf{#1}}
\newcommand{\aligngain}[1]{\textcolor[RGB]{40,144,208}{\textbf{#1}}}
\begin{tabular*}{\textwidth}{@{\extracolsep{\fill}}l c c c c}
\toprule
Metric & BAGEL & STBridge-SFT & STBridge-RL & Gain \\
\midrule
Global Scene Alignment
& 0.6190 & 0.7985 & \alignbest{0.9418} & \aligngain{+0.3228} \\
Entity Presence
& 0.6648 & 0.7558 & \alignbest{0.9275} & \aligngain{+0.2627} \\
Entity Attribute
& 0.5374 & 0.6819 & \alignbest{0.8534} & \aligngain{+0.3160} \\
Entity Part Attribute
& 0.3857 & 0.5998 & \alignbest{0.7618} & \aligngain{+0.3761} \\
Spatial Relation
& 0.6080 & 0.7613 & \alignbest{0.9153} & \aligngain{+0.3073} \\
Inter-Attribute Relation
& 0.4489 & 0.6721 & \alignbest{0.8330} & \aligngain{+0.3841} \\
\midrule
Fine-grained Avg.
& 0.5290 & 0.6942 & \alignbest{0.8582} & \aligngain{+0.3292} \\
Overall
& 0.5740 & 0.7464 & \alignbest{0.9000} & \aligngain{+0.3260} \\
\bottomrule
\end{tabular*}
\end{table}

As shown in Tab.~\ref{tab:ug_alignment}, BAGEL shows limited alignment between its target caption and edited image, especially on fine-grained dimensions such as entity-part attributes and inter-attribute relations. This suggests that the understanding-side target-expression capability and the generation-side visual-realization capability are not naturally aligned, even within a unified model. STBridge-SFT improves the overall alignment score from 0.5740 to 0.7464, suggesting that establishing the shared-target channel already strengthens the connection between target caption and edited image. STBridge-RL further improves the score to 0.9000, with consistent gains across both global and fine-grained dimensions. The largest improvements appear on entity-part attributes and inter-attribute relations, indicating that STBridge particularly improves alignment on local details and intra-entity relational information. These results provide direct evidence that aligning understanding and generation through a shared target improves the consistency between what the model describes and what it generates.

\paragraph{Comparison of SFT paradigms.}
To examine whether the gain of STBridge-SFT comes merely from additional caption supervision, we compare different SFT paradigms on BLINK. We name the two caption-only variants \emph{Source-Caption SFT} and \emph{Target-Caption SFT}. Source-Caption SFT trains the understanding side to predict the source caption from the source image, corresponding to generic source-image caption supervision. Target-Caption SFT trains the understanding side to predict the target caption from the source image and editing instruction, i.e., \((I_s,x)\rightarrow c_t^*\), but does not supervise target-image realization. STBridge-SFT further jointly supervises target-caption prediction and target-image realization under the same shared target. This comparison separates generic caption learning, target-expression-only learning, and the proposed shared-target channel establishment.

As shown in Tab.~\ref{tab:sft_paradigm_comparison}, Source-Caption SFT and Target-Caption SFT both underperform the BAGEL baseline on average, reaching 57.20 and 56.27 compared with 60.88. In contrast, STBridge-SFT improves the BLINK average to 63.85, yielding a +2.97 gain over BAGEL. These results indicate that simply adding caption supervision, even when the caption corresponds to the edited target result, is insufficient. The improvement appears when target-caption prediction is coupled with target-image realization under the same shared target, supporting the importance of establishing a shared-target channel rather than optimizing the textual target expression alone.

\begin{table}[!htbp]
\centering
\caption{Comparison of SFT paradigms on BLINK. Higher scores are better.}
\label{tab:sft_paradigm_comparison}
\definecolor{comparisonhighlight}{RGB}{40,144,208}
\newcommand{\comparisonbest}[1]{\cellcolor{comparisonhighlight!18}\textbf{#1}}
\resizebox{\textwidth}{!}{
\begin{tabular}{l c c c c c c c c c c c c c c c}
\toprule
Model & Visual Sim. & Count. & Rel. Depth & Jigsaw & Art Style & Func. Corr. & Sem. Corr. & Spatial Rel. & Obj. Loc. & Visual Corr. & Multi-view & Rel. Reflect. & Forensic & IQ Test & Avg. \\
\midrule
BAGEL
& 89.63 & 74.17 & 86.29 & \comparisonbest{83.33} & \comparisonbest{76.92} & 30.77 & 36.69 & 81.82 & 66.39 & 73.84 & 51.88 & 32.09 & 37.88 & 30.67 & 60.88 \\
Source-Caption SFT
& \comparisonbest{91.85} & 71.67 & 86.29 & 71.33 & 56.41 & 32.31 & 42.45 & 76.22 & 69.67 & 73.26 & 42.11 & 32.09 & 26.52 & 28.67 & 57.20 \\
Target-Caption SFT
& \comparisonbest{91.85} & 71.67 & \comparisonbest{88.71} & 63.33 & 51.28 & \comparisonbest{33.85} & 42.45 & 75.52 & 66.39 & 69.19 & 44.36 & \comparisonbest{35.82} & 22.73 & 30.67 & 56.27 \\
STBridge-SFT
& 89.63 & \comparisonbest{75.83} & 86.29 & \comparisonbest{83.33} & \comparisonbest{76.92} & \comparisonbest{33.85} & \comparisonbest{46.04} & \comparisonbest{83.22} & \comparisonbest{72.95} & \comparisonbest{77.91} & \comparisonbest{56.39} & 35.07 & \comparisonbest{43.18} & \comparisonbest{33.33} & \comparisonbest{63.85} \\
\bottomrule
\end{tabular}
}
\end{table}

\paragraph{Comparison of Target-Expression Optimization rewards.}
We further compare different reward designs for Target-Expression Optimization. The direct caption--target reward computes the similarity score between the rollout target caption and the target image. Although this design is straightforward, it is vulnerable during RL optimization: since the reward only measures caption--target similarity, the model can increase the reward by producing target-image descriptions that are not directly related to the editing instruction, such as verbose background or scene-level information. In contrast, Diff Reward is designed to focus the target expression on edit-relevant visual changes rather than rewarding generic target-image similarity.

\begin{table}[!htbp]
\centering
\caption{Comparison of Target-Expression Optimization rewards on BLINK. Higher scores are better.}
\label{tab:target_expression_reward_comparison}
\definecolor{rewardhighlight}{RGB}{40,144,208}
\newcommand{\rewardbest}[1]{\cellcolor{rewardhighlight!18}\textbf{#1}}
\resizebox{\textwidth}{!}{
\begin{tabular}{l c c c c c c c c c c c c c c c}
\toprule
Model & Visual Sim. & Count. & Rel. Depth & Jigsaw & Art Style & Func. Corr. & Sem. Corr. & Spatial Rel. & Obj. Loc. & Visual Corr. & Multi-view & Rel. Reflect. & Forensic & IQ Test & Avg. \\
\midrule
BAGEL
& 89.63 & 74.17 & 86.29 & \rewardbest{83.33} & \rewardbest{76.92} & 30.77 & 36.69 & 81.82 & 66.39 & 73.84 & 51.88 & 32.09 & 37.88 & 30.67 & 60.88 \\
STBridge-SFT
& 89.63 & \rewardbest{75.83} & 86.29 & \rewardbest{83.33} & \rewardbest{76.92} & 33.85 & 46.04 & 83.22 & 72.95 & 77.91 & \rewardbest{56.39} & \rewardbest{35.07} & \rewardbest{43.18} & \rewardbest{33.33} & 63.85 \\
Direct Caption--Target Reward
& 91.11 & 75.00 & 85.48 & \rewardbest{83.33} & \rewardbest{76.92} & 36.15 & 43.17 & 82.52 & 74.59 & 80.23 & 55.64 & \rewardbest{35.07} & 42.42 & 32.67 & 63.88 \\
Diff Reward
& \rewardbest{92.59} & 73.33 & \rewardbest{92.74} & 82.67 & \rewardbest{76.92} & \rewardbest{42.31} & \rewardbest{51.08} & \rewardbest{83.92} & \rewardbest{87.70} & \rewardbest{84.88} & 55.64 & 32.84 & 37.88 & 30.00 & \rewardbest{66.03} \\
\bottomrule
\end{tabular}
}
\end{table}

As shown in Tab.~\ref{tab:target_expression_reward_comparison}, Direct Caption--Target Reward only marginally changes the BLINK average from 63.85 to 63.88, suggesting that the direct similarity reward provides almost no additional benefit over STBridge-SFT. In contrast, Diff Reward raises the average to 66.03, yielding a +2.18 gain over STBridge-SFT and a +2.15 gain over the direct reward variant.

\FloatBarrier

\section{Conclusion}

We presented \emph{STBridge}, a shared-target alignment framework for UMM post-training that bridges visual understanding and visual generation through a common target state. In the image-editing setting, the target image defines this shared target: the understanding pathway expresses it as a target caption, while the generation pathway realizes it as the edited image. STBridge is trained through a three-stage post-training procedure. Shared-Target Alignment first establishes the correspondence between target expression and visual realization, while Target-Expression Optimization and Target-Realization Optimization further refine the textual target expression and the generated visual result. Across seven benchmarks, STBridge-RL consistently improves over BAGEL on visual understanding (BLINK +5.15, OddGridBench +5.14, MMVP +2.70), image generation (WISE +1.20), and image editing (ImgEdit +0.46, RISE +21.00). Alignment analysis further shows that caption--image consistency increases from 0.5740 to 0.9000 under VLM evaluation. Ablation studies confirm that the gain comes from coupling target expression with target realization under the same shared target, rather than from isolated caption supervision or direct similarity optimization. These results indicate that shared-target alignment is an effective post-training strategy for bridging understanding and generation in UMMs. Future work includes extending STBridge to broader model families and data regimes, and exploring shared-target alignment beyond image editing.

\FloatBarrier

\bibliography{editbridge}

\end{document}